\documentclass[12pt]{article}
\usepackage{comment}
\usepackage{graphicx,psfrag,epsf}
\usepackage{enumerate}
\usepackage{natbib}
\usepackage{url} 

\usepackage{microtype}
\usepackage{subfigure}
\usepackage{booktabs}

\usepackage{amsmath}
\DeclareMathOperator*{\argmax}{arg\,max}

\usepackage{bbm}
\usepackage{amssymb}
\usepackage{mathtools}
\usepackage{amsthm}

\usepackage{algorithm}
\usepackage{algcompatible}

\usepackage{hyperref}
\usepackage{xcolor}
\usepackage{authblk}

\theoremstyle{plain}
\newtheorem{theorem}{Theorem}[section]
\newtheorem{proposition}[theorem]{Proposition}
\newtheorem{lemma}[theorem]{Lemma}

\theoremstyle{definition}
\newtheorem{definition}[theorem]{Definition}
\newtheorem{assumption}[theorem]{Assumption}
\theoremstyle{remark}

\newcommand{\blind}{0}

\begin{document}

\def\spacingset#1{\renewcommand{\baselinestretch}%
{#1}\small\normalsize} \spacingset{1}


  \title{\bf Deep Spectral Q-learning 
with Application to Mobile Health}

\author[1]{Yuhe Gao\thanks{ygao32@ncsu.edu}}
\author[2]{Chengchun Shi\thanks{c.shi7@lse.ac.uk}}
\author[1]{Rui Song\thanks{rsong@ncsu.edu}}
\affil[1]{Department of Statistics, North Carolina State University, Raleigh, USA}
\affil[2]{Department of Statistics, London School of Economics and Political Science, London, UK}




  \maketitle

\if1\blind
{
  \bigskip
  \bigskip
  \bigskip
  \begin{center}
    {\LARGE\bf Deep Spectral Q-learning 
with Application to Mobile Health}
\end{center}
  \medskip
} \fi

\bigskip
\begin{abstract}
Dynamic treatment regimes assign personalized treatments to patients sequentially over time based on their baseline information and time-varying covariates. 
In mobile health applications, 
these covariates are typically collected at different frequencies over a long time horizon. In this paper, we propose a deep spectral Q-learning algorithm, which integrates principal component analysis (PCA) with deep Q-learning to handle the mixed frequency data. In theory, we prove that the mean return under the estimated optimal policy converges to that under the optimal one and establish its rate of convergence. The usefulness of our proposal is further illustrated via simulations and an application to a diabetes dataset.
\end{abstract}

\noindent%
\vfill

\newpage
\spacingset{1.45} 

\section{Introduction}
\label{Introduction}
Precision medicine 
focuses on providing personalized treatment to patients by 
taking their personal information into consideration \citep[see e.g.,][]{Kosorok2019precisionreview,tsiatis2019dynamic}. 
It has found various applications in numerous studies, 
ranging from the cardiovascular disease study to cancer treatment and gene therapy \citep{Jameson2015precisionmedicine}.
A dynamic treatment regime (DTR) 
consists of a sequence of treatment decisions rules tailored  to each individual patient's status at each time, mathematically formulating the idea behind precision medicine. 
One of the major objectives in precision medicine is to identify the optimal dynamic treatment regime that yields the most favorable outcome on average. 

With the rapidly development of mobile health (mHealth) technology, it becomes feasible to collect rich longitudinal data 
through mobile apps in medical studies. 
A motivating data example is given by the OhioT1DM dataset \citep{Marling2018ohiodata} which contains data from 12 patients suffering from type-I diabetes measured via fitness bands over 8 weeks. Data-driven decision rules estimated from these data have the potential to improve these patients' health \citep[see e.g.,][]{shi2020does,zhu2020causal,zhou2022estimating}. However, it remains challenging to estimate the optimal DTR in these mHealth studies. 
First, the number of treatment stages (e.g., horizon) is no longer fixed whereas the number of patients can be limited. For instance, in the OhioT1DM dataset, only 12 patients are enrolled in the study. Nonetheless, suppose treatment decisions are made on an hourly basis, the horizon is over 1000. Existing proposals in the DTR literature \citep{murphy2003optimal, zhang2013robust, zhao2015new, song2015penalized, shi2018maximin, nie2021learning, fang2022fairness, qi2018d, mo2021learning, ertefaie2021robust, guan2020bayesian, zhang2018interpretable}  become inefficient in these long or infinite horizon settings and require a large number of patients to be consistent. 
Second, patients' time-varying covariates typically contain mixed frequency data. In the OhioT1DM dataset, some of the variables, such as the continuous glucose monitoring (CGM) blood glucose levels, are recorded every 5 minutes. Meanwhile, other variables, such as the  carbohydrate estimate for the meal and the exercise intensity are recorded with a much lower frequency. 
Concatenating these high-frequency variables over each one-hour interval produces a high-dimensional state vector and directly using these states as input of the treatment policy would yield very noisy decision rules. 
A naive approach is to use some ad-hoc summaries of the high-frequency data for policy learning. However, this might produce a sub-optimal policy due to the information loss. 

Recently, 
there is a growing line of research in the statistics literature for policy learning and/or evaluation in infinite horizons. Some references include \citet{Ertefaie2018Gradient, Luckett2020Vlearn, Shi2020SAVE, liao2020batch,  liao2021off, shi2021dynamic, chen2022reinforcement, ramprasad2022online, li2022rate, xu2020latent}. 
In the computer science literature, there is a huge literature on developing reinforcement learning (RL) algorithms in infinite horizons. These algorithms can be casted into as model-free and model-based algorithms. Popular model-free RL methods include  
value-based 
methods that model the expected return starting from a given state (or state-action pair) 
and 
compute the optimal policy as the greedy policy with respect to the value function \citep{Ernst2005FittedQiteration,Riedmiller2005neuralFQI,mnih2015dqn,Hasselt2016doubelDQN,Dabney2018quantileDQN}, and policy-based methods 
that directly 
search the optimal policy among a parameterized class via policy gradient or action critic methods \citep{Schulman2015trustregion,Schulman2016GAE,Cuccu2013evolution,mnih2016a3c,Wang2017experiencereplay}. Model-based algorithms are different from the model-free algorithms in the sense that they model the transition dynamics of the environment and use the model of environment to derive or improve policies. Popular model-based RL methods include \citet{guestrin2002algorithm, janner2019trust, lai2020bidirectional, li2020breaking}, to name a few.   
See also \citet{Arulkumaran2017reviewRL,Sutton2018RLoverview,luo2022survey} for more details. These methods cannot be directly applied to datasets such as OhioT1DM as they haven't considered the mixed frequency data. 
In the RL literature, 
a few works have considered dimension reduction to handle the high dimensional state system. 
In particular, \citet{QLASS1997Murao} 
proposed to segment the state space and learn a cluster representation of the states. 
\citet{tilecoding2007Whiteson} 
proposed to divide the state space into tilings to represent each state.  
Both papers proposed to discretize the state space for dimension reduction. However, 
it can lead to considerable information loss \citep{wangeric2017sufficient}.
\citet{Sprague2007Basis} proposed a iterative dimension reduction method using neighborhood components analysis. 
Their method 
uses a linear basis function to model the Q-function and cannot allow more general nonlinear function approximation. 
Recently, there are a few works that employ principal components analysis (PCA) for dimension reduction in RL \citep{curran2015using,curran2016dimensionality,parisi2017goal}. However, none of the aforementioned papers formally established the theoretical guarantees for their proposals. Moreover, these methods are motivated by applications in games or robotics and their generalization to mHealth applications with mixed frequency data remains unknown.  

In the DTR literature, 
a few works considered mixed frequency data which include both scalar and functional covariates. Specifically, \citet{McKeague2014functional} 
proposed a functional regression model 
for optimal decision making with one functional covariate. 
\citet{Ciarleglio2015Functional} and \citet{Ciarleglio2016functional} extended their proposal to a more general setting 
with multiple scalar and functional covariates.  
\citet{unknownfunction2018Ciarleglio} considered variable selection to handle the mixed frequency data. 
\citet{Staicu2018PCA} applied functional PCA to the functional covariates for dimension reduction. 
All these works considered single stage decision making. Their methods are not directly applicable to the infinite horizon settings. 

Our contributions are as follows. Scientifically, mixed frequency data frequently arise in mHealth applications. Nonetheless, it has been less explored in the infinite horizon settings. Our proposal thus fills a crucial gap, and greatly extends the scope of existing approaches to learning DTRs. 
Methodologically, we propose a deep spectral 
Q-learning algorithm for dimension reduction. 
The proposed algorithm achieves a better empirical performance than those that either directly use the original mixed frequency data or its ad-hoc summaries as input of the treatment policy. 
Theoretically, we derive an upper error bound for the regret of the proposed policy and decompose this bound into the sum of approximation error and estimation error. 
Our theories offer some general guidelines for practitioners to select the number of principal components 
in the proposed algorithm. 

The rest of this paper is organized as follows. We introduce some background about DTR and the mixed frequency data in Section \ref{s:Preliminary}. We introduce the proposed method to estimate the optimal DTR in Section \ref{s:Method} 
and study its theoretical properties in Section \ref{s:asymp_property}. 
Empirical studies are conducted in Section \ref{s:numerical}. In Section \ref{s:real_data}, we apply the proposed method to the OhioT1DM dataset. Finally, we conclude our paper in Section \ref{s:discussion}. 

\section{Preliminary}\label{s:Preliminary}

\subsection{Data and Notations}\label{s: notations}

Suppose 
the study enrolls $N$ patients. 
The dataset for the $i$-th patient can be summarized as
$O_{i}\equiv \{(S_{i,t},A_{i,t},Y_{i,t}):1\le t\le T_i\}$. For simplicity, we assume $T_i=T$ for any $i$. Each state $S_{i,t}=(X_{i,t},\{Z_{i,t,j}\}_{j=1}^{J})\in\mathcal{S}$ 
is composed of a set of low frequency covariates $X_{i,t}\in\mathbb{R}^{m_0}$ and a set of $J$ high frequency covariates $\{Z_{i,t,j}\}_{j=1}^{J}$, where $\mathcal{S}$ is the state space. 
For each high frequency variable $Z_{i,t,j},j\in\{1,2,..,J\}$, we have  $Z_{i,t,j}=(Z_{i,t,j}^{(1)},Z_{i,t,j}^{(2)},...,Z_{i,t,j}^{(m_j)})^T\in\mathbb{R}^{m_j}$ for some large integer $m_j$. 
Let $\tau$ denote the length of a time unit. 
The low frequency variables 
are recorded 
at time points $\tau,2\tau,\cdots,t\tau,\cdots$. 
The $j$th high frequency variables, however, are recorded more frequently at time points $m_j^{-1}\tau,2m_j^{-1}\tau,\cdots$. 
Notice that we allow the $J$ high-frequency variables to be recorded with different frequencies. 
Let $m=\sum_{j=1}^J{m_j}$ and 
$Z_{i,t}^T=[Z_{i,t,1}^T,Z_{i,t,2}^T,\ldots,Z_{i,t,J}^T]\in\mathbb{R}^{m}$ denote a high-dimensional variable that concatenates all the high frequency covariates. As such, the state $S_{i,t}$ can be represented as $
(X_{i,t},Z_{i,t})\in\mathbb{R}^{m_0+m}$. 
In addition, $A_{i,t}$ denotes the treatment
indicator at the $t$th time point and $\mathcal{A}$ denotes the set of all possible treatment options. The reward, $Y_{i,t}$ corresponds to the $i$th patient's response obtained after the $t$th decision point. By convention, we assume a larger value of $Y_{i,t}$ 
indicates a better outcome. 
We require $|Y_{i,t}|$ to be uniformly bounded by some constant $R_{max}>0$,  
and assume $O_1,O_2,...,O_N$ are $i.i.d.$, which are commonly imposed in the RL literature \citep[see e.g.,][]{Sutton2018RLoverview}. 
Finally, we denote the $l_p$-norm of a function aggregated over a given  distribution function $\sigma$ by $\left\Vert.\right\Vert_{p,\sigma}$. We use $[q]$ to represent the indices set $\{1,2,3,...q\}$ for any integer $q\in\mathbb{N}$.

\subsection{Assumptions, Policies and Value Functions}\label{s:assump_policy_value_function}
We will require the system to satisfy the 
Markov assumption such that
\begin{equation*}\label{assum: markov}
    \begin{aligned}
    P(S_{0,t+1}\in \mathbb{S}|S_{0,t}=s,A_{0,t}=a,\{S_{0,t'},A_{0,t'}\}_{0\le t'<t})=P(S_{0,t+1}\in \mathbb{S}|S_{0,t}=s,A_{0,t}=a) ,\forall t, 
    \end{aligned}
\end{equation*} for any $s,a$ and Borel set $\mathbb{S}\in \mathcal{S}$. 
In other words, the distribution of the next state 
depends on the past data history only through the current state-action pair. 
We assume the transition kernel is absolutely continuous with respect to some uniformly bounded transition density function $q(s'|s,a)$ such that $\sup_{s,a,s'}|q(s'|s,a)|\le c_q$ for some constant $c_q>0$. 

In addition, we also impose the following conditional mean independence assumption: 
\begin{equation*}\label{assum: cmia}
    \begin{aligned}
     E(Y_{0,t}|S_{0,t}=s,A_{0,t}=a, \{Y_{0,t'},S_{0,t'},A_{0,t'}\}_{0\le t'<t})=E(Y_{0,t}|S_{0,t}=s,A_{0,t}=a)=r(s,a),\forall t,
    \end{aligned}
\end{equation*} 
where we refer to $r(s,a)=r(x,\{z_j\}_{j=1}^{J},a)$ as the immediate reward function. 

Next, define a policy $\pi:\mathcal{S}\rightarrow P(\mathcal{A})$ as a function that maps a patient’s state at each time point into a probability distribution function on the action space. 
Given $\pi$, we define its (state) value function 
as 
\begin{equation*}
\begin{aligned}
 V^{\pi}(s)=\sum_{t=0}^{\infty}{\gamma^{t}E^{\pi}\{Y_{0,t}|S_{0,0}=s\}},
\end{aligned}
\end{equation*} 
with $\gamma\in[0,1)$ being a discount factor that balances the immediate and long-term rewards. By definition, the state value function 
characterizes the expected return assuming the decision process follows a given target policy $\pi$. 
In addition, we define the state-action value function (or Q-function) as \begin{equation*}
    Q^{\pi}(s,a)=\sum_{t=0}^{\infty}{\gamma^{t}E^{\pi}\{Y_{0,t}|S_{0,0}=s,A_{0,0}=a\}}, 
\end{equation*} 
which is the expected discounted cumulative rewards given an initial state-action pair. 

Under the Markov assumption and the conditional mean independence assumption, there exists an optimal policy $\pi^*$ such that $V^{\pi^*}(s)\ge V^{\pi}(s),\forall \pi, s\in\mathcal{S}$ \citep[see e.g.,][]{Puterman2014MDP}. 
Moreover, $\pi^*$  satisfies the following Bellman optimality equation: 
\begin{equation}\label{Bellmanoptimal}
 \begin{aligned}
     Q^{\pi^*}(s,a)=E\{Y_{0,t}+\gamma \max_{a'\in\mathcal{A}}Q^{\pi^*}{(S_{0,t+1},a')}|S_{0,t}=s,A_{0,t}=a\},
 \end{aligned}
 \end{equation}
 where $Q^{\pi^*}$ denotes the optimal Q-function. 

\subsection{ReLU Network}\label{s: neuralnet}
In this paper, we use value-based methods that learn the optimal policy $\pi^*$ 
by estimating the optimal Q-function. We will use the class of sparse neural network with the Rectified Linear Unit (ReLU) activation function, i.e., $f(x)=\max(x,0)$, to model the Q-function. The advantage of using a neural network over a simple parametric model is that the neural network can better capture the potential non-linearity in 
the high-dimensional state system. 

\begin{figure}[h!]
\centerline{\includegraphics[width=0.6\textwidth]{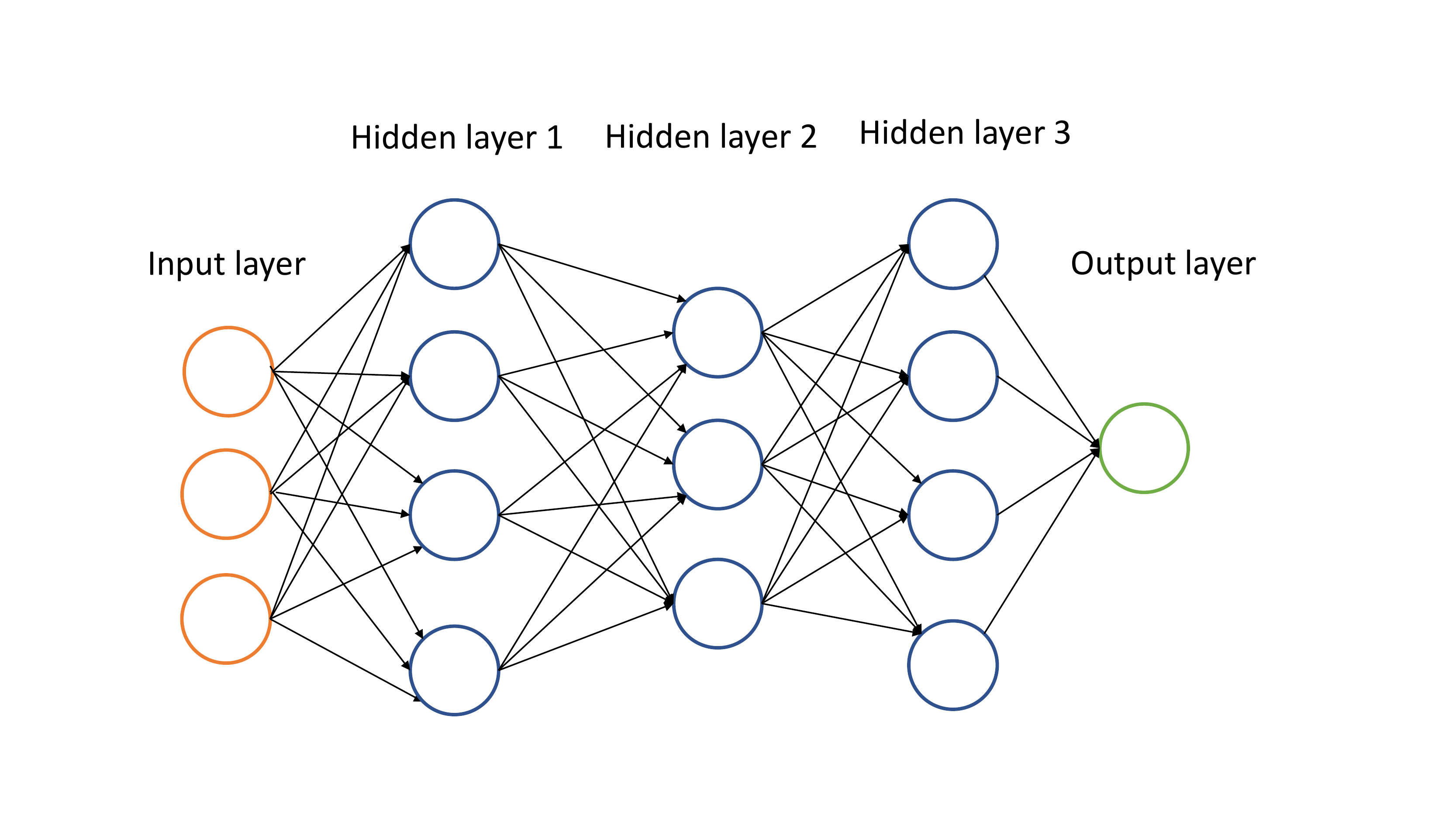}}
\caption{Illustration of a ReLU network.}\label{fig:illustrate_ReLU}
\end{figure}

Formally, the class of sparse ReLU network is defined as \begin{equation*}
\begin{aligned}
\mathcal{F}_{SReLU}(L,\{d_j\}_{j=0}^{L+1},s,V_{max})=\{\mathbb{R}^{d_{0}}\rightarrow\mathbb{R}^{d_{L+1}}:f(x)=W_Lg_{L-1}\circ g_{q-1}\circ...g_{1}\circ g_{0}(x),\\g_j(x)=\sigma(W_j x + v_j), W_j \in \mathbb{R}^{d_{j+1}\times d_{j}},v_j\in \mathbb{R}^{d_{j+1}},j \in\{0,1,...,L\},  \\\max_{j=0,1,...,L}{\{\max(\left\Vert W_j\right\Vert_{\infty},|v_j|_\infty)}\}\le1, \sum_{j=0}^L{(\left\Vert W_j\right\Vert_{0}+|v_j|_0)}\le s,\max_{k\in\{1,2,...,d_{L+1}\}}{\left\Vert f_k\right\Vert_\infty}\le V_{max}\}.
\end{aligned}
\end{equation*} Here, $L$ is the number of hidden layers of the neural network and $d_j$ is the width of each layer. 
The output dimension  $d_{L+1}$ is set to 1 
since the Q-function output is a scalar. The parameters in $\mathcal{F}_{ReLU}(L)$ are the weight matrices $W_j$ and bias vectors $v_j$. The sparsity level $s$ upper bounds the 
total number of nonzero parameters in the model. This constraint can be satisfied using dropout layers in the implementation \citep{Srivastava2014dropout}.  
In theory, sparse ReLU networks can fit smooth functions with a minimax optimal rate of convergence  \citep{Schmidt2020ReLU}. 
The main theorems in Section \ref{s:asymp_property} will rely on this property. 
An illustration of sparse ReLU network is in Figure \ref{fig:illustrate_ReLU}.

\section{Spectral Fitted Q-Iteration}
\label{s:Method}
Neural network with ReLU activation functions in \ref{s: neuralnet} is commonly used in value-based reinforcement learning algorithms.  
However, in medical studies, the training dataset 
is often of limited size, with a few thousands or 10 thousands of observations in total \citep[see e.g.,][]{Marling2018ohiodata, liao2021off}. 
Meanwhile, the data contains high frequency state variables, which yields a high-dimensional state system. Directly using these states as input will procedure a very noisy policy. This motivates us to consider dimension reduction in RL. 


A naive approach for dimension reduction is to 
use some summary statistics of the high frequency state as input for policy learning. For instance, on the OhioT1DM dataset, the average of CGM blood glucose levels between two treatment decision points can be used as the summary statistic, as in \citet{zhu2020causal,Shi2020SAVE,zhou2022estimating}. 
In this paper, we propose to use principal component analysis to reduce the dimensionality of $\{Z_{i,t,j}\}_{j=1}^{J}$. We expect that using PCA can preserve more information than some ad-hoc summaries (e.g., average).  

To apply PCA in the infinite horizon setting, we need to impose some stationarity assumptions on the concatenated high dimensional variables $Z_{i,t}^T\in\mathbb{R}^{m}$: $E[Z_{i,t}]=\bf \mathbf{\mu}$ and $Cov[Z_{i,t}]=\mathbf{G}$ for some mean vector $\mathbf{\mu}\in\mathbb{R}^{m}$ and covariance matrix $\mathbf{G}\in\mathbb{R}^{m\times m}$ that are independent of $t$. In real data application, we can test  whether the concatenated high-frequency variable $Z_{i,t}$ is weak stationary \citep[see e.g.,][]{dickey1979distribution, said1984testing, kwiatkowski1992testing}. If it is weak stationary, the concatenated high frequency covariate $Z_{i,t}$ will automatically satisfy the two assumptions above.   
Similar assumptions have been widely imposed in the literature \citep[see e.g.,][]{shi2021deeply,kallus2022efficiently}. Without loss of generality, we assume $\mathbf{\mu} = 0_{m}$ for simplicity of notations.  
For the covariance matrix $\mathbf{G}$, it is generally unknown. 
In practice, we recommend to use the sample covariance estimator $\hat{\mathbf{G}}$.

We describe our procedure as follows. By the spectral decomposition, we have $\mathbf{\hat{G}}=\sum_{k=1}^{m}{\hat{\lambda}_{k}\hat{U}_{k}\hat{U}_{k}^T}$, where $\lambda_{1}\ge\lambda_{2}\ge...\ge0$ are the eigenvalues 
and $\hat{U}_{k}$'s are the corresponding eigenvectors. 
This allows us to represent ${Z}_{i,t}$ as $\sum_{k=1}^{m}{\hat{\lambda}_{k}^{1/2}\hat{V}_{k}^{(i,t)}\hat{U}}_{k}$, where 
$\hat{V}_{k}^{(i,t)}$s are the estimated principal component scores, given by $\hat{\lambda}_{k}^{-1/2}{Z}_{i,t}^T\hat{U}_{k}$. 
For any $\kappa$, the estimated principal component scores $\hat{\mathbf{V}}_{i,t,\kappa}=(\hat{V}_{1}^{(i,t)},\hat{V}_{2}^{(i,t)},...,\hat{V}_{\kappa}^{(i,t)})$  correspond to the $\kappa\le m$ largest eigenvalues of the concatenated high frequency variable $Z_{i,t}$. 
When $\kappa=m$, using these principal component scores is equivalent to using the original high-frequency variable ${Z}_{i,t}$. 
We will approximate $Q^\pi({X}_{i,t},\mathbf{V}_{i,t,m},{A}_{i,t})$ by $Q^\pi({X}_{i,t},\hat{\mathbf{V}}_{i,t,\kappa},{A}_{i,t})$ and propose to use neural fitted Q-iteration algorithm by \citet{Riedmiller2005neuralFQI} to learn the estimated optimal policy. We detail our procedure in Algorithm \ref{alg:algorithm1}.


\begin{algorithm}[!tph]
   \caption{Spectral Fitted Q-Iteration}
   \label{alg:algorithm1}
\begin{algorithmic}

   \STATE {\bfseries Input:} $\{S_{i,t}=(X_{i,t},Z_{i,t}),A_{i,t},Y_{i,t},\gamma\}$ with $i\in\{1,2,...,N\}, t\in\{1,2,...,T\}$; ReLU network function class $\mathcal{F}_{SReLU}=\mathcal{F}_{SReLU}(L,\{d_j\}_{j=0}^{L+1},s,V_{max})$; sampling distribution $\sigma$; sample size $n$; number of principal components $\kappa$; number of iterations $K$; estimated covariance matrix $\hat{\bf G}$ for $Z_{i,t}$;
   \STATE Calculate first $\kappa$ PCA $\hat{\mathbf{V}}_{i,t,\kappa}$ for high frequency data part $Z_{i,t}$;
   \STATE For each $a\in\mathcal{A}$, initialize a sparse ReLU network $\tilde{Q}_0(x,v_{\kappa},a)\in\mathcal{F}_{SReLU}$;
   \FOR{$k=1$ {\bfseries to} $K$}
   \STATE Sample $n$ observations $I_k=\{(i,t):1\le i\le N,1\le t\le T-1\}$ based on $\sigma$ from data;
   \STATE Define a response $R_{i,t}$ based on $\tilde{Q}_{k}$:  $R_{i,t}(\tilde{Q}_{k})=Y_{i,t}+\gamma \operatorname*{max}_{a\in\mathcal{A}}\tilde{Q}_{k}(X_{i,t+1},\hat{\mathbf{V}}_{i,t+1,\kappa},a)$;
   \STATE Update $\tilde{Q}_{k}$ to $\tilde{Q}_{k+1}$:
   \begin{equation*}
   \begin{aligned}
   \tilde{Q}_{k+1}\leftarrow\operatorname*{argmin}_{f(.,a)\in\mathcal{F}_{SReLU}}\frac{1}{n}\sum_{(i,t)\in I_j}{[ R_{i,t}(\tilde{Q}_{k})-f(X_{i,t},\hat{\mathbf{V}}_{i,t,\kappa},A_{i,t}) ]^2}
   \end{aligned}
   \end{equation*}
   \ENDFOR
  \STATE \textbf{Return} The greedy policy: $\pi_K(a|x,v_{\kappa})=0$, if $a\notin argmax_{a'} \tilde{Q}_{K}(x,v_{\kappa},a'),\forall x,v_{\kappa},a$
\end{algorithmic}
\end{algorithm}

In Algorithm \ref{alg:algorithm1}, we fit $|\mathcal{A}|$ neural networks corresponding to each $a$ in $Q(s,a)$. This is reasonable in settings where the action space is small. 
When the dataset is small (such as the OhioT1DM dataset), we recommend to set $n$ to $N(T-1)$ such that all the data transactions (instead of a random subsample) will be used in each iteration. 

Similar to the original neural fitted Q-iteration algorithm in \citet{Riedmiller2005neuralFQI}, the intuition of this algorithm is also based on the Bellman optimality equation \eqref{Bellmanoptimal}. In each step $k$ of Algorithm \ref{alg:algorithm1}, $\tilde{Q}_{k}$ estimates $Q^{\pi^*}$  and response $R_{i,t}(\tilde{Q}_{k})=Y_{i,t}+\gamma \operatorname*{max}_{a\in\mathcal{A}}\tilde{Q}_{k}(X_{i,t+1},\hat{\mathbf{V}}_{i,t+1,\kappa},a)$ corresponds to the right hand side of equation \eqref{Bellmanoptimal}. Therefore, fitting the regression of $R_{i,t}$ with $\tilde{Q}_{k+1}$ is to solve the Bellman optimality equation. The key difference between Algorithm \ref{alg:algorithm1} and original neural fitted Q-iteration algorithm is that the high dimensional inputs  $Z_{i,t}^T=[Z_{i,t,1}^T,Z_{i,t,2}^T,\ldots,Z_{i,t,J}^T]$ is involved in the state space, and is mapped to a lower dimensional vector $\hat{\mathbf{V}}_{i,t,\kappa}$ during the learning process, so the neural networks $\tilde{Q}_0(x,v_{\kappa},a)$ takes principle components $\hat{\mathbf{V}}_{i,t,\kappa}$ rather than original high dimensional $Z_{i,t}$ as input.

\section{Asymptotic Properties}\label{s:asymp_property}
Before discussing the asymptotic properties of our proposed Q-learning method, we introduce some notations. 
\begin{definition}\label{def_1}

Define \begin{equation*}
\begin{aligned}
\mathcal{F}_{0}(L,\{d_i\}_{i=0}^{L+1},s,V_{max})=\{f:\mathcal{S}\times\mathcal{A}\rightarrow\mathbb{R},f(.,a)\in \mathcal{F}_{SReLU}(L,\{d_i\}_{i=0}^{L+1},s,V_{max}),\forall  a\in\mathcal{A}\},
\end{aligned}
\end{equation*} where $\mathcal{F}_{SReLU}(L,\{d_j\}_{i=0}^{L+1},s,V_{max})$ is the class of sparse ReLU network with $L$ layers and sparsity parameter $s$ and 
$V_{max}=\frac{R_{max}}{1-\gamma}$, the uniform upper bound for the cumulative reward. 
\end{definition}  We 
define $L_{\mathcal{F}_0}=\sup_{f\in\mathcal{F}_0}{\sup_{x\ne y}{\frac{{|f(y)-f(x)|}^{2}}{\left\Vert y-x\right\Vert^{2}}}}$ as the Lipschitz constant for the sparse ReLU class $\mathcal{F}_{SReLU}(L,\{d_j\}_{i=0}^{L+1},s,V_{max})$ used in $\mathcal{F}_{0}$. Note that this class $\mathcal{F}_{0}$ is used in the original neural fitted Q-iteration algorithm to model the Q-function, where the dimension of high frequency part $Z$ in state is not reduced through PCA.
We 
further define a function class $\mathcal{F}_{2}$ such that it models the Q-function by first converting high frequency part $Z$ into its principal component scores and then use a sparse ReLU neural network to obtain the resulting Q-function. More specifically, $\mathcal{F}_{2}$ is a set of functions $\{f_2\}$, such that $f_2(x,z,a) =  f_0(x,\hat{v}_{\kappa},a)$, where $f_0\in \mathcal{F}_0$ and $\hat{v}_{\kappa}$ is the vector containing first $\kappa$ principle components of $Z$. Note that $\mathcal{F}_{2}$ is the function class that we use in Algorithm \ref{alg:algorithm1} to model the Q-function. That is, $\tilde{Q}_k\in\mathcal{F}_{2},k\in\{1,2,...,K\}$ (formal definition of $\mathcal{F}_{2}$ and another function class $\mathcal{F}_{1}$ not mentioned here are in Section \ref{sec:def_two_func_class} of Appendix).

In addition, we introduce the H{\"o}lder smooth function class by $\mathcal{C}_r(\mathcal{D},\beta,H)$ with $\mathcal{D}\in\mathbb{R}^r$ to be the set of function input. The definition is: 
\begin{definition}\label{def_2}

Define \begin{equation*}
\begin{aligned}
    \mathcal{C}_r(\mathcal{D},\beta,H)=\{f:\mathcal{D}\rightarrow \mathbb{R}:\sum_{\gamma<\beta}{\left\Vert \partial^\gamma f\right\Vert_\infty} + \sum_{\alpha\in\mathbb{N}_0^{r}: \left\Vert\alpha\right\Vert=\lfloor \beta \rfloor}{\sup_{x,y\in\mathcal{D},x\ne y}{\frac{|\partial^{\alpha}f(x)-\partial^{\alpha}f(y)|}{\left\Vert x-y\right\Vert_\infty^{\beta-\lfloor \beta \rfloor}}}} \le H \},
\end{aligned}
\end{equation*} where $\alpha\in\mathbb{N}_0^{r}$ is a $r$-tuple multi-index for partial derivatives. 
\end{definition}

We next construct a network structure $\mathcal{G}(\{p_j,t_j,\beta_j,H_j\}_{j\in[q]})$ with the component function on each layer of this network belonging to Holder smooth function class $\mathcal{C(\mathcal{D},\beta,H)}$, which is called composition of Holder Smooth functions. This composition network contains $q$ layers, with each layer being $g_j:{[a_j,b_j]}^{p_j}\rightarrow {[a_{j+1},b_{j+1}]}^{p_{j+1}}$, such that $g_{jk}$ the $k$th component ($k\in[p_{j+1}]$) in layer $j$ satisfies that  $g_{jk}\in\mathcal{C}_{t_j}({[a_j,b_j]}^{t_j},\beta_j,H_j)$ with $t_j\le p_j$ inputs. Similar to the definition of $\mathcal{F}_{0}$, we can define the function class $\mathcal{G}_0(\{p_j,t_j,\beta_j,H_j\}_{j\in[q]})$(simply denoted as $\mathcal{G}_{0}$) on $\mathcal{S}\times\mathcal{A}\rightarrow\mathbb{R}$ such that each function $g\in\mathcal{G}_{0}$ satisfies that $g(.,a)\in\mathcal{G}(\{p_j,t_j,\beta_j,H_j\}_{j\in[q]})$ for $\forall a\in\mathcal{A}$. The relation between function class $\mathcal{G}_0$ and the network structure $\mathcal{G}$ is similar to the relation between function class $\mathcal{F}_0$ and the neural network $\mathcal{F}_{SReLU}$ in Definition \ref{def_1}.
See Definition 4.1 of \citet{Fan2020Qlearn} for more details on $\mathcal{G}_0(\{p_j,t_j,\beta_j,H_j\}_{j\in[q]})$. 

Next we will introduce the three major assumptions for our theorems: 
\begin{assumption}
\label{ass:a1}
The eigenvalues of $Cov(Z)$ follow an exponential decaying trend $\lambda_k=O(e^{-\zeta k}),k=1,2,...,m$ for some constant $\zeta>0$.
\end{assumption}
\begin{assumption}
\label{ass:a2}
The estimator $\hat{\mathbf{G} }=\sum_{k=1}^{m}{\hat{\lambda}_{k}\hat{U}_{k}{\hat{U}_{k}}^T}$  satisfies that $\left\Vert \hat{U}_{k}-U_{k}\right\Vert_2=O_p(n^{-\Delta})$ for $1\le k\le m$ such that $\Delta>0$ is some constant.
\end{assumption}
\begin{assumption}
\label{ass:a3}
First, we define the Bellman optimality operator $\mathcal{T}$ as \begin{equation*}
    \begin{aligned}
     \mathcal{T}f(x,z,a)=E\{Y_{0,t}+\gamma \max_{a'\in\mathcal{A}}f{(X_{0,t+1},Z_{0,t+1},a')}|A_{0,t}=a,X_{0,t}=x,Z_{0,t}=z\}.
    \end{aligned}
 \end{equation*} Then we assume $\mathcal{T}f\in \mathcal{G}_0$ for $f\in\mathcal{F}_{2}$.
 
\end{assumption}



Among the three assumptions, the exponential decaying structure of eigenvalues in Assumption \ref{ass:a1} can be commonly found in the literature of  high-dimensional and functional data analysis \citep[see e.g.,][]{reiss2020nonasymptotic, crambes2013asymptotics, jirak2016optimal}. 
This assumption is to control the information loss caused by using the first $\kappa$ principal component scores of $Z_{i,t}$ only. Assumption \ref{ass:a2} is about the consistency of the estimators $\hat{\mathbf{G}}$ and similar assumptions are imposed in the literature of functional data analysis \citep[see e.g.,][]{Staicu2018PCA, staicu2014likelihood}. Using similar arguments in proving Theorem 5.2 of \citet{Zhang2016consistentfuncional}, we can show that such an assumption holds in our setting as well. 
It is to bound the error caused by the estimation of the covariance matrix. 
Assumption \ref{ass:a3} 
is referred to as the completeness assumption in the literature \citep[see e.g.,][]{Chen2019completeness, uehara2022future, uehara2021finite}. 
This assumption is automatically satisfied 
when the transition kernel 
and the reward function 
satisfy certain smoothness conditions. 

Our first theorem is concerend with the nonasymptotic convergence rate of $\tilde{Q}_K$ in Algorithm \ref{alg:algorithm1}. 
\begin{theorem}[Convergence of Estimated Q-Function]\label{theorem_1}

Let $\mu$ be some distribution on $\mathcal{S}$ such that $\frac{d\mu(s)}{ds}$ bounded away from $0$. Under the Assumptions \ref{ass:a1} to \ref{ass:a3}, with sufficiently large $n$, there exists a sparse ReLU network structure for the function class $\mathcal{F}_{2}$ modeling $\tilde{Q}(s,a)$, such that $\tilde{Q}_K$ obtained from our Algorithm \ref{alg:algorithm1} satisfies that: 



\begin{equation*}
    \begin{aligned}
    \label{main_thm}
    \frac{1}{|\mathcal{A}|}\sum_{a\in\mathcal{A}}{\left\Vert Q^{\pi^*}(.,a)-\tilde{Q}_K(.,a)\right\Vert_{2,\mu}^2}=O_p(|\mathcal{A}|(n^{\alpha^*}\log^{\xi^*}{n}+d_1^*\kappa)n^{-1}\log^{\xi^*+1}{n}\\+L_{\mathcal{F}_0}(e^{-\zeta\kappa}-e^{-\zeta m}+n^{-2\Delta})+\frac{\gamma^{2K}}{(1-\gamma)^2}R_{max}^2),
    \end{aligned}
\end{equation*} 
where $\xi^{*}>1, 0<\alpha^*<1 $ are some constants, $|\mathcal{A}|$ is number of treatment options, and $d_1^*$ is the width of the first layer of the sparse ReLU network used in $\mathcal{F}_2$ satisfying the bound $m_0 + m \le d_1^* \le n^{\alpha^*}$. 
\end{theorem}

More details of neural network structure and sample size assumptions for Theorem \ref{theorem_1} can be found in Section \ref{sec: full_thm_1} of Appendix. Theorem \ref{theorem_1} provides an error bound on the estimated Q-function $\tilde{Q}_K$. Based on this theorem, we further 
establish the 
regret bound of the estimated policy $\pi_K$ obtained via Algorithm \ref{alg:algorithm1}. Toward that end, we need another assumption:
\begin{assumption}
\label{ass:a6}
Assume there exist $\eta>0,\delta_0 >0$, such that 
    \begin{equation*}
    \begin{aligned}
    P(s:\max_a{Q^{\pi^*}(s,a)}-\max_{a\in {\mathcal{A}-\argmax_{a'}{Q^{\pi^*}(x,a')}}}{Q^{\pi^*}(s,a)} \le \epsilon)=O(\epsilon^\eta)
    \end{aligned}
    \end{equation*} for $0<\epsilon\le \delta_0$. 
\end{assumption}
The margin type condition Assumption \ref{ass:a6} is commonly used in the literature. 
Specifically, in classification, the margin conditions are imposed to bound the excess risk \citep{Tsybakov2004classification,Tsybakov2007classification}. In dynamic treatment regime, a similar assumption is introduced for proving the convergence of State-Value function in a finite horizon setting \citep{qian2011qlearnguarantee,luedtke2016statistical}. In RL, these assumptions were introduced by \citep{Shi2020SAVE} to obtain sharper regret bound for the estimated optimal policy. 


\begin{theorem}[Convergence of State-Value Function]\label{theorem_2}
Under the Assumptions \ref{ass:a1} to \ref{ass:a6} and the conditions of $\mu,\mathcal{F}_2,n$ in Theorem \ref{theorem_1}, we have:

\begin{equation*}
    \begin{aligned}
    \label{second_thm}
    E_{\mu}[V^{\pi^*}(s)-V^{\pi_K}(s)]=O_p(\frac{1}{1-\gamma}\{ |\mathcal{A}|(n^{\alpha^*}{\log^{\xi^*}{n}}+d_1^*\kappa)n^{-1}{\log^{\xi^*+1}{n}}\\+L_{\mathcal{F}_0}(e^{-\zeta\kappa}-e^{-\zeta m}+n^{-2\Delta})+\frac{\gamma^{2K}}{(1-\gamma)^2}R_{max}^2\}^{\frac{\eta+1}{\eta+2}}).
    \end{aligned}
\end{equation*}

\end{theorem} 
The proofs of the two theorems are included in Section \ref{s:proof_of_thms} of Appendix. We 
summarize our theoretical findings below. 
First, we notice that the convergence rate of regret in Theorem \ref{theorem_2} is faster than the convergence of estimated Q-functions in Theorem \ref{theorem_1}. This is due to the margin type Assumption \ref{ass:a6} which enables us to obtain a sharper error bound. Similar results have been established in the literature. See e.g., Theorem 3.3 in \citet{Tsybakov2007classification}, Theorem 3.1 in \citet{qian2011qlearnguarantee}, Theorems 3 and 4 in \citet{Shi2020SAVE}. 

From Theorem \ref{theorem_2}, 
it can be seen that the regret bound is mainly determined by three parameters: the sample size $n$, the number of principal components $\kappa$ and the number of iterations $K$. 
Here, the first term $|\mathcal{A}|(n^{\alpha^*}{(\log^{\xi^*}{n})}+d_1^*\kappa)n^{-1}\log^{\xi^*+1}{n}$ 
on the right-hand-side corresponds to the estimation error, which decreases with $n$ and increases with $\kappa$. The second term $L_{\mathcal{F}_0}(e^{-\zeta\kappa}-e^{-\zeta m}+n^{-2\Delta})$ corresponds to the approximation error, which decreases with both $n$ and $\kappa$. The remaining term $\frac{\gamma^{K+1}}{(1-\gamma)^2}R_{max}$ is 
the optimization error which will decrease as the iteration number $K$ in Algorithm \ref{alg:algorithm1} grows. 

Compared with the existing results on the convergence rate of deep fitted Q-iteration algorithm \citep[see Theorem 4.4 of ][]{Fan2020Qlearn}, 
our theorems additionally characterize the dependence upon the number of principal components. 
Specifically, selecting the first $\kappa$ principal components induces the information loss (e.g., bias) that is of the order  $e^{-\zeta\kappa}-e^{-\zeta m}$, but reduces the 
model complexity caused by high frequency variables from $d_1^*m$ to $d_1^*\kappa$ and hence the variance of the policy estimator. 
This represents a bias-variance trade-off. Notice that the bias decays at an exponential order, when the training data is small, reducing the model complexity can be more beneficial. 
Thus, our algorithm is expected to perform better than the original fitted Q-iteration algorithm in small samples, 
as shown in our numerical studies.

Finally, 
The number of principal components shall diverge with $n$ to ensure the consistency of the proposed algorithm. 
Based on the two theorems, the optimal $\kappa^*$ that balances the bias and variance trade-off shall satisfy $\kappa^*\asymp \log(n)$ (details are given in Section \ref{s:appendix_kappa_select} of Appendix). 
Thus, when $n$ goes to infinity, we will eventually take $\kappa^* =m$ and our Algorithm \ref{alg:algorithm1} will be equivalent to the original neural fitted Q-iteration. This is just an asymptotic guideline 
for selecting the number of principal components. 
We provide some practical guidelines 
in the next section.


\section{Empirical Studies}\label{s:numerical}
\subsection{Practical Guidelines for Number of Principal Components}\label{s:choice_pca}
In 
functional data analysis, 
several criteria have been developed to select number of principal components, including the percentage of variance explained, Akaike information criterion (AIC) and Bayesian information criterion (BIC) \citep[see e.g.,][]{yao2005functional,li2013selecting}. In our setting, it is difficult to apply AIC/BIC, since there does not exist a natural objective function (e.g., likelihood) for Q-function estimation. One possibility is to extend the value information criterion \citep{shi2021concordance} developed in single-stage decision making to infinite horizons. Nonetheless, it remains unclear how to determine the penalty parameter for consistent tuning parameter selection. 



Here,  we select $\kappa$ based on 
the percentage of variance explained. 
That is, we can look at the minimum value of $\kappa$ such that the total variance explained by PCA reaches a certain level (e.g., $95\%$). This method is also employed in \citet{Staicu2018PCA} in single-stage decision making. 
To illustrate the empirical performance of this method, we apply Algorithm \ref{alg:algorithm1} with 
$\kappa\in\{2,6,10,14...,74\}$ and 
evaluate the expected return of these policies in the following numerical study. 

The data generating process can be described as follows. We set the low frequency covariate $X_{i,t}$ to be a $2$-dimensional vector and the high frequency variables $Z_{i,t}$ to be a $108$-dimensional vector ($m=108$). Both are sampled from mean zero normal distributions. The covariance matrix of $Z$ is set to 
satisfies Assumption \ref{ass:a1}. 
The action space is binary and the behavior policy to generate actions in training data is a uniform random policy. The reward function $r(x,z,a)$ is set to $=x\beta_{1,a}+z\beta_{2,a}+c \max{\{(x\beta_{1,a}+z\beta_{2,a}),0\}}$ for some constant $c$ and coefficient vectors $\beta_{1,a},\beta_{2,a}$ such that it is a mixture of a linear function and a neural network with a single layer and ReLU activation function. 
Next state $(X_{i,t+1},Z_{i,t+1})$ will be generated from a normal distribution with mean being a linear function of state $({X}_{i,t},{Z}_{i,t})$ and action ${A}_{i,t}$. 
The number of trajectories $N$ is fixed to $6$ and the length of trajectory $T$ is set to be $80$. 

The ReLU network is constructed with 3 hidden layers and width $d_1=15, d_2=d_3=5$. Dropout layers with $10\%$  dropout rate are added between layer 1, layer 2 and layer 3. During training, the dropout layers randomly sets the output from previous layers to $0$ with the probability $10\%$, which can introduce sparsity to the neural network and reduce over-fitting \citep{Srivastava2014dropout}. 
The hyperparamters of neural network structure can be tuned via cross-validation. 
The discounted factor $\gamma$ is fixed to $0.5$. 


To evaluate the policy performance, we can use a Monte-Carlo method to approximate the 
expected return under each estimated policy. Specifically, for each estimated policy $\pi$, we generate $N_{mc} = 100$ trajectories each of length $T_{mc} = 20$ (in our setting with $\gamma=0.5$, the cumulative reward after $T_{mc} = 20$ is negligible). The initial state 
distribution is the same as the one in the training dataset. The actions 
are assigned according to $\pi$. The expected return can then be approximated via the average of the empirical cumulative rewards over the 100 trajectories. 

\begin{figure}[h!]
  
  \centering
  \includegraphics[scale=0.4]{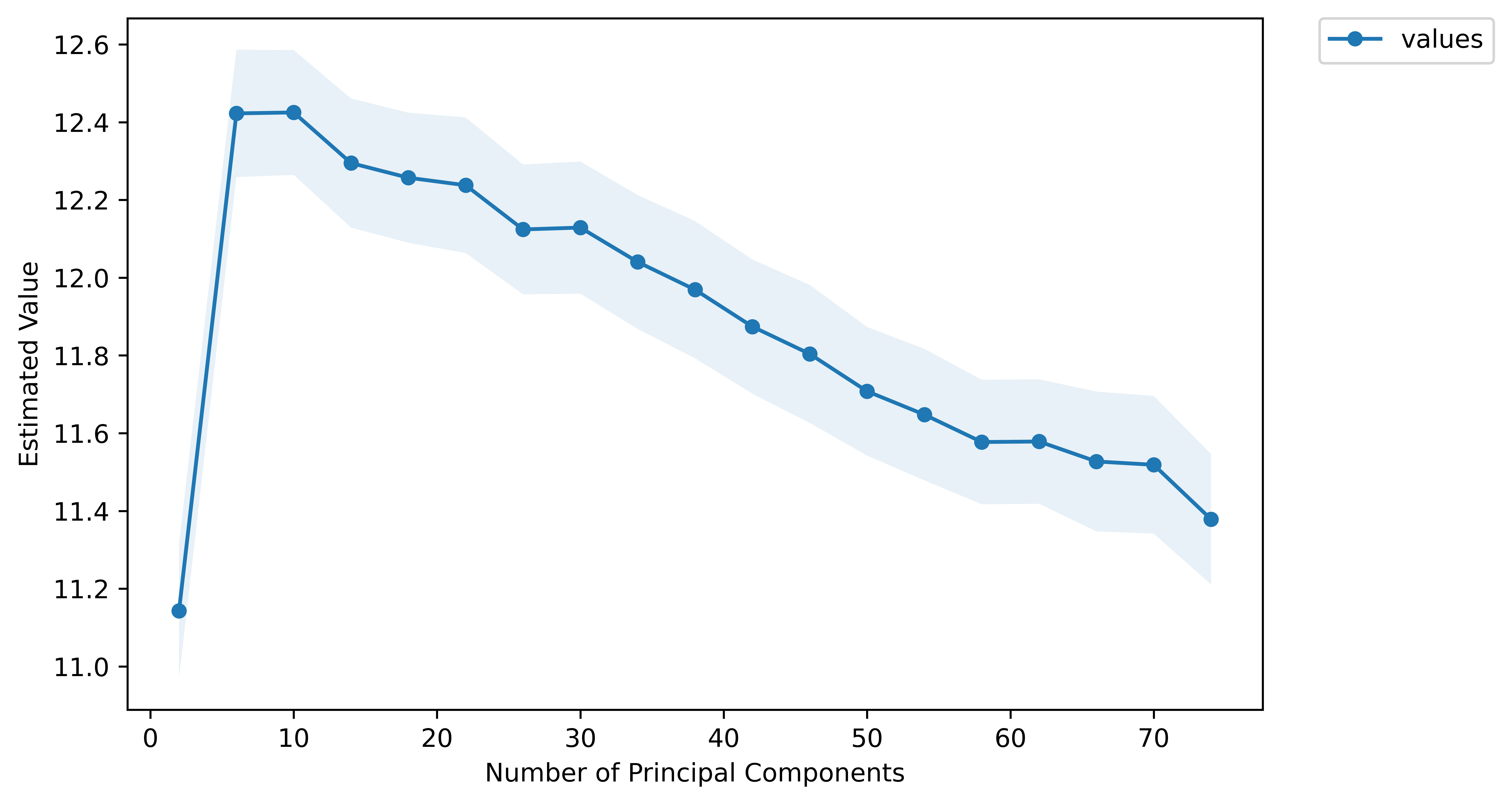}
  \includegraphics[scale=0.4]{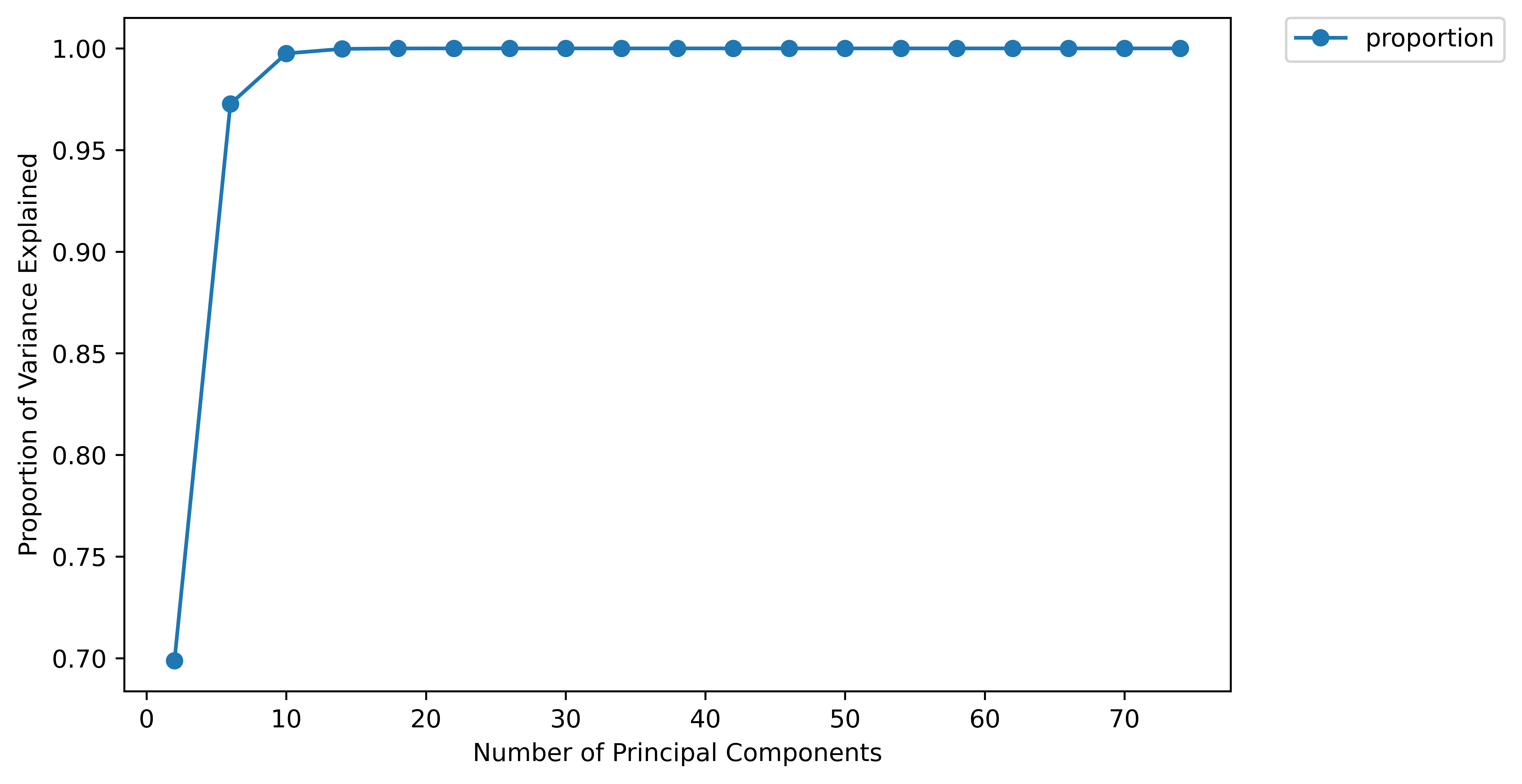}
  \caption{\textbf{Left}: Estimated value of policies by Algorithm \ref{alg:algorithm1} when $\kappa$ varies in $\{2,6,10,14...,74\}$ (shaded area is $95\%$ confidence interval); \textbf{Right}: Proportion of variance explained by the first $\kappa$ principal components}
  \label{f:figure_3}
\end{figure}

For each $\kappa$ in the list $\{2,6,10,14...,74\}$, we apply Algorithm \ref{alg:algorithm1} to learn the optimal policy over $80$ random seeds and evaluate their expected return using the Monte Carlo method. We then take the sample average and standard error of these $80$ expected returns to estimate the value of policy and construct the margin of error. Figure \ref{f:figure_3} depicts the estimated values of these expected returns as well as their confidence intervals. 
It can be seen that 
increasing $\kappa$ from $2$ to $6$ leads to a significant improvement. However, further increasing $\kappa$ worsens the performance. This trend is consistent with our theory since the bias term will dominate the estimation error for small value of $\kappa$. When $\kappa$ increases, the bias decays at exponentially fast and the model complexity term becomes the leading term. Meanwhile, the percentage of variance explained increases quickly when $\kappa\le 6$ and remains stable afterwards. As such, it makes sense to use this criterion for $\kappa$ selection. In our implementation, we select the smallest $\kappa$ such that the variance explained is at least $95\%$. 


\subsection{Simulation Study}\label{s:simulation_study}
In the simulation study, we 
compare the proposed policy $\pi_K^{PCA}$ against two baseline policies obtained by directly using 
the original high frequency variable $Z_{i,t,j}$ (denoted by $\pi_K^{ALL}$) and its average 
as input (denoted by $\pi_K^{AVE}$). 
Both policies are computed in a similar manner based on the deep fitted Q-iteration algorithm. 
We additionally include one more baseline policy, 
denoted by $\pi_K^{BOTTLE}$, which adds one bottleneck layer after the input of original $Z$ in the neural network architecture such that the width of this bottleneck layer is the same as the input dimension of the proposed policy $\pi_K^{PCA}$. 
This policy 
differs from the proposed policy in that it uses this bottleneck layer for dimension reduction instead of PCA. 
Both the data generating setting and the neural network structure are the same 
to Section \ref{s:choice_pca}. 
However, in this simulation study we vary the sample size 
and the dimension of the 
high frequency variables. Specifically, we consider 9 cases of training size ($N=6, T=60, 75, 90, 105, 120$ and $N=7,8,9,10, T=120$ ) and $5$ different high frequency part dimension $m = 27, 54, 108, 162, 216$. Furthermore, we have two settings of generating high frequency variables that will be discussed below. We similarly compare the proposed policy against 
$\pi_K^{ALL}$, $\pi_K^{BOTTLE}$, and $\pi_K^{AVE}$ and use the Monte Carlo method to evaluate their values. 

\begin{figure}[h!]
  
  \centering
  \includegraphics[scale=0.6]{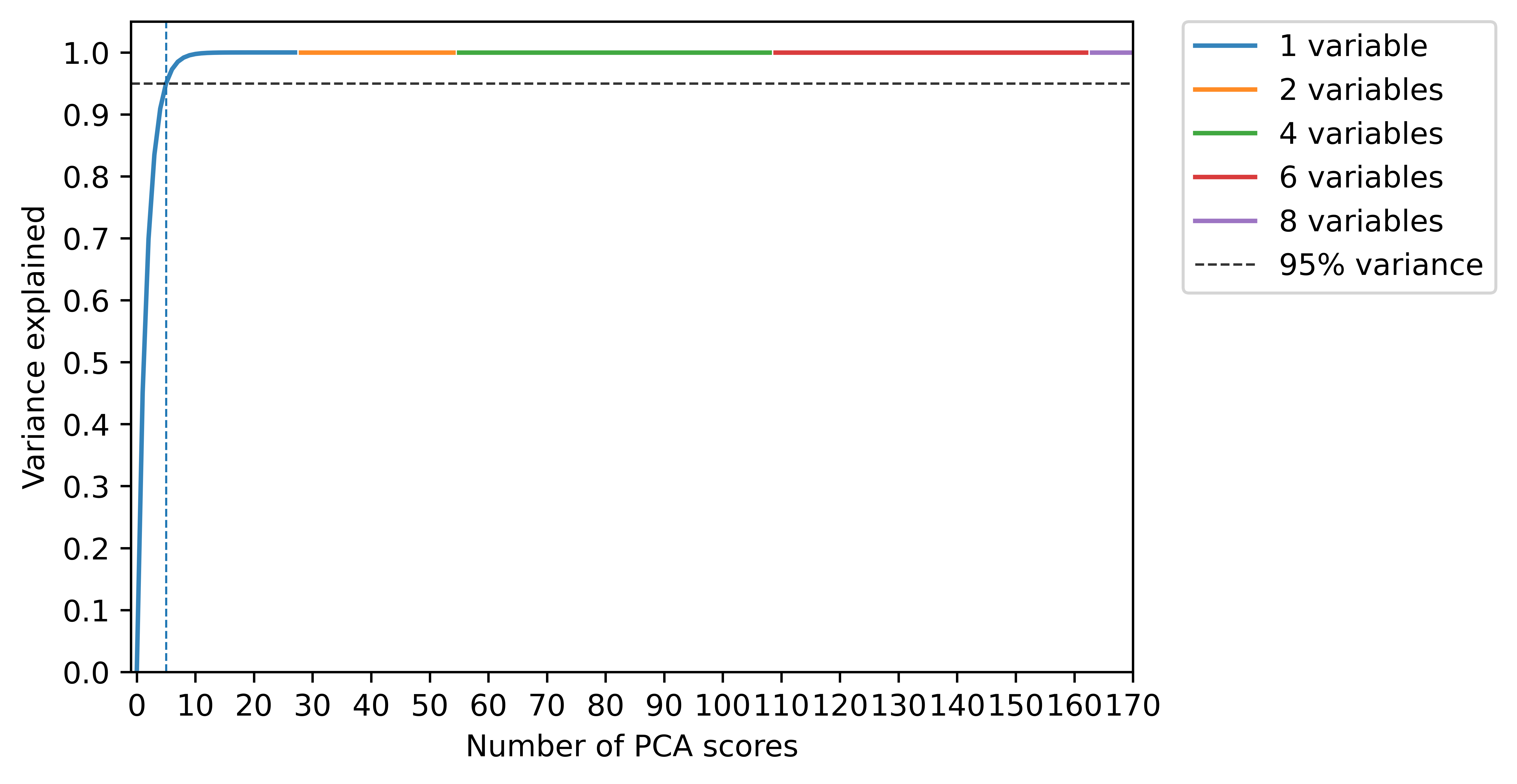}
  \includegraphics[scale=0.6]{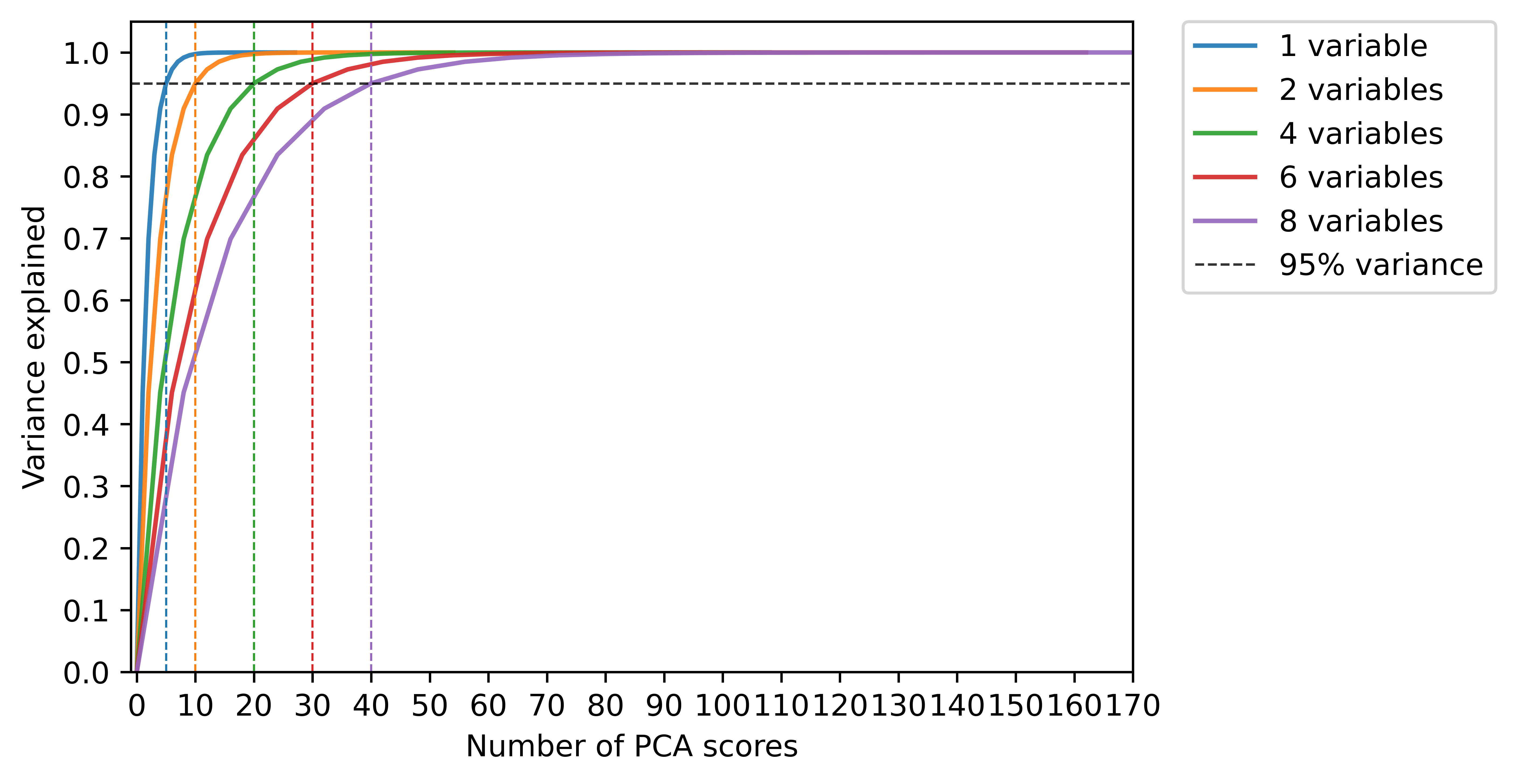}  
  \caption{\textbf{Upper}: Variance explained by first $\kappa$ principal component scores when there are $J= 1,2,4,6,8$ high frequency variables in the \textbf{first setting} of simulation (corresponding to Figure \ref{f:figure_2}). Horizontal dash line is $95\%$ of variance explained. $\kappa=5$ can explain $95\%$ variance in all $J= 1,2,4,6,8$ cases of the first setting; \textbf{Lower}: Variance explained by first $\kappa$ principal component scores when there are $J= 1,2,4,6,8$ high frequency variables in the \textbf{second setting }of simulation (corresponding to Figure \ref{f:figure_2_2}). Here the number of principal components are $\kappa=5,10,20,30,40$ corresponding to the five cases $J=1,2,4,6,8$ respectively to ensure $95\%$ of variance explained. Here eigenvalues of the concatenated high frequency $Z$ decays at an exponential order $\lambda_k=O(e^{-\zeta k})$ with $\zeta = 0.6$.} 
  \label{f:figure_2_pca_num}
\end{figure}

\begin{figure}[h!]
  \centering
     \includegraphics[width=0.42\textwidth]{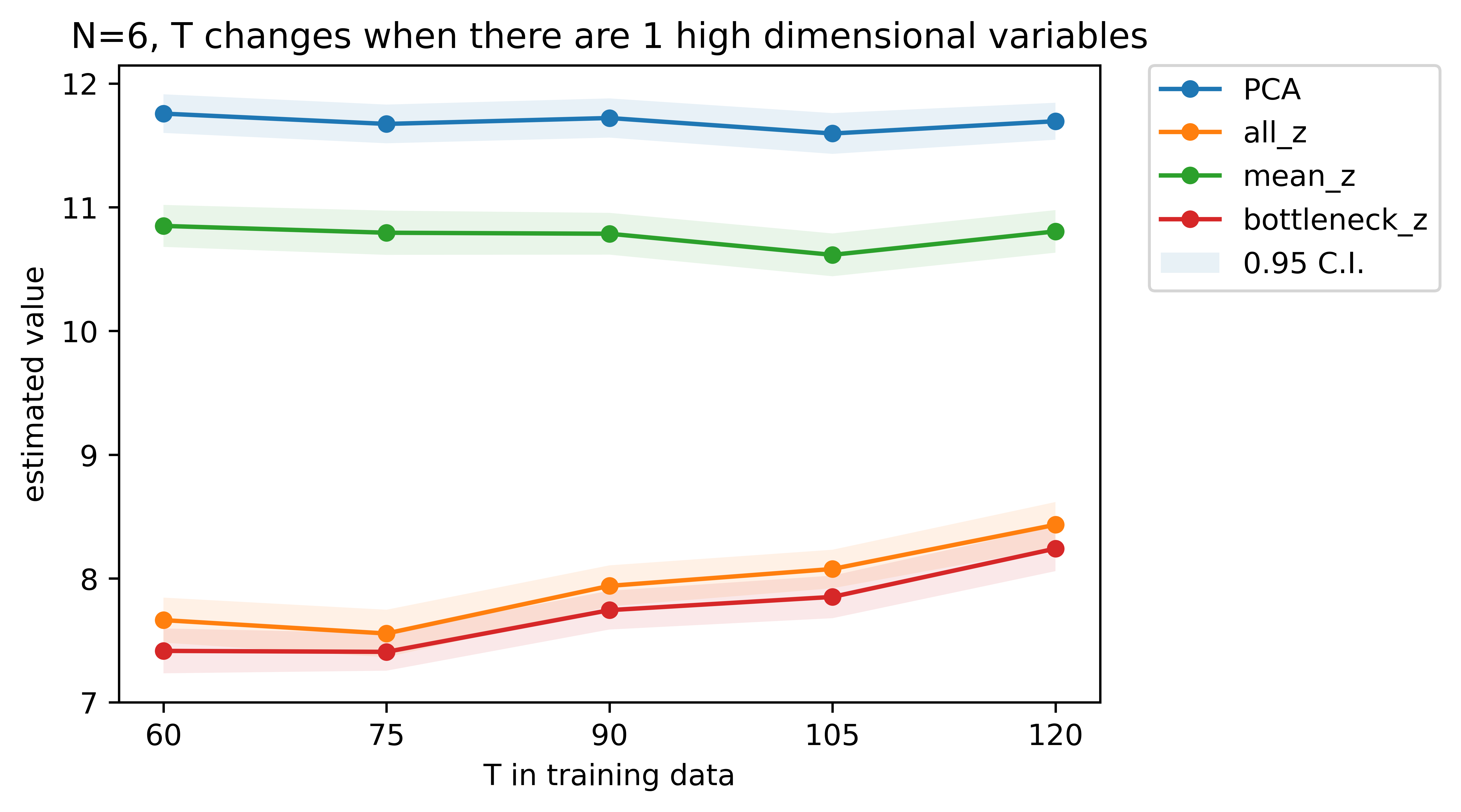}
     \includegraphics[width=0.42\textwidth]{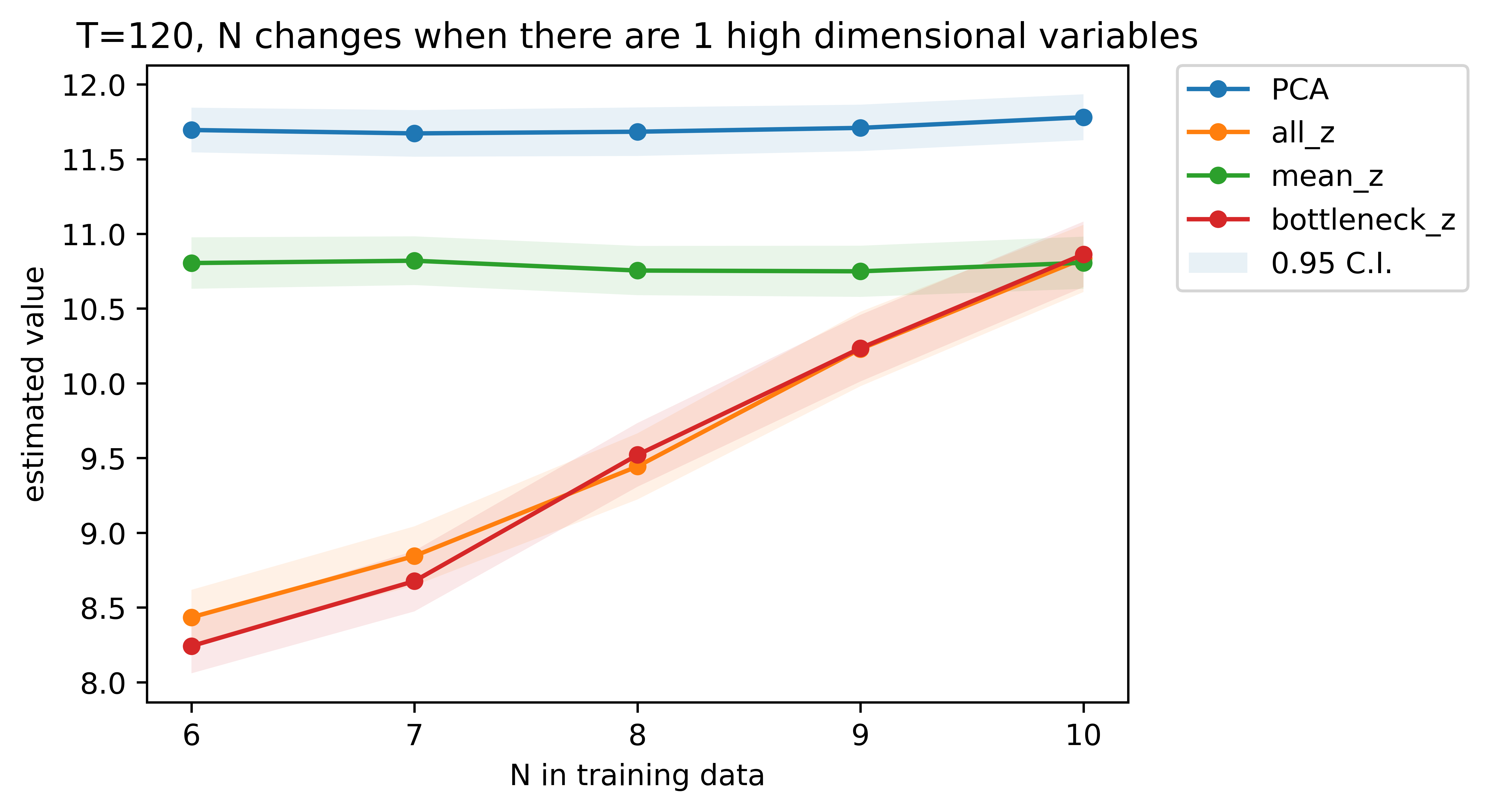}
      \includegraphics[width=0.42\textwidth]{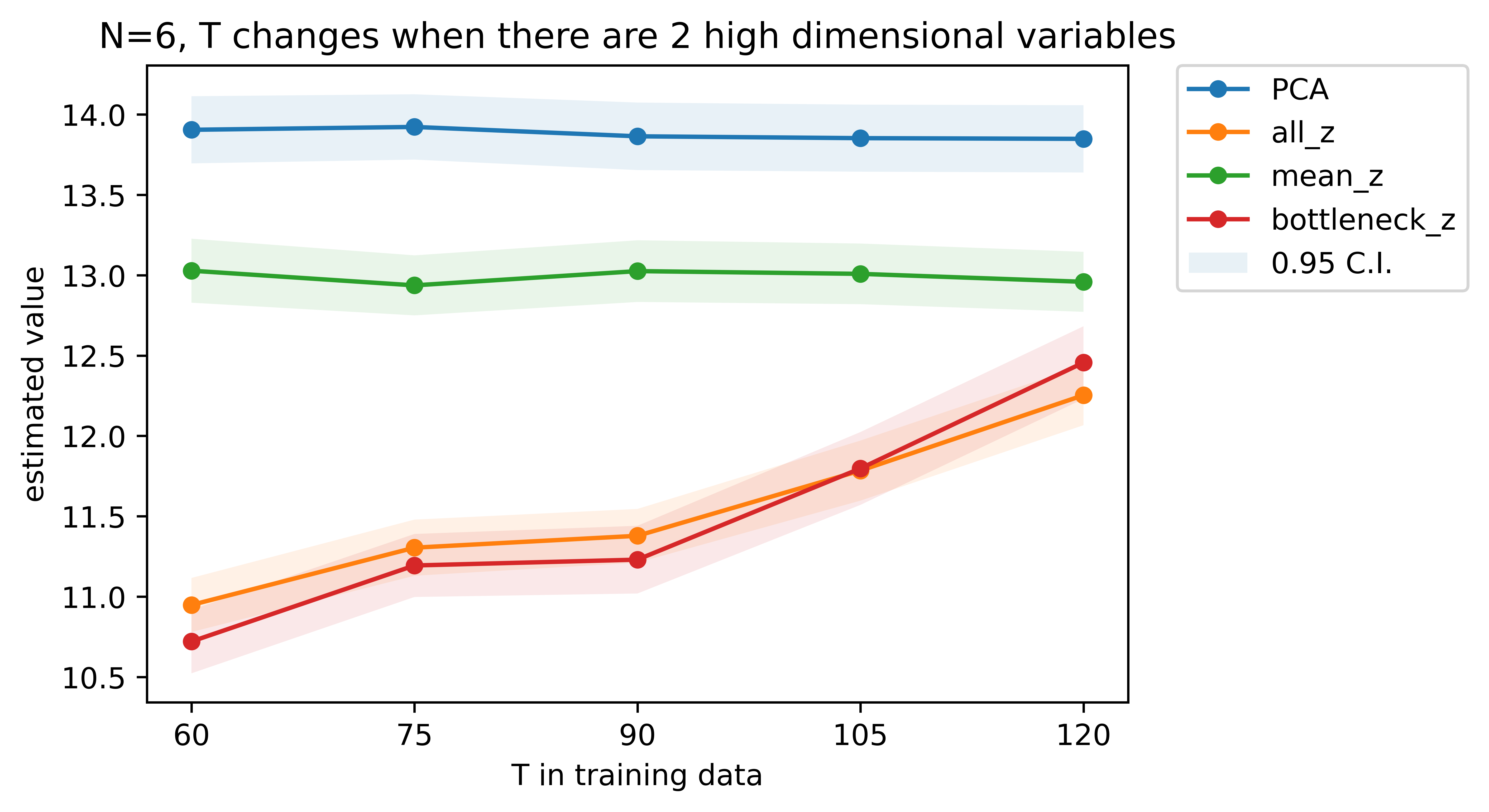}
    \includegraphics[width=0.42\textwidth]{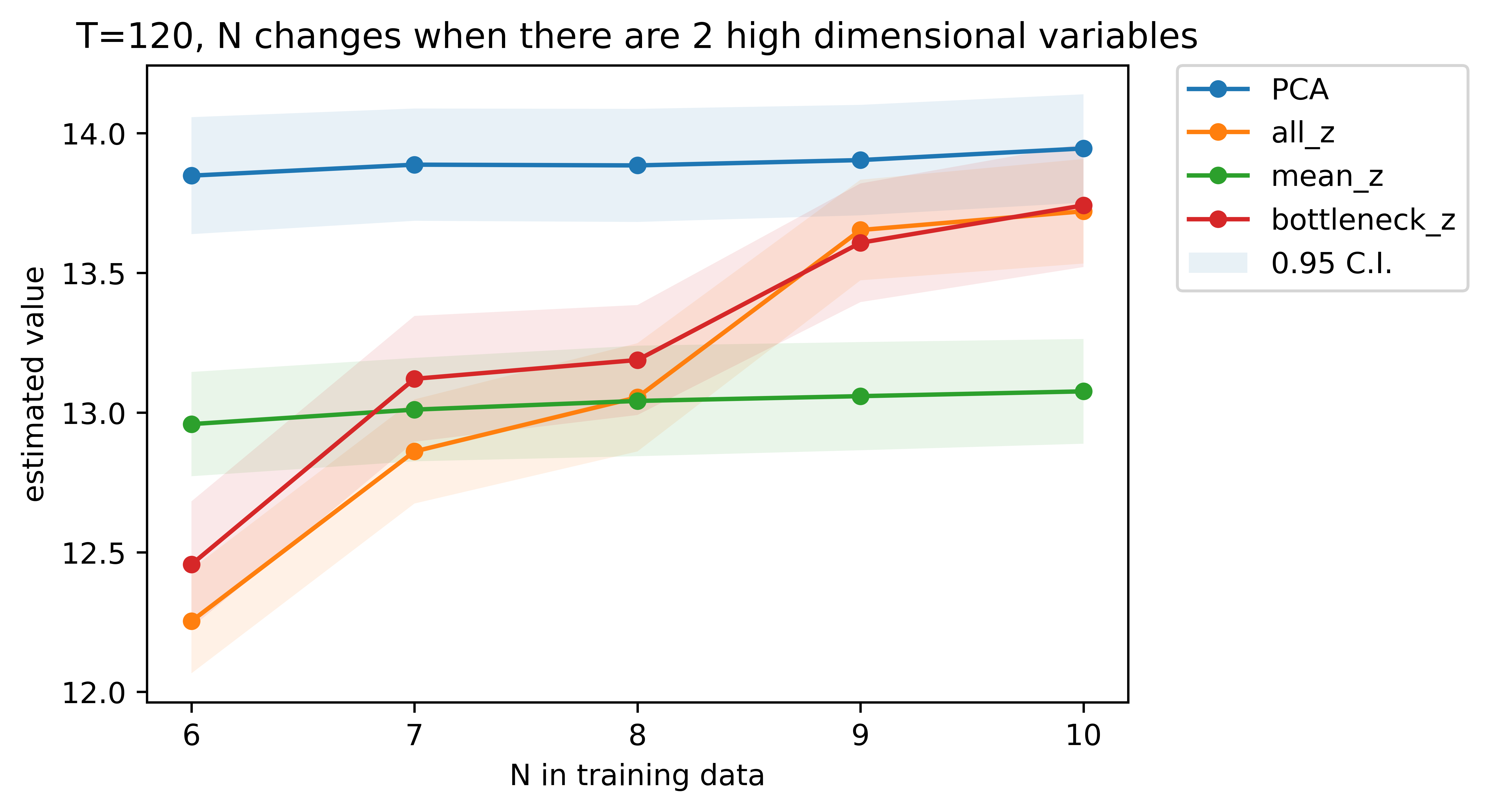}
      \includegraphics[width=0.42\textwidth]{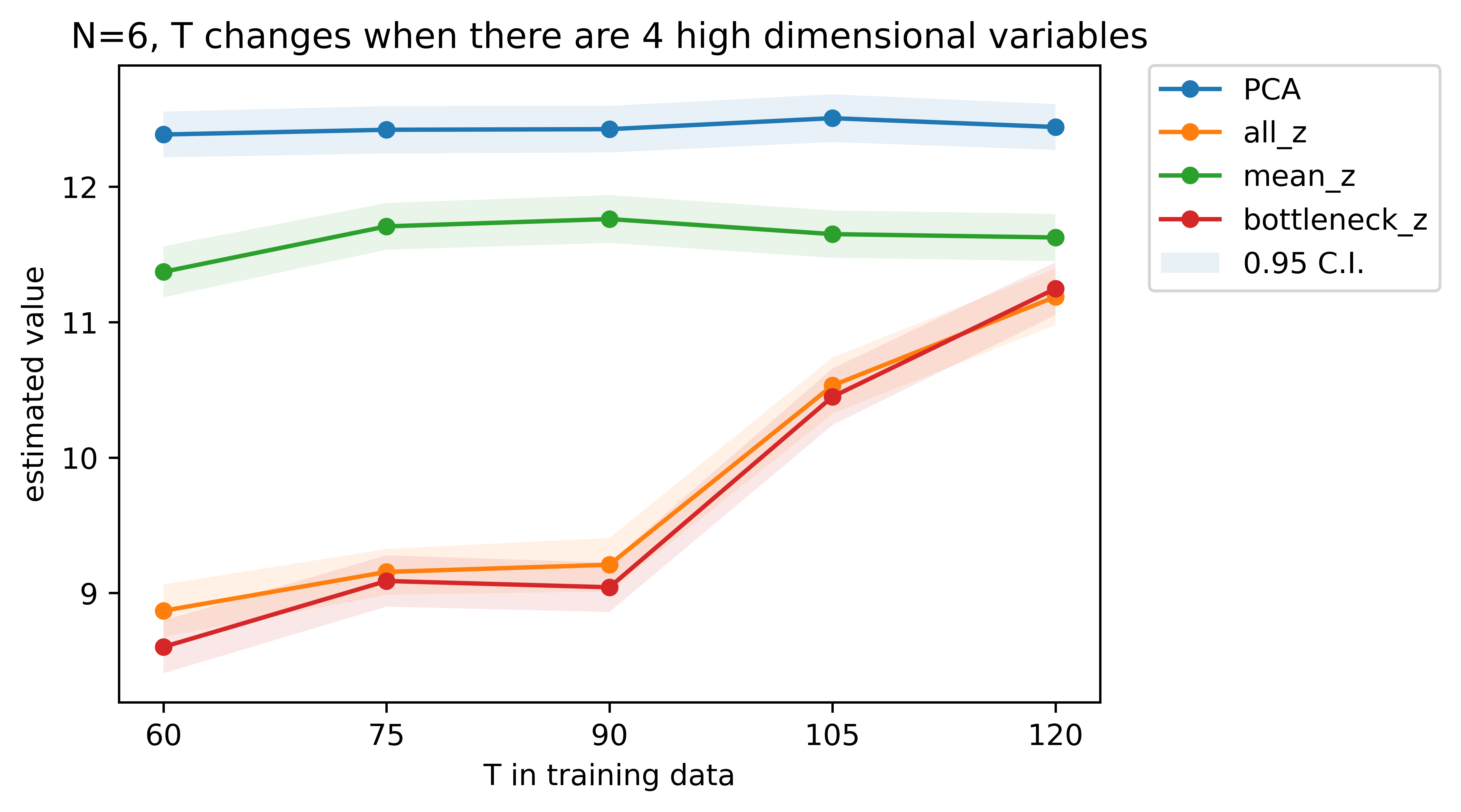}
      \includegraphics[width=0.42\textwidth]{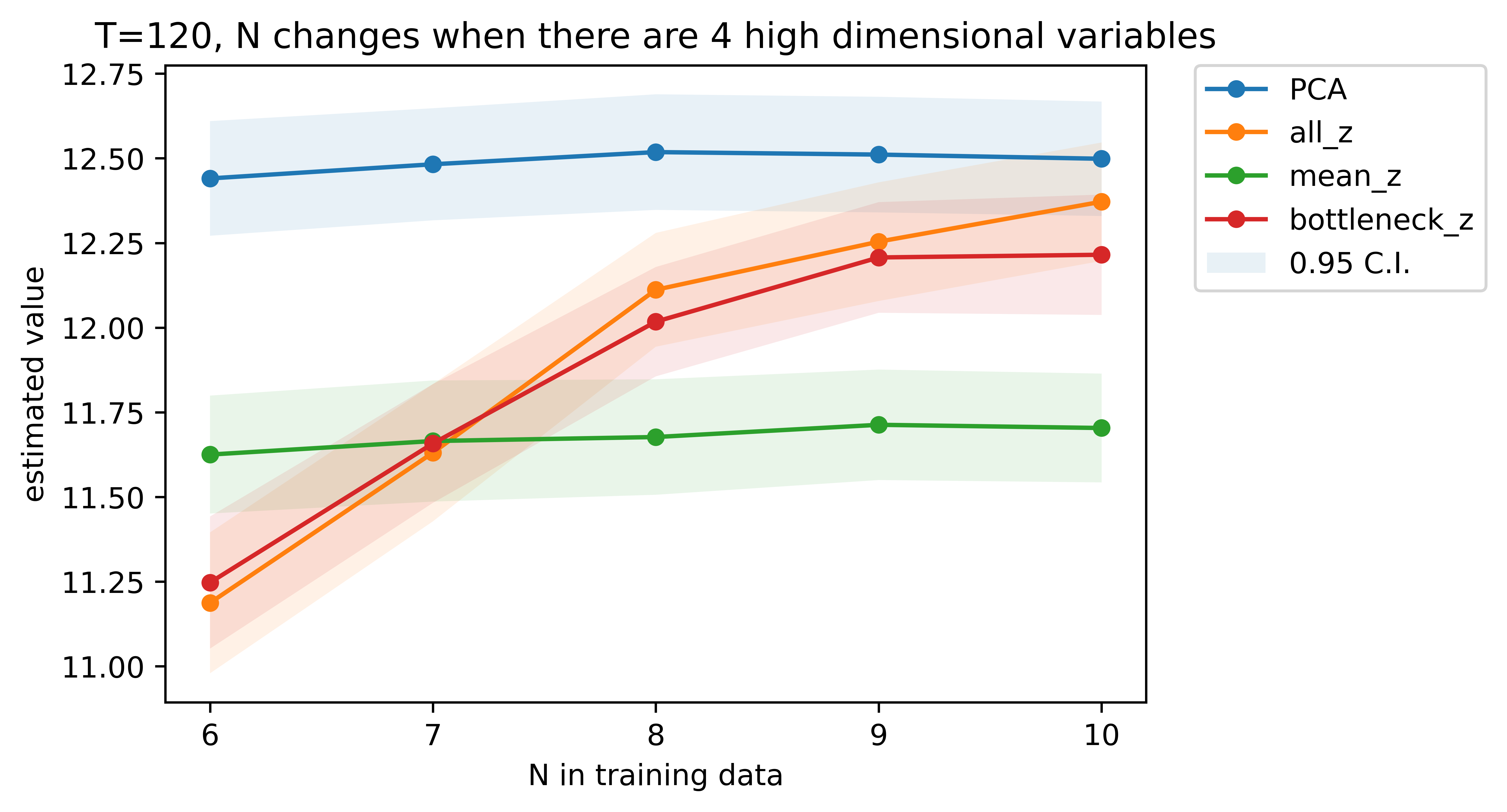}
  \includegraphics[width=0.42\textwidth]{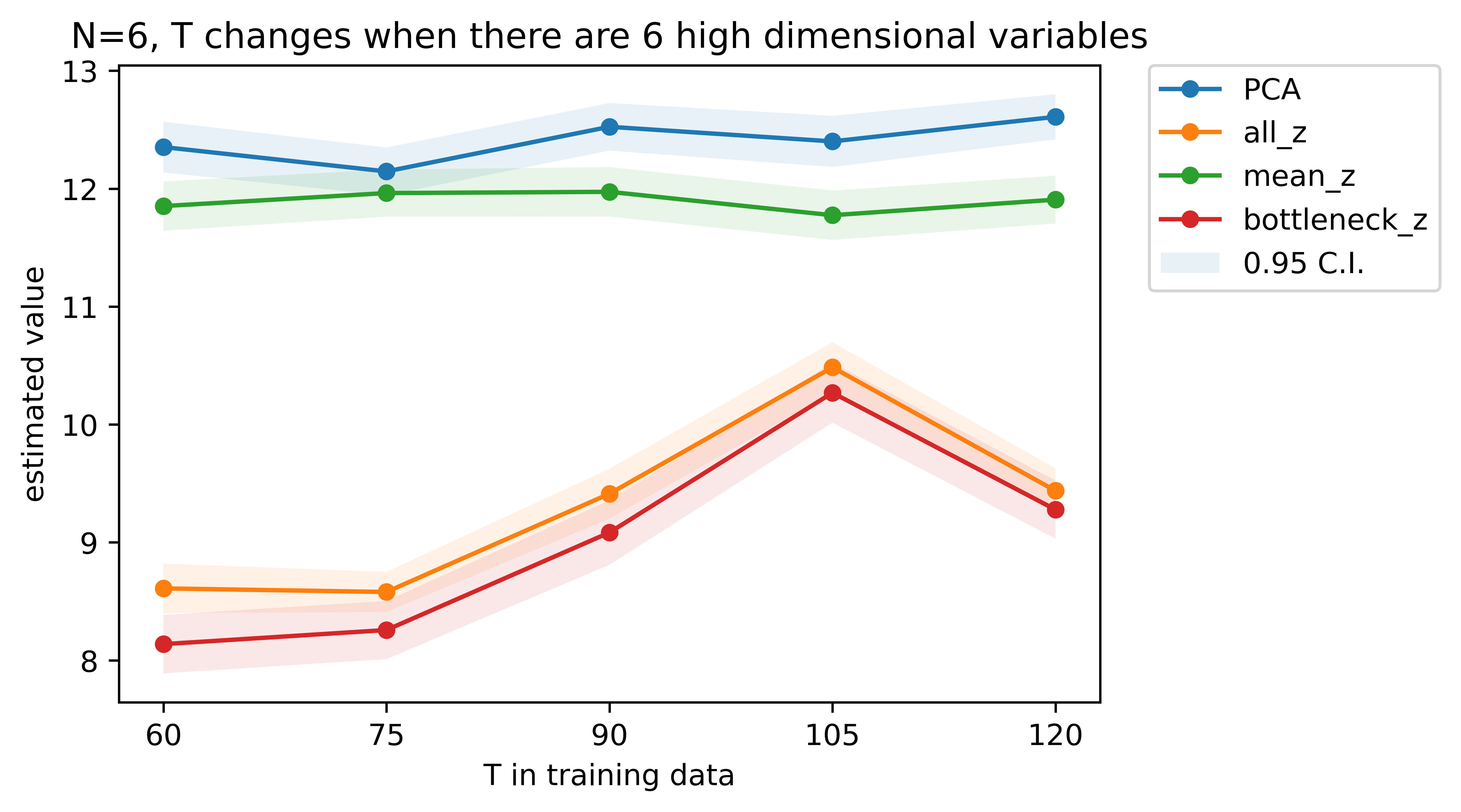}
  \includegraphics[width=0.42\textwidth]{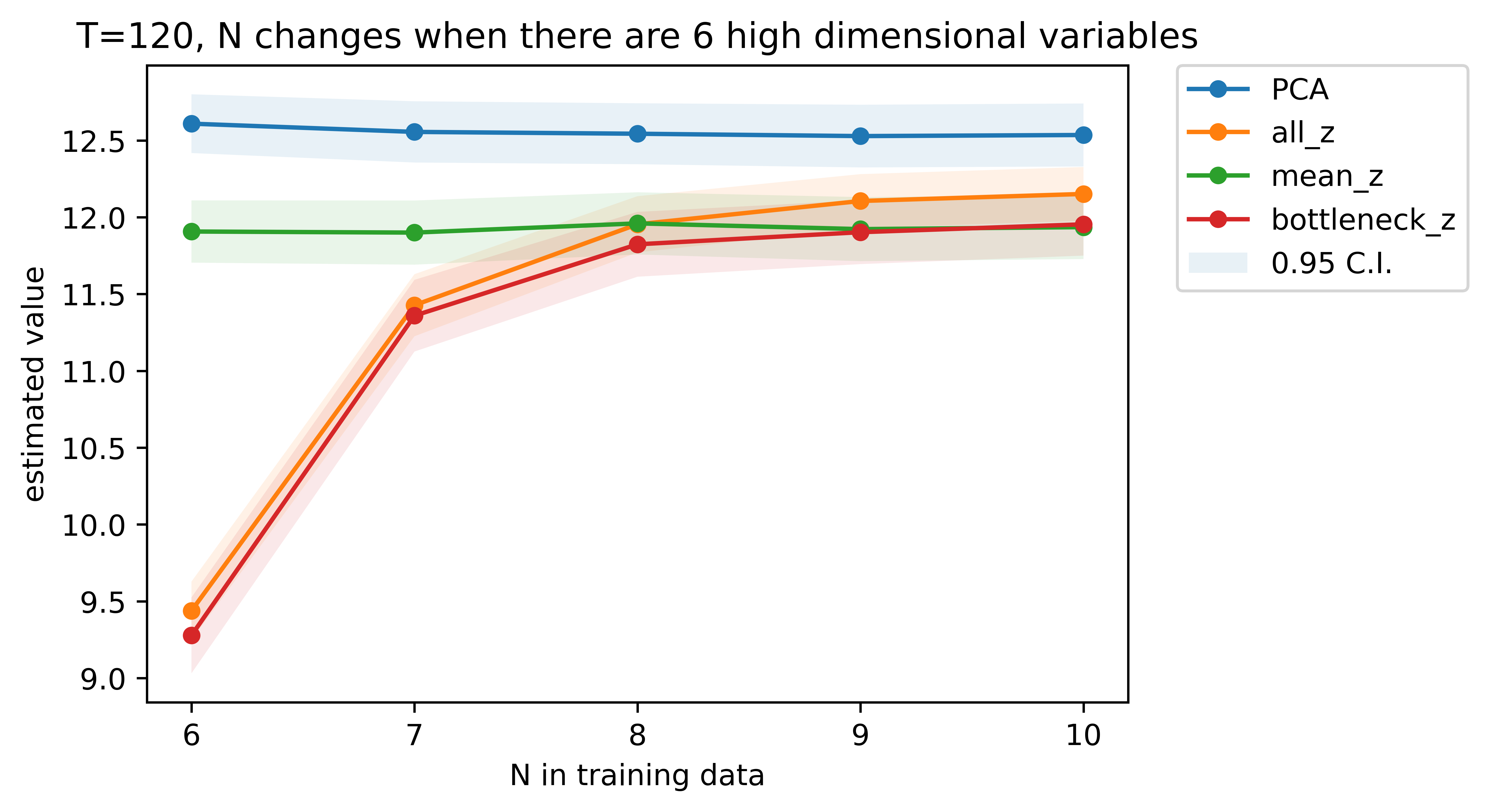}
  \includegraphics[width=0.42\textwidth]{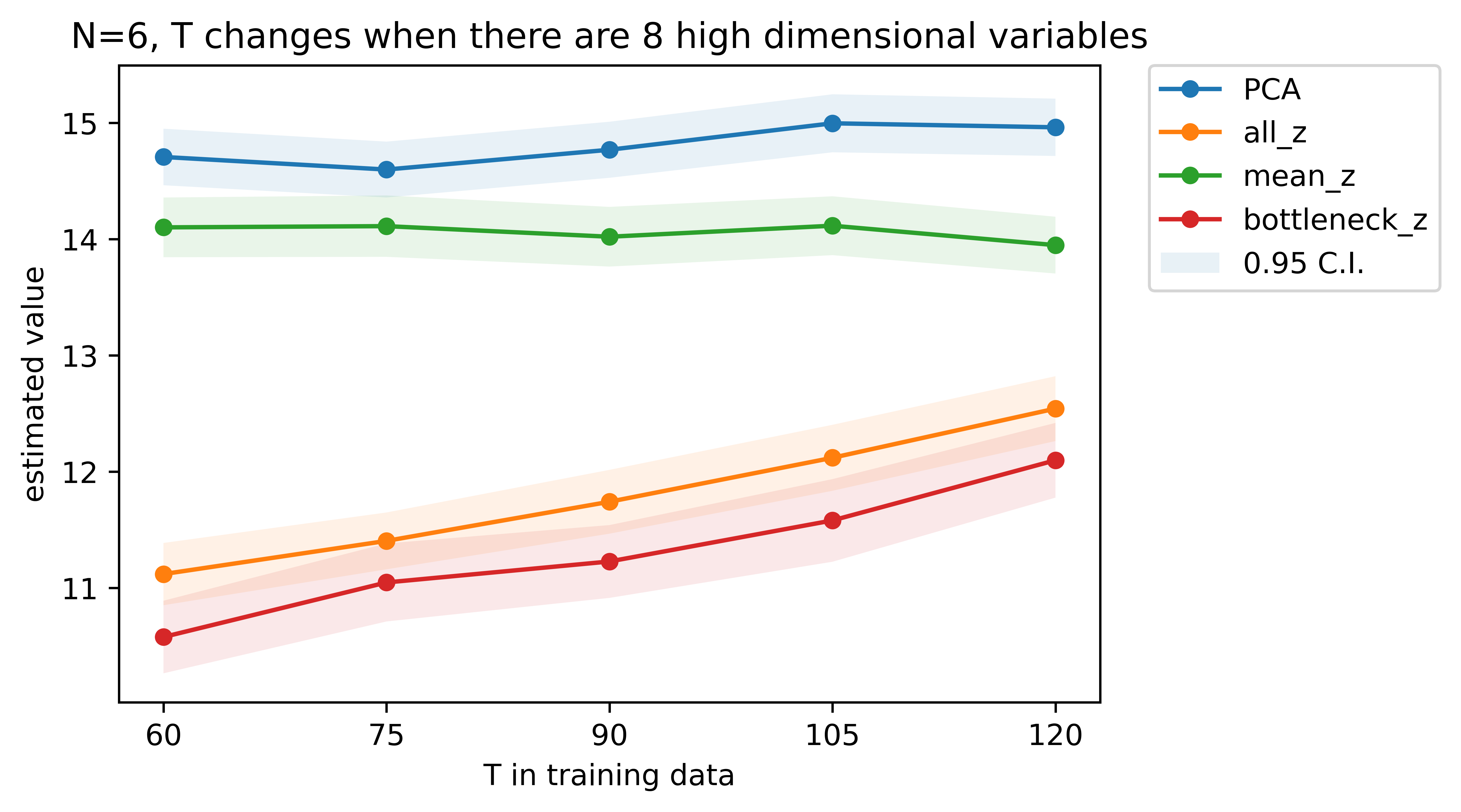}
  \includegraphics[width=0.42\textwidth]{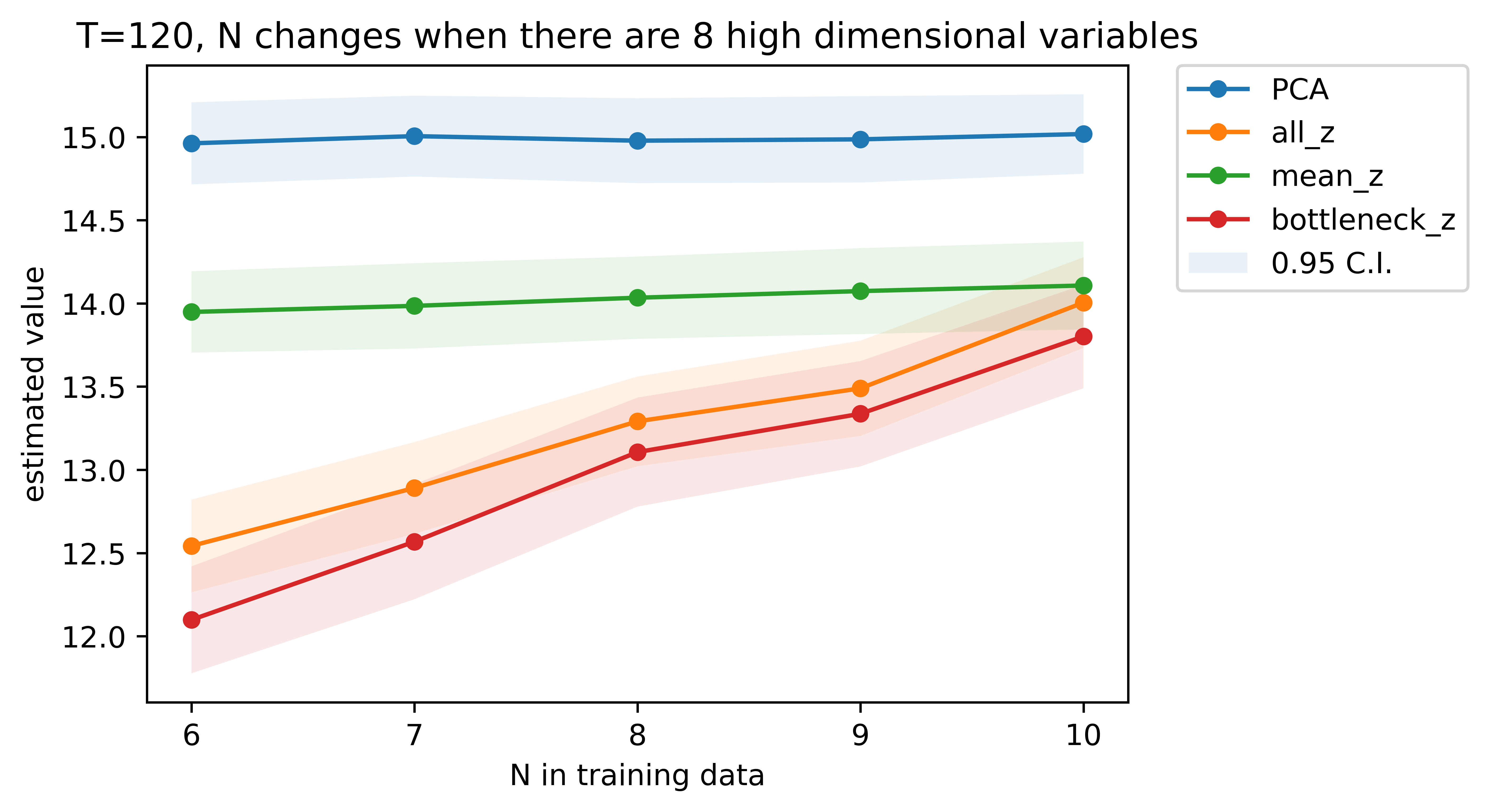}
  
  \caption{\textbf{First setting} in simulation. \textbf{Left}: training data with $N=6$ and $T=60,75,90,105,120$ ; \textbf{Right}: training data with $T=120$ and $N=6,7,8,9,10$ when there are $1,2,4,6,8$ variables with dimension $27$ (shaded area is $95\%$ confidence interval). In the legend, ``PCA'' refers to  $\pi_K^{PCA}$; ``all z'' refers to $\pi_K^{ALL}$; ``bottleneck z'' refers to  $\pi_K^{BOTTLE}$; ``mean z''  refers to  $\pi_K^{AVE}$.} 
  \label{f:figure_2}
\end{figure}

In the first setting, we consider $5$ cases with $J$ (the number of high frequency variables) equals $1,2,4,6,8$  and each high frequency variable $Z_j,j\in\{1,2,...,J\}$ is of dimension $27$. In this setting, 
all $J$ high frequency variables are dependent and eigenvalues of concatenated high frequency variable $Z$ decays at an exponential order. 
We find that the first $5$ principal components explains over $95\%$ of variance in all the $5$ cases, 
as shown in Figure \ref{f:figure_2_pca_num}. Therefore, we set the number of principal components $\kappa=5$ 
and plot the results in Figure \ref{f:figure_2}.

\begin{figure}[h!]
  \centering
  \includegraphics[scale=0.42]{APA/plots/S1_S2_J1_T_change.png}
  \includegraphics[scale=0.42]{APA/plots/S1_S2_J1_N_change.png}
  \includegraphics[scale=0.42]{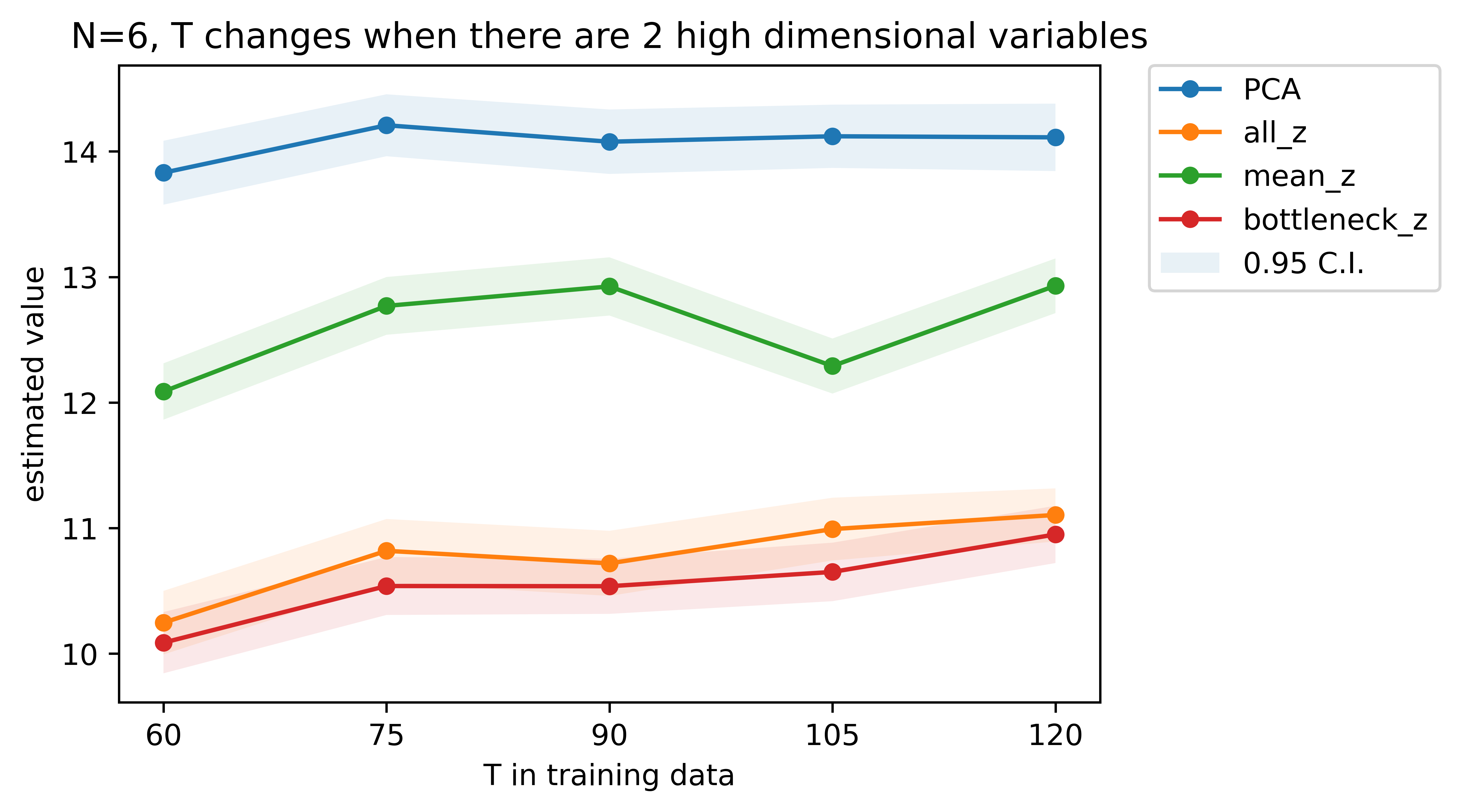}
  \includegraphics[scale=0.42]{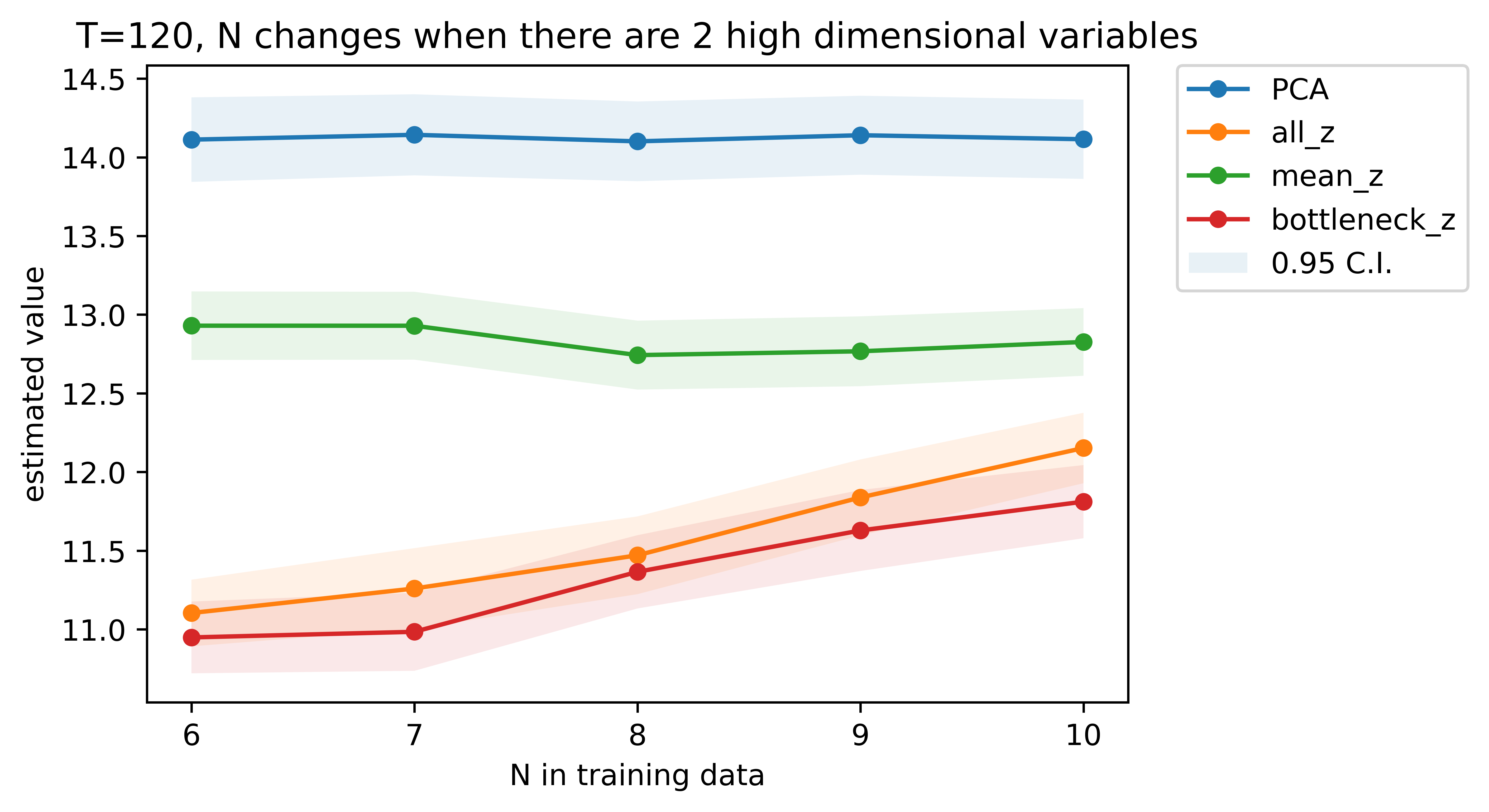}
  \includegraphics[scale=0.42]{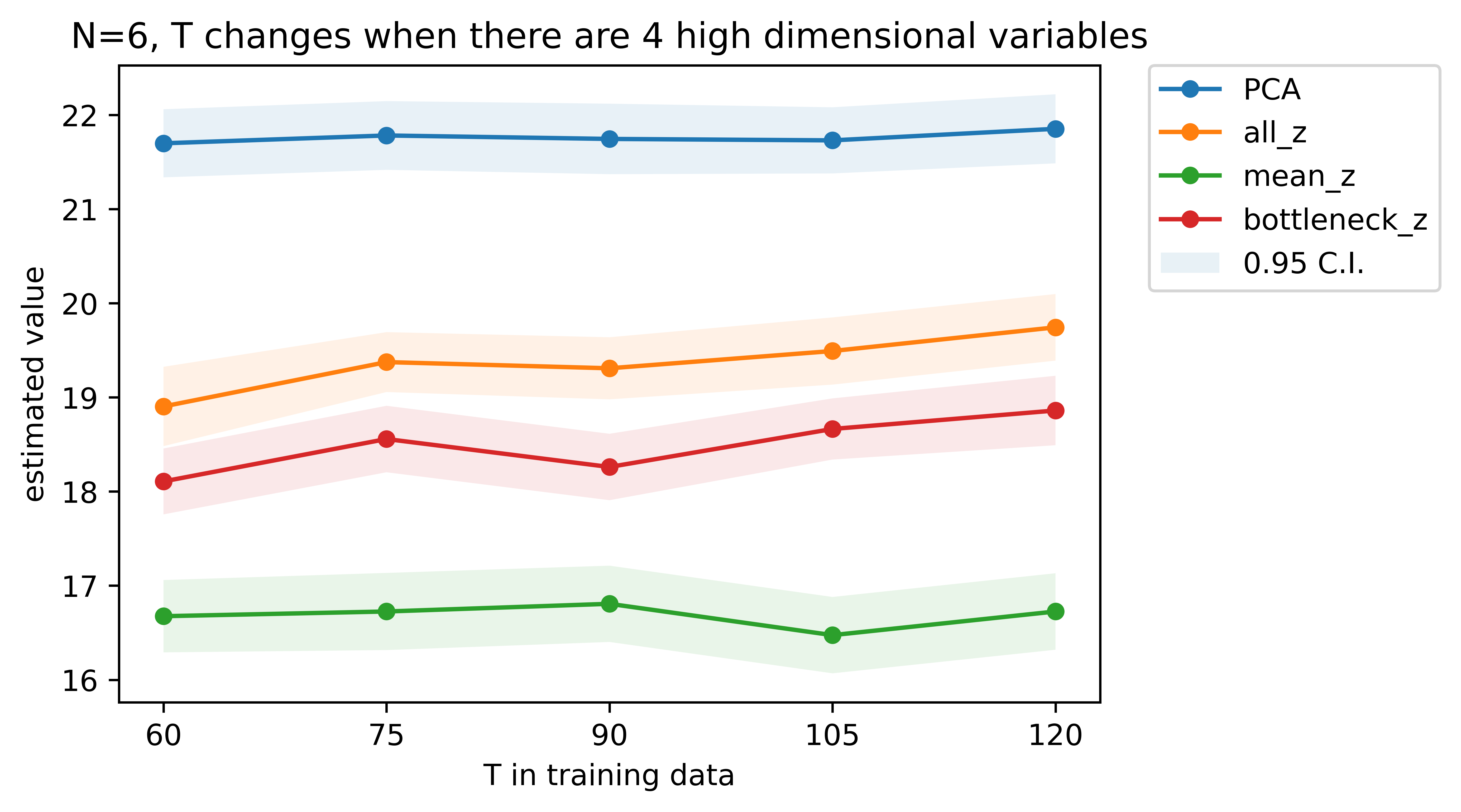}
  \includegraphics[scale=0.42]{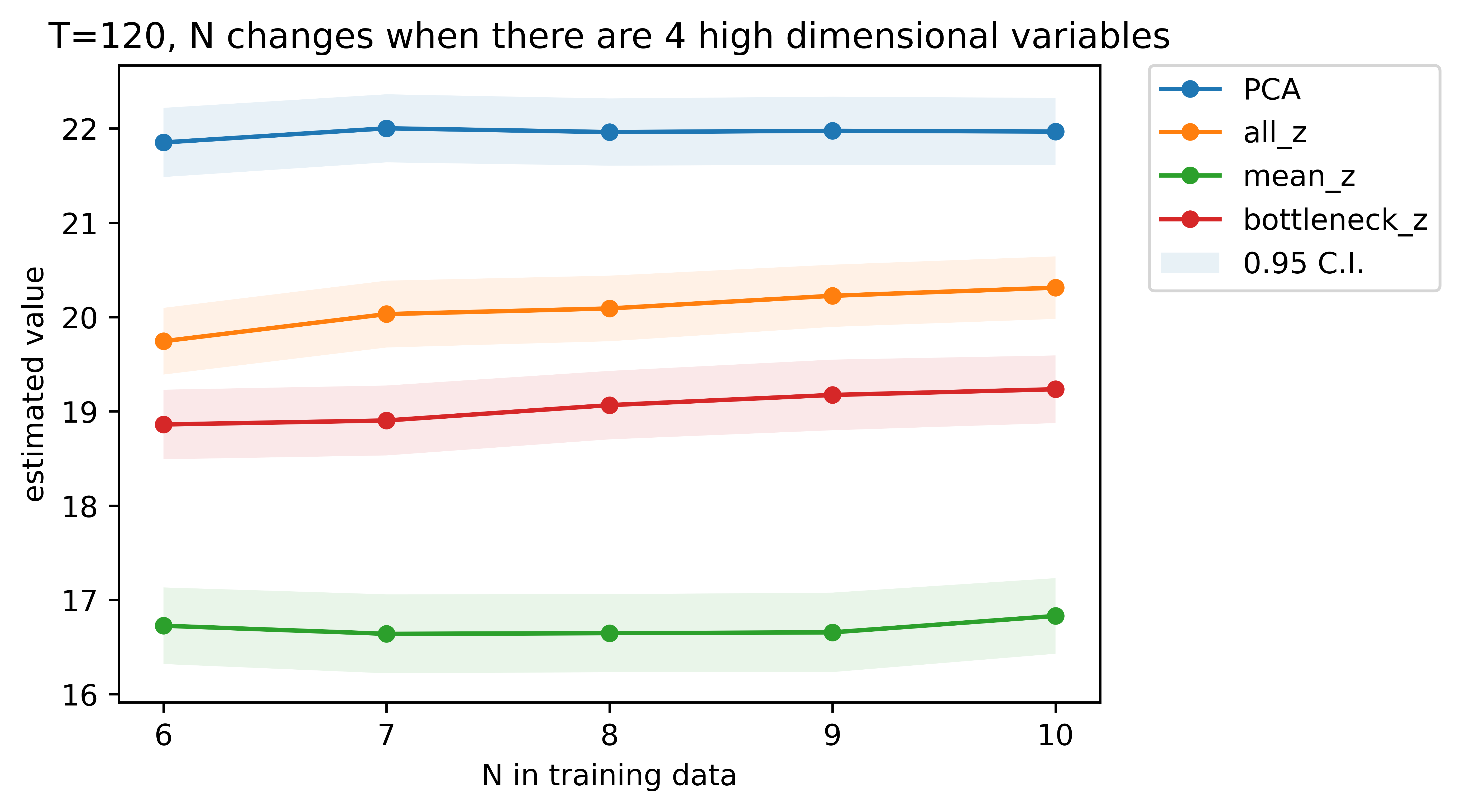}
    \includegraphics[scale=0.42]{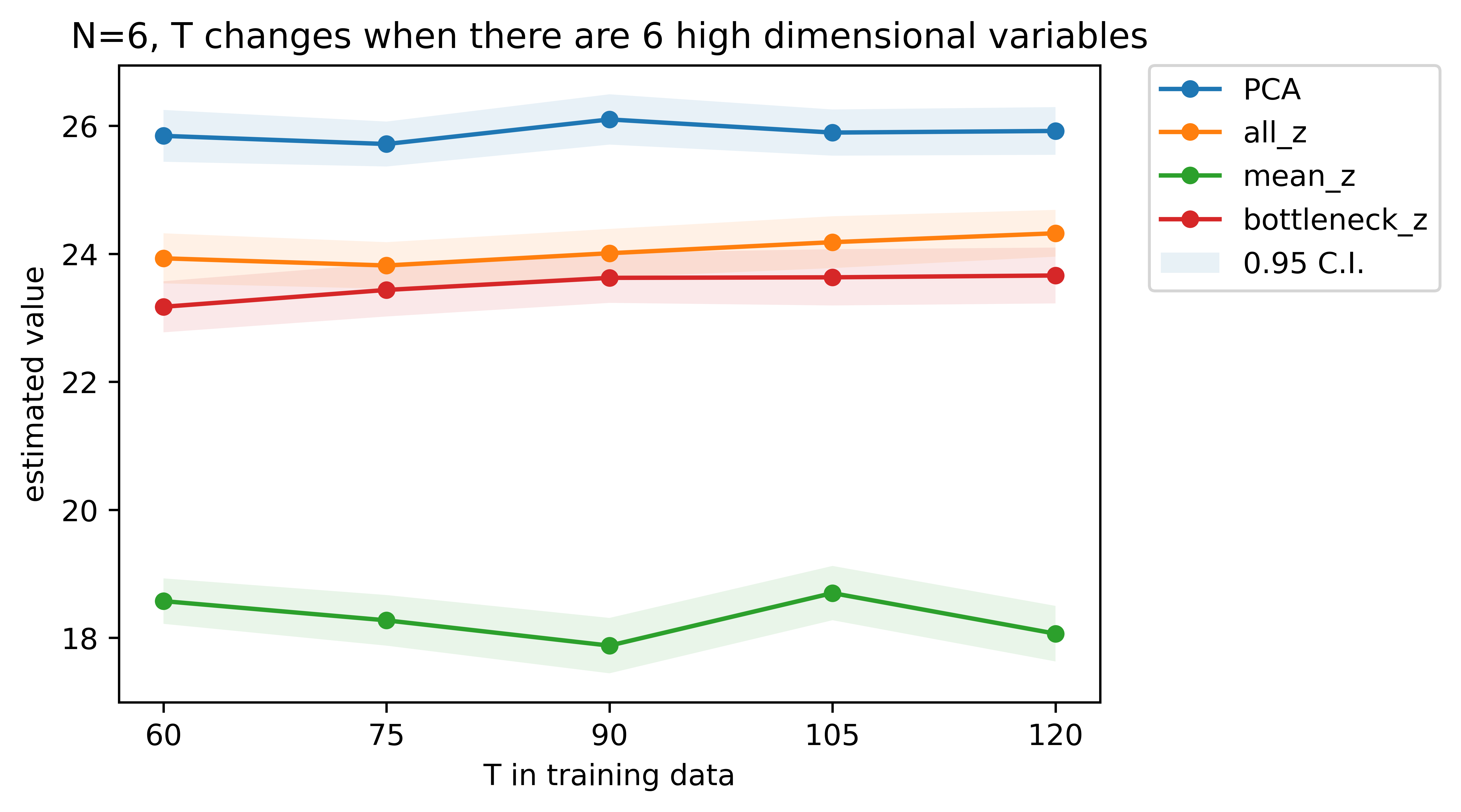}
  \includegraphics[scale=0.42]{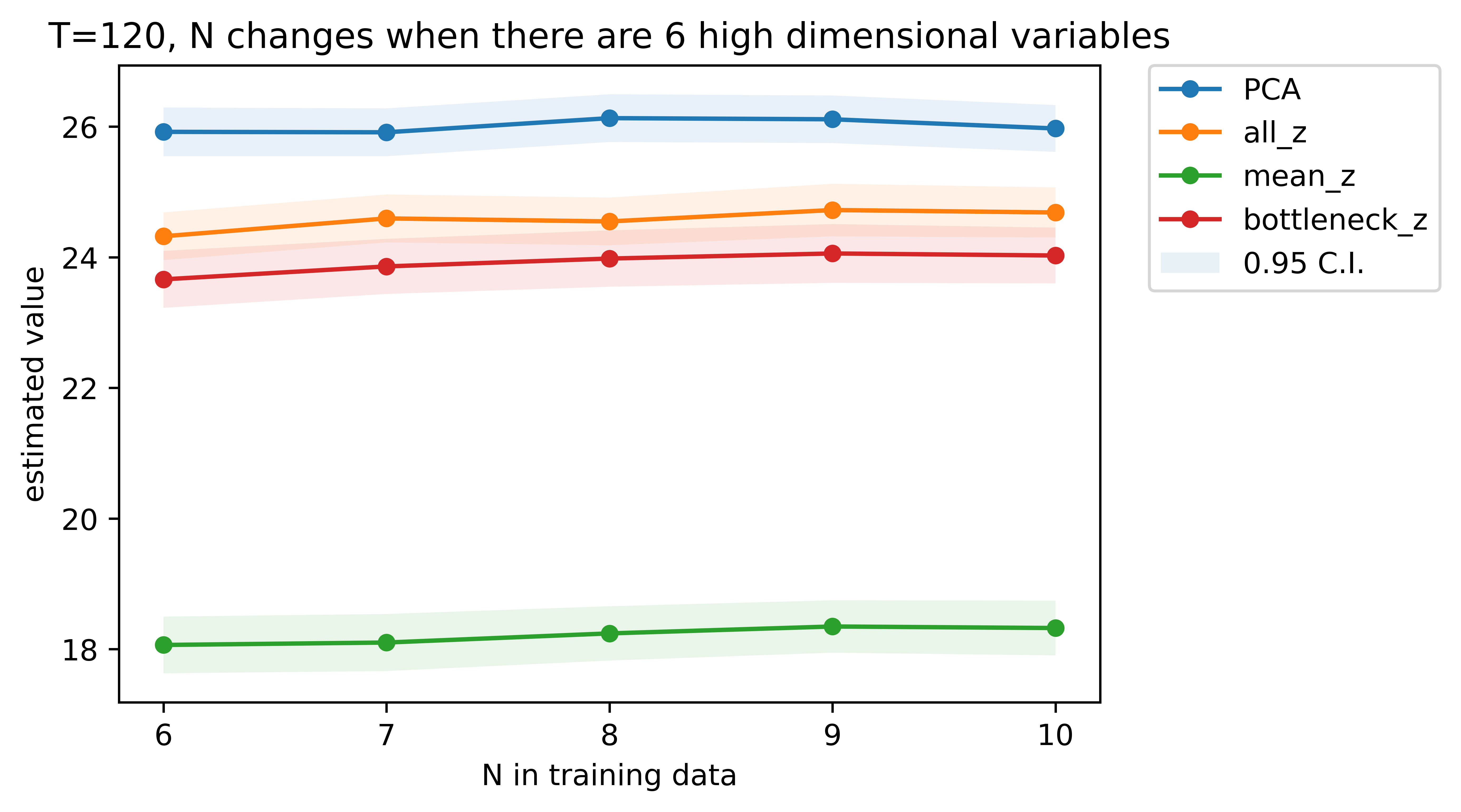}
  \includegraphics[scale=0.42]{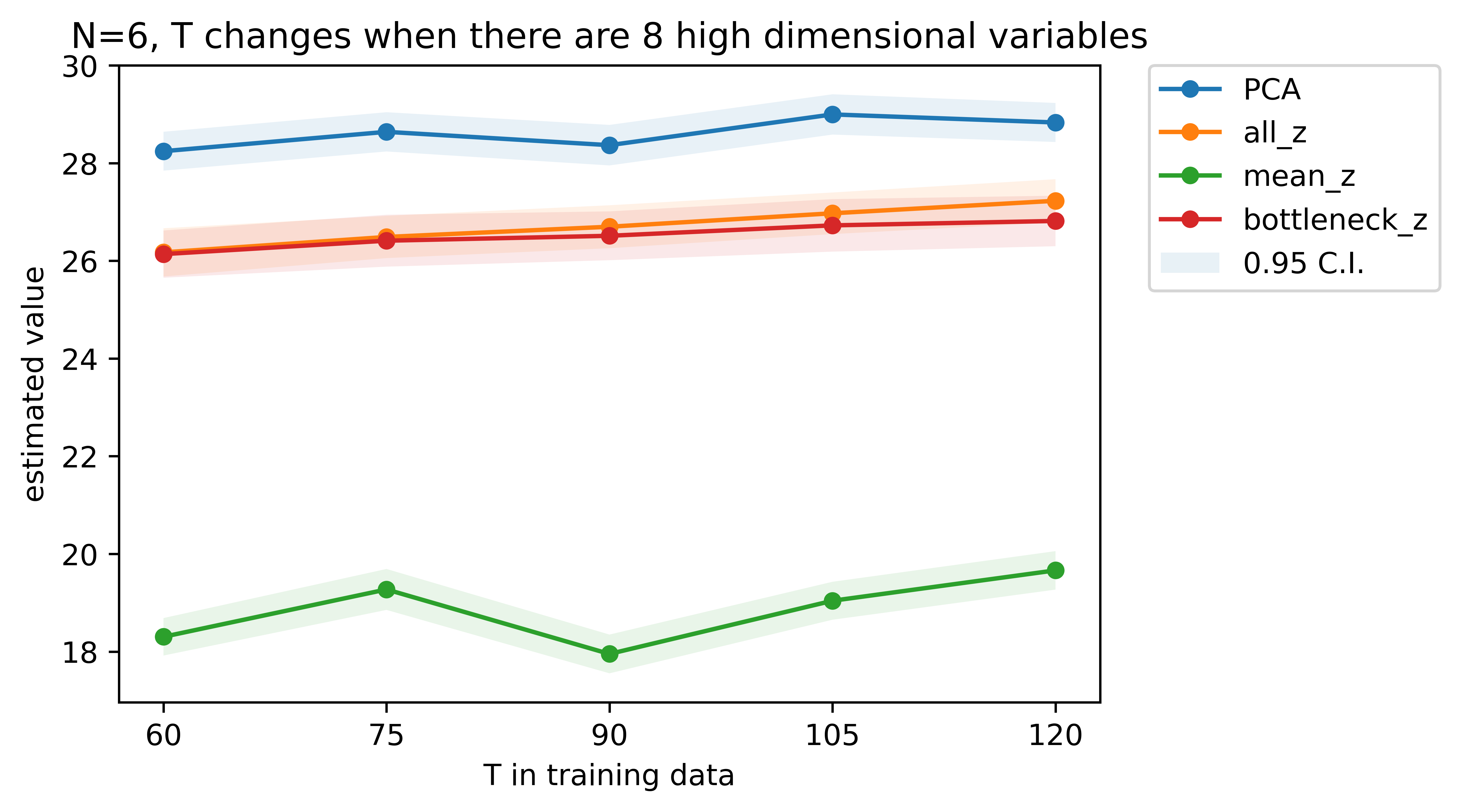}
  \includegraphics[scale=0.42]{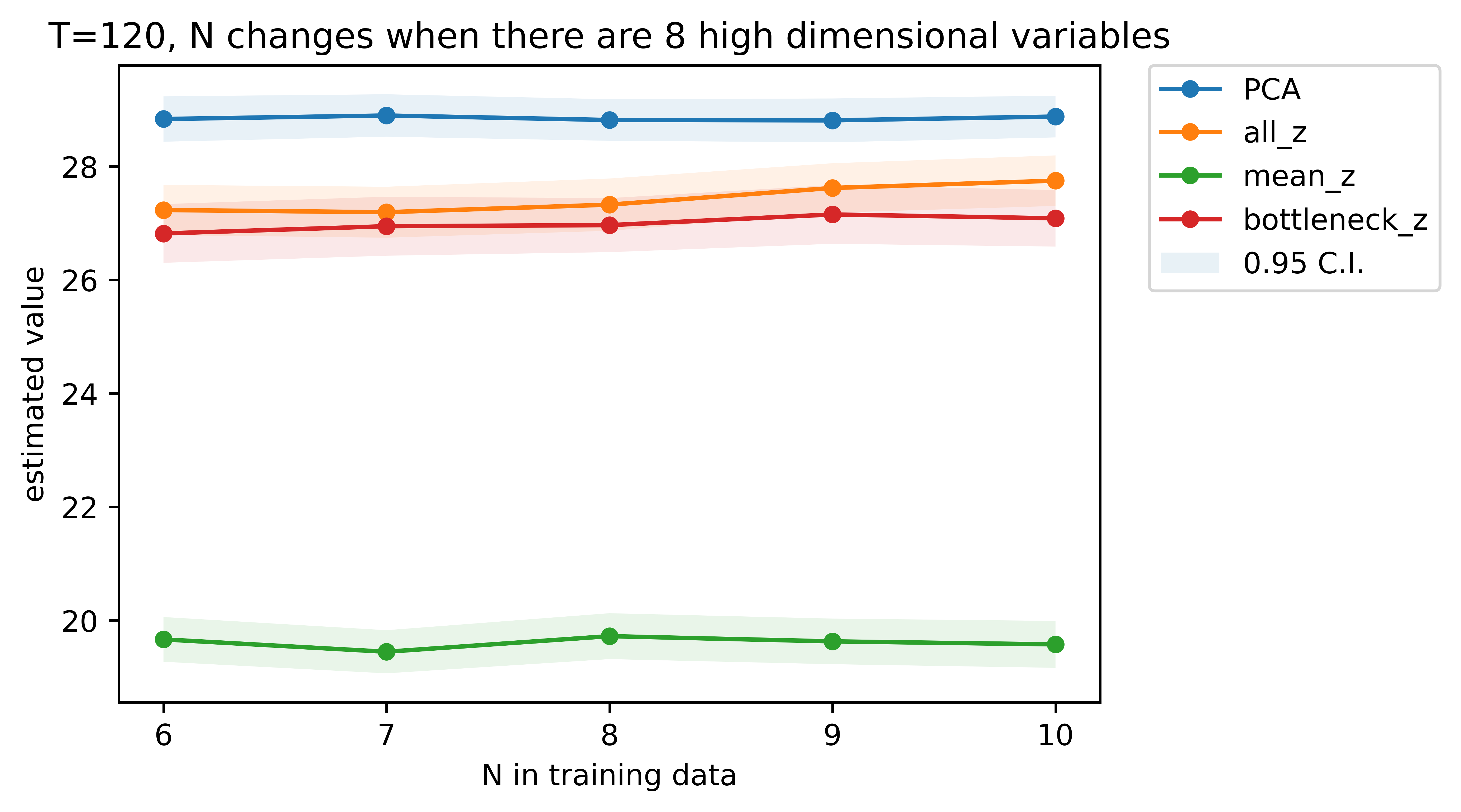}
  \caption{\textbf{Second setting} in simulation. \textbf{Left}: training data with $N=6$ and $T=60,75,90,105,120$ ; \textbf{Right}: training data with $T=120$ and $N=6,7,8,9,10$ when there are $1,2,4,6,8$ variables with dimension $27$ (shaded area is $95\%$ confidence interval). In the legend, ``PCA'' refers to  $\pi_K^{PCA}$; ``all z'' refers to $\pi_K^{ALL}$; ``bottleneck z'' refers to  $\pi_K^{BOTTLE}$; ``mean z''  refers to  $\pi_K^{AVE}$.} 
  \label{f:figure_2_2}
\end{figure}

In the second setting, the $J$ high frequency variables are independent with each other. For each $j$, all the elements in $Z_j$ are dependent and eigenvalues of $Z_j$ decays at the same exponential order as the eigenvalues of the concatenated high frequency variable $Z$ in the first setting. Therefore, more principal components are needed to guarantee that the number of variance explained exceeds $95\%$, as $J$ increases. 
Specifically, when $J=1,2,4,6,8$ high frequency variables, the corresponding $\kappa$ is given by $5,10,20,30,40$ accordingly. See Figure \ref{f:figure_2_pca_num} for details. The expected returns of all estimated optimal  policies are plotted in Figure \ref{f:figure_2_2}. 

From Figures \ref{f:figure_2} and \ref{f:figure_2_2}, 
it can be seen that the proposed policy $\pi_K^{PCA}$ always achieves a larger value than the three baseline policies. Meanwhile, $\pi_K^{ALL}$ and $\pi_K^{BOTTLE}$ perform comparably. The value of both of them are significantly affected by the training size $n$. In addition, 
$\pi_K^{AVE}$ outperforms $\pi_K^{ALL}$ and $\pi_K^{BOTTLE}$ in small samples, but performs worse than the two policies when the sample size is large. In the second setting, 
$\pi_K^{ALL}$ and $\pi_K^{BOTTLE}$ tend to perform much better than $\pi_K^{AVE}$ when $J = 4,6,8$, since averaging over several high frequency variables will lose more relevant information for policy learning. 

\begin{figure}[h!]
  
  \centering
  \includegraphics[scale=0.5]{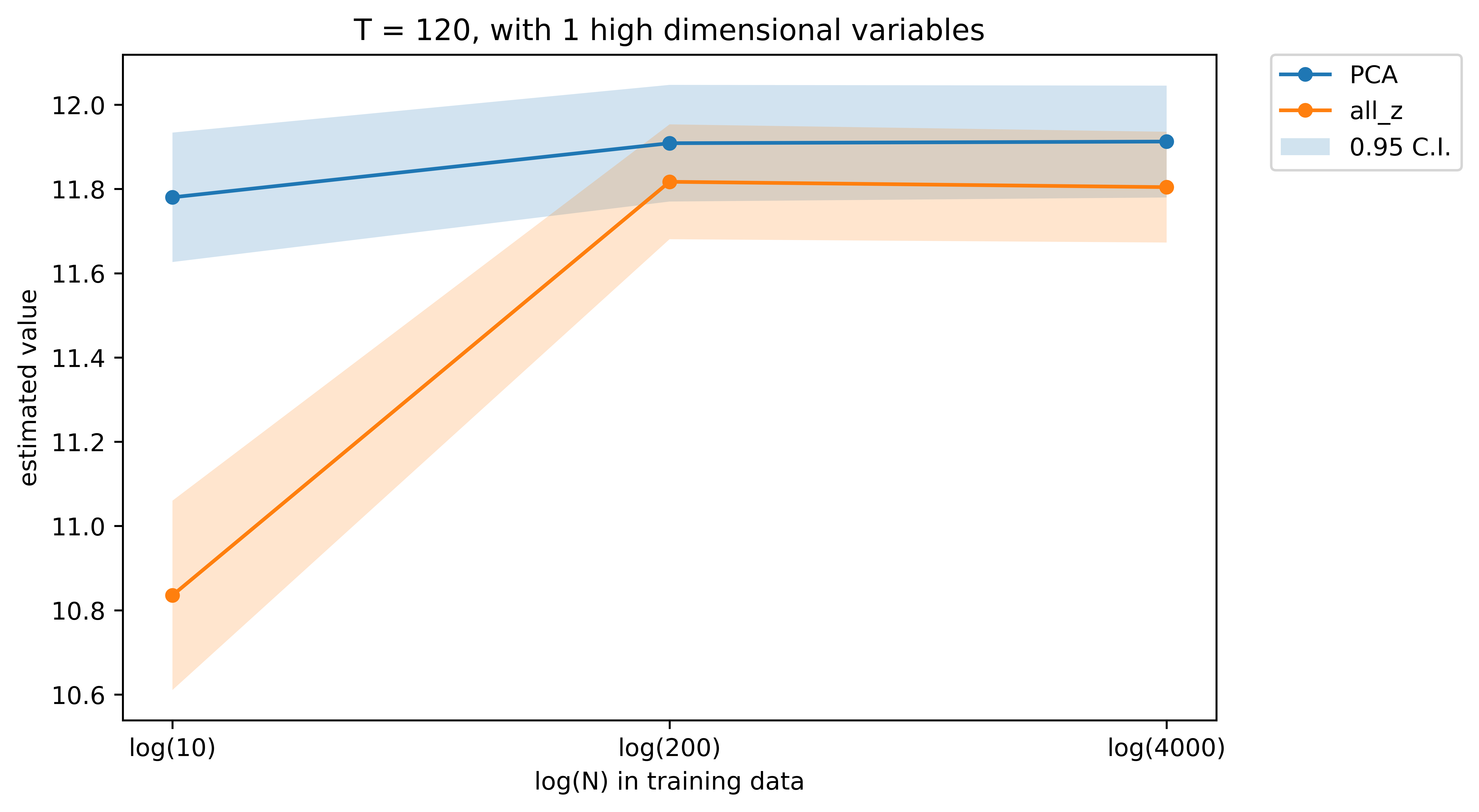}
  \caption{Performance of $\pi^{PCA}$ and $\pi^{ALL}$ as training data size $n$ goes to extremely large ($T$ fixed to be $120$, $N$ from $10$ to $200$ and $4000$). The x-axis presents $\log{N}$. In this figure, shaded area is $95\%$ confidence interval.In the legend, ``PCA'' refers to  $\pi_K^{PCA}$; ``all z'' refers to $\pi_K^{ALL}$.} 
  \label{f:figure_2_ablation}
\end{figure}

Finally, we conduct an additional simulation study with large training datasets where $N=200$ or $4000$ and $T=120$. This setting might be unrealistic in an mHealth dataset. It is included only to test the performance of $\pi_K^{ALL}$. As $\pi_K^{ALL}$ is consistent as well, we anticipate that the difference  between the value under $\pi_K^{ALL}$ and the proposed policy will be negligible as the sample size grows to infinity. 
Figure \ref{f:figure_2_ablation} depicts the results. 
As expected, we observe 
no significant difference between $\pi^{PCA}$ and $\pi^{ALL}$ 
when $N\ge 200$. 

\subsection{Application on OhioT1DM Dataset}\label{s:real_data}
We apply the proposed Algorithm \ref{alg:algorithm1} on the updated OhioT1DM Dataset by \citet{Marling2018ohiodata}. In the real data case, we would still like to compare the behaviors of the four policies: $\pi_K^{PCA}$, $\pi_K^{ALL}$, $\pi_K^{BOTTLE}$ and $\pi_K^{AVE}$. 
OhioT1DM Dataset contains medical information of 12 type-I diabetes patients, including the CGM blood glucose levels of the patients, insulin doses applied during this period, self-reported information of meals and exercises, and other variables recorded by mobile phone apps and physiological sensors. 
The high frequency variables in the OhioT1DM Dataset, such as CGM blood glucose levels, are recorded every 5 minutes. The data for exercises and meals are collected with a much lower frequency, say recorded every few hours. Moreover, considering the basal insulin rate of in this dataset, although this variable is also collected every $5$ minutes, it usually remains a constant for several hours in a day. Thus, the basal insulin rate can also be regarded as a low-frequency scalar variable by taking the average of it. The time period between two decision points is set as $3$ hours, as we only consider non-emergency situations where patients don't need to take bolus injection promptly. In other studies using the OhioT1DM Dataset, the treatment decision frequency is also set to be much lower than the recording frequency of CGM blood glucose levels \citep[see e.g.,][]{Shi2020SAVE,Zhou2021ohio1dtmptlearning,zhu2020causal}.

For the low frequency covariate $X_{i,t}=(X_{i,t}^{(1)},X_{i,t}^{(2)})$, $X_{i,t}^{(1)}$  is constructed based on the $i$th patient’s
self-reported carbohydrate estimate for the meal during the past $3$-hour-interval $[t-1,t)$. The second scalar variable in $X_{i,t}$ is defined as the average of the basal rate of insulin dose
during the past interval $[t-1,t)$. We consider one high-frequency element, $Z_{i,t}$, which contains CGM blood glucose levels
recorded every 5 minutes during the past $3$ hours (its dimension is $m=36$). The action variable $A_{i,t}$ is set to $1$ when the total amount of insulin delivered to the
$i$-th patient is greater than one unit in the past interval. The response variable $Y_{i,t}$ is defined according to the Index of Glycemic
Control \citep[IGC,][]{Rodbard2009glucose}, which is a non-linear function of the blood glucose levels in the following stage. 
A higher IGC value indicates that the blood glucose level of this patient stays close or falls in to the proper range of glucose level. 

\begin{figure}[h!]
  
  \centering
  \includegraphics[scale=0.5]{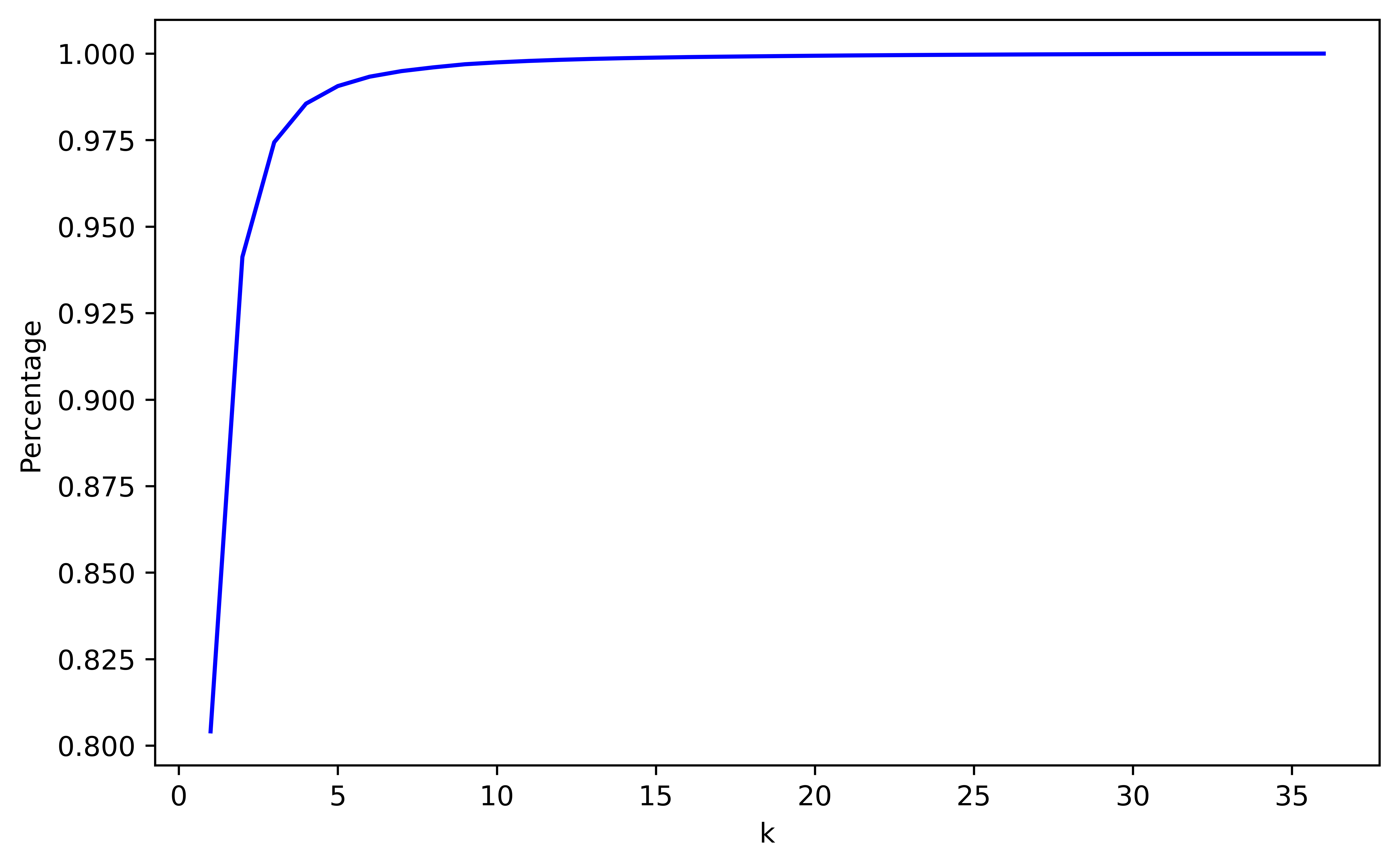}
  \caption{Proportion of variance explained by the first $\kappa$ principal components for CGM blood glucose level in OhioT1DM data.}

  \label{f:figure_4}
\end{figure}

In this study, $\kappa=5$ is selected when training $\pi_K^{PCA}$, as the proportion of variance explained by the first $5$ principal components is over $99\%$, as is shown in Figure \ref{f:figure_4}. The ReLU network here is with 2 hidden layers and width $d_i=6, i=1,2$ (dropout layers with $10\%$ dropout rate added between layer 1, layer 2 and the output layer). 
To estimate the value $V^{\pi}(x,z)$ of the four policies, we use the Fitted Q Evaluation algorithm proposed by \citet{Le2019FQE}. When applying the Fitted Q Evaluation algorithm, a random forest model is used to fit the estimated Q-function of the policy to be evaluated. 
By dividing $12$ patients into a training set of $9$ patients and testing set of $3$ patients, there are $220$ repetitions with different patient combinations. 
In each repetition, the data of $9$ patients is used to train the policy and fit the random forest for Fitted Q Evaluation corresponding to this policy. 
The data of the other $3$ patient is used for approximating the value of the policy using the estimated Q-function from Fitted Q Evaluation. The sample mean of estimated values from all $220$ repetitions is taken as our main result and the standard errors are used to construct the margin of error. To compare the performance of our proposed policies $\pi_K^{PCA}$ against $\pi_K^{ALL}$, $\pi_K^{AVE}$, and $\pi_K^{BOTTLE}$, we present the difference of estimated values of $\pi^{PCA}$ and the three other policies in Table \ref{t:table1}, where margin of error is standard error of the mean difference multiplied by the critical value $1.96$. The estimated values of the four policies is in Table \ref{t:supplementary_table}.

  




Based on the result, it can be shown that the estimated value of $\pi_K^{PCA}$ is higher than all three baselines. The policy $\pi_K^{AVE}$ obtained by using the average of CGM blood glucose levels is commonly used in literature \citep{Zhou2021ohio1dtmptlearning,zhu2020causal}. The less plausible performance of $\pi_K^{AVE}$ is probably due to the information loss by simply replacing the CGM blood glucose levels with its average.
On the other hand, the size of training data is relatively small, as we didn't use all the data recorded in eight weeks due to the large chunks of missing values. Eventually training data from about 5 consecutive weeks are used for training DTR policies. In such scenarios, using the original high frequency vector $Z_{i,t}$ will significantly increase the complexity of the ReLU network structure, such that the number of parameters to be trained is too large compared to the size of training data. Thus, $\pi_K^{ALL}$ and $\pi_K^{BOTTLE}$ cannot outperform $\pi_K^{PCA}$ where input dimension is reduced by PCA. The results shown in Table \ref{t:table1} and Table \ref{t:supplementary_table} agree with the results in Section \ref{s:simulation_study}.

\section{Discussions}\label{s:discussion}
In summary, we propose a deep spectral fitted Q-iteration algorithm to handle mixed frequency data in infinite horizon settings. The algorithm 
relies on the use of PCA for dimension reduction and the use of deep neural networks to capture the non-linearity in the high-dimensional system. In theory, we establish a regret bound for the estimated optimal policy. Our theorem provides an asymptotic guideline 
for selecting the number of principal components. 
In empirical studies, 
we demonstrate the superiority of the proposed algorithm over baseline methods without dimension reduction or use ad-hoc summaries of the state. 
We further offer practical guidelines to select the number of principal components. 
The proposed paper is built upon on PCA and deep Q-learning. It is worthwhile to investigate the performance of other dimension reduction and RL methods. We leave it for future research.


\bibliographystyle{jasa}

\bibliography{reference}

\appendix

\section{Supplementary Definitions}\label{sec:def_two_func_class}

In this section we would like to introduce more definitions, which will be used in the rest of supporting information.

Note that $\mathcal{F}_{0}$ is a function class to model the Q-function without using PCA.
We need to define two more function classes based on PCA of the high dimensional input $z$: 
\begin{definition}\label{def_F_12}
\begin{equation*}
    \begin{aligned}
    \mathcal{F}_{1}(L,\{d_i\}_{i=0}^{L+1},s,V_{max},G,\kappa)=\{f_1:\mathcal{S}\times \mathcal{A}\rightarrow\mathbb{R},f_1(x,z,a)=\\f_0(x,(\sum_k^{\kappa}{U_{k}{U_{k}}^{T}})z,a)=f_0(x,z^{*},a), f_0\in \mathcal{F}_0(L,\{d_i\}_{i=0}^{L+1},s,V_{max})\},
    \end{aligned}
\end{equation*} where $\mathcal{F}_0$ is defined in Section \ref{s:asymp_property} and $G=\sum_{k=1}^{m}{\lambda_{k}\mathbf{U_{k}}\mathbf{U_{k}}^T}$ is the covariance matrix of $z$. $z^{*}=(\sum_k^{\kappa}{U_{k}{U_{k}}^{T}})z$ is the vector recovered from the first $\kappa$ PCA values of the original high-frequency vector $z$. 


The definition of $\mathcal{F}_{2}$ is already mentioned in Section \ref{s:asymp_property}. Here we would like to formally define it as below: 
\begin{equation*}
    \begin{aligned}
        \mathcal{F}_{2}(L,\{d_i\}_{i=0}^{L+1},s,V_{max},G,\kappa)= \{f_2:\mathcal{S}\times \mathcal{A}\rightarrow\mathbb{R},f_2(x,z,a)=\\f_0(x,\begin{pmatrix}
I_{\kappa} & 0_{\kappa\times(m-\kappa)}
\end{pmatrix}{\Sigma}^{-\frac{1}{2}}U^{T}z,a)=f_0(x,v_{\kappa},a),f_0\in \mathcal{F}_0(L,\{d_i\}_{i=0}^{L+1},s,V_{max})\},
    \end{aligned}
\end{equation*} where $v_{\kappa}$ is the vector containing first $\kappa$ PCA values of the original high-frequency vector $z$. 

\end{definition}

 Note that $\mathcal{F}_{2}(L,\{d_i\}_{i=0}^{L+1},s,V_{max},G,\kappa)$ is the function class that we use in Spectrum Q-Iterated Learning Algorithm to model the Q-function. In practice, we will use the estimated $\hat{G}$ and the corresponding PCA vector $\hat{v}_{\kappa}$ in $\mathcal{F}_{2}$. 
 The relationship between the two classes $\mathcal{F}_{1}$ and $\mathcal{F}_{2}$ will be used in the proof of the main theorems (Proposition 1 and Lemma 2 in Section \ref{s:proof_of_thms} of Appendix).

Here we list the definition of the concentration coefficient as below:
\begin{definition}\label{def:concentration_coef}
Let $\sigma_1$ and $\sigma_2$ be two absolutely continuous distributions on $\mathcal{S}\times\mathcal{A}$. Let $\{\pi_t\}_{t\ge 1}=\{\pi_1,…,\pi_m\}$ be a sequence of policies. For a  action-state pair $S_0,A_0$ from the distribution $\sigma_1$, denote the state at $t$-th decision point by $S_t, t=1,2,...,m$. Suppose we take action $A_t$ at $t$-th decision point according to $\pi_t$. Then we denote the marginal distribution of $(S_t,A_t)$ by $P^{\pi_{m}}P^{\pi_{m-1}}...P^{\pi_{1}}\sigma_1$. Then we can define the concentration coefficient by \begin{equation}\label{eqa:def_concentration_coef}
    \begin{aligned}
    \omega_{\infty}(m;\sigma_1,\sigma_2)=\sup_{\pi_1,...,\pi_m}{\left\Vert \frac{dP^{\pi_{m}}P^{\pi_{m-1}}...P^{\pi_{1}}\sigma_1(s,a)}{d\sigma_2(s,a)}\right\Vert_{\infty,\sigma_2}}.
    \end{aligned}
\end{equation}
\end{definition}

Similar definitions of concentration coefficients can be commonly found in the literature of reinforcement learning \citep{Fan2020Qlearn,Lazaric2016concentration}. 


\section{Detailed Version of the First Theorem}\label{sec: full_thm_1}




The detailed version of Theorem \ref{theorem_1} is: Let $\mu$ be some distribution on $(\mathcal{S}\times \mathcal{A})$ such that $\mu(\mathcal{S}\times \mathcal{A})=\mu_1(\mathcal{S})\times \mu_2(\mathcal{A})$ with $\mu_2(a)>0, \forall a\in\mathcal{A}$ and $\frac{d\mu_1(s)}{ds}$ bounded away from $0$ (here we would like to include a more general measure $\mu_2$ for the action space $\mathcal{A}$). Define $\beta_j^{*}=\beta_j\times\prod_{l=j+1}{\min{(\beta_l,1)} }$ and $\alpha^*=\max_{j\in[q]}{ \frac{t_j}{2\beta_j^*+t_j}}$, where $\beta_j,q,t_j,p_j$ come from Definition \ref{def_2} and definition of $\mathcal{G}_0(\{p_j,t_j,\beta_j,H_j\}_{j\in[q]})$ in Section \ref{s:asymp_property}.


Under the Assumptions \ref{ass:a1} to \ref{ass:a3}, and the condition that $n$ is large enough such that $\max\{\sum_{j}{{(t_j+\beta_j+1)}^{3+t_j}},\max_j{p_j}\}\le C_\xi{(\log(n))}^\xi$ for some $\xi>0,C_\xi>0$, there exists a function class $\mathcal{F}_{2}(L^*,\{d_j^*\}_{i=0}^{L+1},s^*,V_{max},\hat{G},\kappa)$ modeling $Q(s,a)$ such that its components are sparse ReLU networks with hyperparameters satisfying: layer number $L^*=O({\log^{\xi^*}{n}})$, first layer width satisfying the bound $d_1^*\le n^{\alpha^*}$, output dimension $d^*_{L^*+1}=1$, other layers width satisfying the bound $d_0^* = m_0 + m\le min_{j\in\{1,2,...,L\}}{d_j^*}\le  max_{j\in\{1,2,...,L\}}{d_j^*}=O(n^{\xi^*})$, and sparsity with order $s^*\asymp d_1^*\kappa + n^{\alpha^{*}}\log^{\xi^*}(n)$ for some $\xi^{*}>1+2\xi$. In this case, we can show that $\tilde{Q}_K$ (modeled by $\mathcal{F}_{2}$)  obtained from our Algorithm \ref{alg:algorithm1} satisfies that:   


\begin{equation*}
    \begin{aligned}
    \left\Vert Q^{\pi^*}-\tilde{Q}_K\right\Vert_{2,\mu}^2=O_p(|\mathcal{A}|(n^{\alpha^*}\log^{\xi^*}{n}+d_1^*\kappa)n^{-1}\log^{\xi^*+1}{n}\\+L_{\mathcal{F}_0}(e^{-\zeta\kappa}-e^{-\zeta m}+n^{-2\Delta})+\frac{\gamma^{2K}}{(1-\gamma)^2}R_{max}^2),
    \end{aligned}
\end{equation*} 
where $|\mathcal{A}|$ is the number of treatment options. 


\section{Proof of theorems}\label{s:proof_of_thms}



\subsection{Proof Sketch of Theorem \ref{theorem_1}}\label{s:proof_sketch_thm_1}





The detailed version of Theorem \ref{theorem_1} is: Let $\mu$ be some distribution on $(\mathcal{S}\times \mathcal{A})$ such that $\mu(\mathcal{S}\times \mathcal{A})=\mu_1(\mathcal{S})\times \mu_2(\mathcal{A})$ with $\mu_2(a)>0, \forall a\in\mathcal{A}$ and $\frac{d\mu_1(s)}{ds}$ bounded away from $0$ (here we would like to include a more general measure $\mu_2$ for the action space $\mathcal{A}$). Define $\beta_j^{*}=\beta_j\times\prod_{l=j+1}{\min{(\beta_l,1)} }$ and $\alpha^*=\max_{j\in[q]}{ \frac{t_j}{2\beta_j^*+t_j}}$, where $\beta_j,q,t_j,p_j$ come from Definition \ref{def_2} and definition of $\mathcal{G}_0(\{p_j,t_j,\beta_j,H_j\}_{j\in[q]})$ in Section \ref{s:asymp_property}.


Under the Assumptions \ref{ass:a1} to \ref{ass:a3}, and the condition that $n$ is large enough such that $\max\{\sum_{j}{{(t_j+\beta_j+1)}^{3+t_j}},\max_j{p_j}\}\le C_\xi{(\log(n))}^\xi$ for some $\xi>0,C_\xi>0$, there exists a function class $\mathcal{F}_{2}(L^*,\{d_j^*\}_{i=0}^{L+1},s^*,V_{max},\hat{G},\kappa)$ modeling $Q(s,a)$ such that its components are sparse ReLU networks with hyperparameters satisfying: layer number $L^*=O({\log^{\xi^*}{n}})$, first layer width satisfying the bound $d_1^*\le n^{\alpha^*}$, output dimension $d^*_{L^*+1}=1$, other layers width satisfying the bound $d_0^* = m_0 + m\le min_{j\in\{1,2,...,L\}}{d_j^*}\le  max_{j\in\{1,2,...,L\}}{d_j^*}=O(n^{\xi^*})$, and sparsity with order $s^*\asymp d_1^*\kappa + n^{\alpha^{*}}\log^{\xi^*}(n)$ for some $\xi^{*}>1+2\xi$. In this case, we can show that $\tilde{Q}_K$ (modeled by $\mathcal{F}_{2}$)  obtained from our Algorithm \ref{alg:algorithm1} satisfies that:   


\begin{equation*}
    \begin{aligned}
    \left\Vert Q^{\pi^*}-\tilde{Q}_K\right\Vert_{2,\mu}^2=O_p(|\mathcal{A}|(n^{\alpha^*}\log^{\xi^*}{n}+d_1^*\kappa)n^{-1}\log^{\xi^*+1}{n}\\+L_{\mathcal{F}_0}(e^{-\zeta\kappa}-e^{-\zeta m}+n^{-2\Delta})+\frac{\gamma^{2K}}{(1-\gamma)^2}R_{max}^2),
    \end{aligned}
\end{equation*} 
where $|\mathcal{A}|$ is the number of treatment options.




\hfill \break
First, we need to denote \begin{equation*}
    \mathcal{F}_1(L^*,\{\{d_i^*\}^{L+1}_{i=1},d_0=m_0+m\},s^*-d_1\kappa,V_{max}, \hat{G},\kappa )
\end{equation*} as $\mathcal{F}_1$,   \begin{equation*}
    \mathcal{F}_2(L^*,\{d_i^*\}^{L+1}_{i=0},s^*,V_{max},\hat{G},\kappa)
\end{equation*} as $\mathcal{F}_2$ and \begin{equation*}
    \mathcal{F}_0(L^*,\{\{d_i^*\}^{L+1}_{i=1},d_0=m_0+m\},s^*-d_1\kappa,V_{max})
\end{equation*} as $\mathcal{F}_0$. 

Note that it has been assumed that the transition density function $q(s'|s,a)$ satisfies that $\sup_{s',s,a}{q(s'|s,a)}\le c_q$ (as is stated in Section \ref{s:assump_policy_value_function}). Let $q^j(s'|s,a)$ denote the density of the state $s'$ following the decisions of the sequence of policy $\pi_1,\pi_2,...,\pi_{j-1}$. Then it can be shown that \begin{equation*}
    \begin{aligned}
    q^j(s'|s,a)=\int_{x\in\mathcal{S}}{\sum_{a_x\in\mathcal{A}}{q(s'|x,a)\pi_{j-1}(a_x|x)q^{j-1}(x|s,a)dx}}\\\le c_q \int_{x\in\mathcal{S}}{\sum_{a_x\in\mathcal{A}}{\pi_{j-1}(a_x|x)q^{j-1}(x|s,a)dx}}=c_q.
    \end{aligned}
\end{equation*}
Then the density for joint distribution of $s',a'$ after $\{\pi_1,...,\pi_m\}$ is 
\begin{equation*}
    \begin{aligned}
    \int_{(s,a)\in\mathcal{S}\times\mathcal{A}}{\pi_m(a'|s')q^{(m)}}(s'|s,a)d\mu(s,a) \le c_q \int_{(s,a)\in\mathcal{S}\times\mathcal{A}}{d\mu(s,a)}=c_q.
    \end{aligned}
\end{equation*}
Thus, $\frac{dP^{\pi_{m}}P^{\pi_{m-1}}...P^{\pi_{1}}\mu(s,a)}{d\sigma(s,a)}$ can be bounded as long as the sampling distribution $\sigma$ is bounded away from $0$. That is, $\omega_\infty(m;\mu,\sigma)\le c_{\mu,\sigma}$ for some constant $c_{\mu,\sigma}>0$ ( recall that $\omega_\infty(m;\mu,\sigma)$ is from Definition \ref{eqa:def_concentration_coef}). 

Then it can be shown that \begin{equation*}\label{eqa:omega_bound}
    \begin{aligned}
    \frac{1}{(1-\gamma)^2}\sum_{m=0}^{\infty}{\gamma^{m-1}(m+1)[\omega_\infty(m;\mu,\sigma)]^{\frac{1}{2}}}\le \frac{1}{(1-\gamma)^2}\sum_{m=0}^{\infty}{\gamma^{m-1}(m+1){c_{\mu,\sigma}}^{\frac{1}{2}}}=\phi_{\mu,\sigma},
    \end{aligned}
\end{equation*} for some constant $\phi_{\mu,\sigma}>0$.


Then we can show a lemma regarding the approximation error of $\tilde{Q}_K$ as below: 

\begin{lemma}\label{lemma_6_1}
For the estimated Q-function obtained in iteration $K$, $\tilde{Q}_K$ in Algorithm \ref{alg:algorithm1}, it leads to  
\begin{equation}\label{equation_lemma_6_1}
    \begin{aligned}
    \left\Vert Q^*-\tilde{Q}_K\right\Vert_{2,\mu}^2 \le 2 \gamma^2(1-\gamma)^4\epsilon_{max}^2\phi_{\mu,\sigma}^{2}+8\frac{\gamma^{2K}R_{max}^2}{(1-\gamma)^2},
    \end{aligned}
\end{equation} where $\epsilon_{max}=\max_{k\in\{1,2...,K\}}{\left\Vert T\tilde{Q}_{k-1}-\tilde{Q}_{k}\right\Vert_{\sigma}}$.
\end{lemma}

That is, the distance of the $K$-th iteration result $\tilde{Q}_K$ in our algorithm with $Q^*$ corresponding to the optimal policy can be bound by the largest approximation error $\left\Vert T\tilde{Q}_{k-1}-\tilde{Q}_{k}\right\Vert_{\sigma}^2$ among all the $K$ iterations in the Spectrum Q-Iteration algorithm. We know $\tilde{Q}_{k-1},\tilde{Q}_{k}\in\mathcal{F}_2$. 

To bound the term $\left\Vert T\tilde{Q}_{k-1}-\tilde{Q}_{k}\right\Vert_{\sigma}^2$ in each iteration, we will apply Theorem 6.2 in \citet{Fan2020Qlearn}, and it can be shown that \begin{equation*}\label{equation_2}
\left\Vert T\tilde{Q}_{k-1}-\tilde{Q}_{k}\right\Vert_{\sigma}^2 \le(1+\epsilon)^2 \omega(\mathcal{F}_2)+C\frac{V_{max}^2}{n\epsilon}\log{N_{\delta,2}}+C^{'}V_{max}\delta
\end{equation*}
for $\forall \epsilon\in(0,1]$, where $\omega(\mathcal{F}_2)=\sup_{g\in\mathcal{F}_{2}}{\inf_{f\in\mathcal{F}_{2}}{\left\Vert f-Tg \right\Vert_\sigma^2}}$. Here $N_{\delta,2}$ is the cardinality of the minimal $\sigma$-covering set of the function class $\in\mathcal{F}_{2}$ with respect to $l_{\infty}$ norm. 

If we take $\epsilon=1$ and fix $\delta=\frac{1}{n}$, we have \begin{equation}\label{equation_3}
\left\Vert T\tilde{Q}_{k-1}-\tilde{Q}_{k}\right\Vert_{\sigma}^2 \le4 \omega(\mathcal{F}_2)+C\frac{V_{max}^2}{n}\log{N_{\frac{1}{n},2}}+\frac{C^{'}V_{max}}{n}.
\end{equation}

Then we need to bound $\omega(\mathcal{F}_2)=\sup_{g\in\mathcal{F}_{2}}{\inf_{f\in\mathcal{F}_{2}}{\left\Vert f-Tg \right\Vert_\sigma^2}}$ and $\log{N_{\frac{1}{n},2}}$ respectively.

The term $\sup_{g\in\mathcal{F}_{2}}{\inf_{f\in\mathcal{F}_{2}}{\left\Vert f-Tg \right\Vert_\sigma^2}}$ is related to the error of using a function $f\in\mathcal{F}_{2}$ to approximate a function $g$ from the same class after optimal Bellman operation. Note that in each iteration of the Q-iterative Algorithm, we use $\tilde{Q}_k$ from a ReLU-like class to fit the responses $Y_i=R_i+\gamma \max_{a\in\mathcal{A}}{\tilde{Q}_{k-1}(S_i^{'},a)}$. If we take expectation of $Y_i$, we know the true model underlying the responses $Y_i$ would be a function from the class of $Tg:g\in\mathcal{F}_{2}$. As is argued in \citet{Fan2020Qlearn},  the term $\sup_{g\in\mathcal{F}_{2}}{\inf_{f\in\mathcal{F}_{2}}{\left\Vert f-Tg \right\Vert_\sigma^2}}$ can be represented as the bias term when we fit $Y_i$ by using $\tilde{Q}_k\in\mathcal{F}_{2}$.

We will first bound this bias-related term. We need to show the following proposition: 
\begin{proposition}\label{statement_1}
The function class $\mathcal{F}_1(L,\{\{d_i\}^{L+1}_{i=1},d_0=m_0+m\},s,V_{max}, \hat{G},\kappa)$ is a subset of $\mathcal{F}_2(L,\{\{d_i\}^{L+1}_{i=1},d_0= \kappa+m_0\},s+d_1\kappa,V_{max}, \hat{G},\kappa)$.  
\end{proposition}

This proposition implies $\mathcal{F}_1\subset\mathcal{F}_2$. Then we can propose the following lemma, which relies on the Proposition \ref{statement_1} we've just shown:
\begin{lemma}\label{lemma_1}
Under the same assumptions of the Theorem \ref{theorem_1},  
\begin{equation*}
\sup_{g\in\mathcal{F}_{2}}{\inf_{f\in\mathcal{F}_{2}}{\left\Vert f-Tg \right\Vert_\sigma^2}}\le 2\sup_{f^{'}\in\mathcal{G}_{0}}{\inf_{f_0\in\mathcal{F}_{0}}{\left\Vert f_0-f^{'} \right\Vert_\sigma^2}}+2L_{\mathcal{F}_0}{\left\Vert Z-Z^{*}\right\Vert_{\sigma}^{2}} ,
\end{equation*} where $L_{\mathcal{F}_0}=\sup_{f\in\mathcal{F}_0}{\sup_{x\ne y}{\frac{{|f(y)-f(x)|}^{2}}{\left\Vert y-x\right\Vert^{2}}}}$. $\{Z\}_{j=1}^{J}$ is the high-frequency vector from the distribution $\sigma$ and $Z^{*}=(\sum_{k=1}^{\kappa}{\hat{U}_k{\hat{U}_k}^{T}})Z$ is the vector recovered from $\kappa$ PCA values of $Z$.
\end{lemma}

By applying Lemma \ref{lemma_1}, we now can bound the term  $\sup_{g\in\mathcal{F}_{2}}{\inf_{f\in\mathcal{F}_{2}}{\left\Vert f-Tg \right\Vert_\sigma^2}}$ by $\sup_{f^{'}\in\mathcal{G}_{0}}{\inf_{f_0\in\mathcal{F}_{0}}{\left\Vert f_0-f^{'} \right\Vert_\sigma^2}}$ and ${\left\Vert Z-Z^{*}\right\Vert_{\sigma}^{2}}$. 

Recall that $\mathcal{F}_0$ is the simple notation of \begin{equation*}
    \mathcal{F}_0(L^*,\{\{d_i^*\}^{L+1}_{i=1},d_0=m_0+m\},s^*-d_1\kappa,V_{max}),
\end{equation*} as is specified in the beginning of this section. We know sparsity of $\mathcal{F}_0$ satisfies $s^*-d_1\kappa \asymp  n^{\alpha^{*}}\log^{\xi^*}(n)$ and architectures of $\mathcal{F}_0$ satisfying  $d_1^*\le n$ and $m_0 + m\le min_{j\in\{1,2,...,L\}}{d_j^*}\le  max_{j\in\{1,2,...,L\}}{d_j^*}=O(n^{\xi^*})$. Thus, we can use equation (4.18) of \citet{Fan2020Qlearn} to bound term $\sup_{f^{'}\in\mathcal{G}_{0}}{\inf_{f_0\in\mathcal{F}_{0}}{\left\Vert f_0-f^{'} \right\Vert_\sigma^2}}$: \begin{equation*}\label{equation_5}
    \sup_{f^{'}\in\mathcal{G}_{0}}{\inf_{f_0\in\mathcal{F}_{0}}{\left\Vert f_0-f^{'} \right\Vert_\sigma^2}}\le n^{\alpha^{*}-1},
\end{equation*} where the conditions of neural networks in $\mathcal{F}_0$ matches the conditions of applying equation (4.18) of \citet{Fan2020Qlearn}. 

Now we can bound the term $\left\Vert Z-Z^{*}\right\Vert_{\sigma}^{2}$ by the next lemma.

\begin{lemma}\label{lemma_2}
For the high-frequency vector $Z$, we assume $E(Z)=0$ and the the eigenvalues of $Cov(Z)$ following this exponential decaying trend $\lambda_k=O(e^{-\zeta k})$. We also assume the estimation of the $Cov(Z)$ satisfies that $\left\Vert \hat{U}-U\right\Vert=O_p(n^{-\Delta})$. Then we have
\begin{equation*}\label{equation_6}
\left\Vert  Z-Z^{*}\right\Vert_{\sigma}^2=O_p(e^{-\zeta \kappa}-e^{-\zeta m} + n^{-2\Delta} )     
\end{equation*}

\end{lemma}

Based on these two lemmas, we know that \begin{equation*}\label{equation_7}
    \sup_{g\in\mathcal{F}_{2}}{\inf_{f\in\mathcal{F}_{2}}{\left\Vert f-Tg \right\Vert_\sigma^2}}=O_p(n^{\alpha^{*}-1}+L_{\mathcal{F}_0}{(e^{-\zeta \kappa}-e^{-\zeta m} +n^{-2\Delta})} ). 
\end{equation*}

Then, when we take $\delta=\frac{1}{n}$, a bound for $\log{N_{\delta,2}}$ in equation \eqref{equation_3} is required. Here we need to show another proposition, Proposition \ref{statement_2}:
\begin{proposition}\label{statement_2}
When $\delta=\frac{1}{n}$, $\log{N_{\delta,2}}\le C_1|\mathcal{A}|(n^{\alpha^*}(\log{n})^{\xi^{*}}+d_1^*\kappa)(\log{n})^{1+\xi^{*}}$ for some constant $C_1$.
\end{proposition}

Then from equation \eqref{equation_3}, we know 
\begin{equation}\label{equation_8}
    \left\Vert T\tilde{Q}_{k-1}-\tilde{Q}_{k}\right\Vert_{\sigma}^2=O_p(|\mathcal{A}|n^{-1}(n^{\alpha^*}{(\log{n})}^{\xi^*}+d_1^*\kappa){(\log{n})}^{\xi^*+1}+L_{\mathcal{F}_0}{(e^{-\zeta \kappa}-e^{-\zeta m} +n^{-2\Delta})})
\end{equation}

Then from equation \eqref{equation_8} and equation \eqref{equation_lemma_6_1}, the proof of Theorem \ref{theorem_1} is complete.



\subsection{Proof of Theorem \ref{theorem_2}}\label{s:proof_thm_2}


Theorem \ref{theorem_2}: Under the Assumptions \ref{ass:a1} to \ref{ass:a6} and the conditions of $\mu,\mathcal{F}_2,n$ in Theorem \ref{theorem_1}, we can show:

\begin{equation*}
    \begin{aligned}
    E_{\mu_1}[V^{\pi^*}(s)-V^{\pi_K}(s)]=O_p(\frac{1}{1-\gamma}\{ |\mathcal{A}|(n^{\alpha^*}{\log^{\xi^*}{n}}+d_1^*\kappa)n^{-1}{\log^{\xi^*+1}{n}}\\+L_{\mathcal{F}_0}(e^{-\zeta\kappa}-e^{-\zeta m}+n^{-2\Delta})+\frac{\gamma^{2K}}{(1-\gamma)^2}R_{max}^2\}^{\frac{\eta+1}{\eta+2}}).
    \end{aligned}
\end{equation*}



\hfill \break
\begin{proof}

The policy $\pi_K$ obtained in Algorithm \ref{alg:algorithm1} is a greedy policy with respect to $\tilde{Q}_K$.

First, a proposition from \citet{Shi2020SAVE} can be applied:
\begin{proposition}[Inequality E.10 and E.11 in \citet{Shi2020SAVE}]
\begin{equation}\label{eqa:save_e10_e11}
    \begin{aligned}
    E_{\mu_1}[V^{\pi^*}(s)-V^{\pi_K}(s)]\le \int_{s\in\mathcal{S}}{\sum_{a\in\mathcal{A}}{Q^{\pi^*}(s,a)\{\pi^*(a|s)-\pi_K(a|s)\}d\mu_1(s)}}\\+\frac{c\gamma}{1-\gamma}\int_{s\in\mathcal{S}}{\sum_{a\in\mathcal{A}}{Q^{\pi^*}(s,a)\{\pi^*(a|s)-\pi_K(a|s)\}ds}}.
    \end{aligned}
\end{equation}

\end{proposition}

Then it suffices to bound the two terms $\int_{s\in\mathcal{S}}{\sum_{a\in\mathcal{A}}{Q^{\pi^*}(s,a)\{\pi^*(a|s)-\pi_K(a|s)\}d\mu_1(s)}}$ and $\int_{s\in\mathcal{S}}{\sum_{a\in\mathcal{A}}{Q^{\pi^*}(s,a)\{\pi^*(a|s)-\pi_K(a|s)\}ds}}$. 
We will first show how to bound the following term:\begin{equation*}
    \int_{s\in\mathcal{S}}{\sum_{a\in\mathcal{A}}{Q^{\pi^*}(s,a)\{\pi^*(a|s)-\pi_K(a|s)\}d\mu_1(s)}}.
\end{equation*} 

Define $\hat{a}_{\pi^K}(s)=\argmax_{a\in\mathcal{A}}{\tilde{Q}_K(s,a)}$. For $\epsilon>0$, let $A_*=\{0<\max_{a\in\mathcal{A}}{Q^{\pi^*}(s,a)}-Q^{\pi^*}(s,\hat{a}_{\pi^K})\le \epsilon\}$. Then we know the complement set $A_*^C=\{\max_{a\in\mathcal{A}}{Q^{\pi^*}(s,a)}-Q^{\pi^*}(s,\hat{a}_{\pi^K})>\epsilon\}\cup \{\max_{a\in\mathcal{A}}{Q^{\pi^*}(s,a)}-Q^{\pi^*}(s,\hat{a}_{\pi^K})=0\}$. Then it leads to  \begin{equation}\label{eqa:29}
    \begin{aligned}
    \int_{s\in\mathcal{S}}{\sum_{a\in\mathcal{A}}{Q^{\pi^*}(s,a)\{\pi^*(a|s)-\pi_K(a|s)\}d\mu_1(s)}}=\\
    \int_{s\in\mathcal{S}}{\sum_{a\in\mathcal{A}}{Q^{\pi^*}(s,a)\{\pi^*(a|s)-\pi_K(a|s)\}\mathbbm{1}(s\in A_*)d\mu_1(s)}}+\\
    \int_{s\in\mathcal{S}}{\sum_{a\in\mathcal{A}}{Q^{\pi^*}(s,a)\{\pi^*(a|s)-\pi_K(a|s)\}\mathbbm{1}(s\in A_*^C)d\mu_1(s)}}.
    \end{aligned}
\end{equation} 

The first step is to bound the first part of equation \eqref{eqa:29}, \begin{equation*}
    \int_{s\in\mathcal{S}}{\sum_{a\in\mathcal{A}}{Q^{\pi^*}(s,a)\{\pi^*(a|s)-\pi_K(a|s)\}\mathbbm{1}(s\in A_*)d\mu_1(s)}}.
\end{equation*}

It is known that \begin{equation*}
    \max_{a\in\mathcal{A}}{Q^{\pi^*}(s,a)}=\sum_{a\in\mathcal{A}}{Q^{\pi^*}(s,a)\pi^*(a|s)}
\end{equation*} and \begin{equation*}
    Q^{\pi^*}(s,\hat{a}_{\pi^K})=\sum_{a\in\mathcal{A}}{Q^{\pi^*}(s,a)\pi^K(a|s)},
\end{equation*} as both $\pi^*$ are $\pi^K$ are greedy policies ($\pi^*$ can be viewed as the greedy policy with respect to $Q^{\pi^*}$). Then $\sum_{a\in\mathcal{A}}{Q^{\pi^*}(s,a)\{\pi^*(a|s)-\pi_K(a|s)\}}=\max_{a\in\mathcal{A}}{Q^{\pi^*}(s,a)}-Q^{\pi^*}(s,\hat{a}_{\pi^K})\le \epsilon$ when $s\in A_*$. That is, \begin{equation}\label{eqa:31}
    \begin{aligned}
    \int_{s\in\mathcal{S}}{\sum_{a\in\mathcal{A}}{Q^{\pi^*}(s,a)\{\pi^*(a|s)-\pi_K(a|s)\}}\mathbbm{1}(s\in A_*)d\mu_1(s)}\le \epsilon \int_{s\in\mathcal{S}}{\mathbbm{1}(s\in A_*)d\mu_1(s)}.
    \end{aligned}
\end{equation}

Also, when $s\in A_*$, we know that $\hat{a}_{\pi^K}$ will never be the same as $a\in\argmax_{a'\in\mathcal{A}}{Q^{\pi^*}(s,a')}$ based on the definition of $A_*$, as $\max_{a\in\mathcal{A}}{Q^{\pi^*}(s,a)}-Q^{\pi^*}(s,\hat{a}_{\pi^K})>0$ is always true in $A_*$. Thus, $\max_{ a\in{\mathcal{A}- \argmax_{a'}{Q^{\pi^*}(x,a')} } }{Q^{\pi^*}(s,a)}\ge Q^{\pi^*}(s,\hat{a}_{\pi^K})$. Then we know $\max_{a\in\mathcal{A}}{Q^{\pi^*}(s,a)}-Q^{\pi^*}(s,\hat{a}_{\pi^K})\le \epsilon$ induces $\max_{a\in\mathcal{A}}{Q^{\pi^*}(s,a)} - \max_{ a\in{\mathcal{A}- \argmax_{a'}{Q^{\pi^*}(x,a')} } }{Q^{\pi^*}(s,a)}\le \epsilon$. 

Therefore, it can be obtained that \begin{equation*}
    \begin{aligned}
    \int_{s\in\mathcal{S}}{\mathbbm{1}(s\in A_*)d\mu_1(s)} \le 
    \int_{s\in\mathcal{S}}\mathbbm{1}\{s:\max_{a\in\mathcal{A}}{Q^{\pi^*}(s,a)} \\- \max_{ a\in{\mathcal{A}- \argmax_{a'}{Q^{\pi^*}(x,a')} } }{Q^{\pi^*}(s,a)}\le \epsilon\}d\mu_1(s) = O(\epsilon^\eta).
    \end{aligned}
\end{equation*} by assumption \ref{ass:a6}. Together with equation \eqref{eqa:31}, we know that \begin{equation}\label{eqa:32}
    \begin{aligned}
    \int_{s\in\mathcal{S}}{\sum_{a\in\mathcal{A}}{Q^{\pi^*}(s,a)\{\pi^*(a|s)-\pi_K(a|s)\}}\mathbbm{1}(s\in A_*)d\mu_1(s)} = O(\epsilon^{\eta+1}).
    \end{aligned}
\end{equation} 

Then we can bound the second part of equation \eqref{eqa:29},  \begin{equation*}
    \int_{s\in\mathcal{S}}{\sum_{a\in\mathcal{A}}{Q^{\pi^*}(s,a)\{\pi^*(a|s)-\pi_K(a|s)\}\mathbbm{1}(s\in A_*^C)d\mu_1(s)}}.
\end{equation*}

Recall the definition of $A_*^C=\{\max_{a\in\mathcal{A}}{Q^{\pi^*}(s,a)}-Q^{\pi^*}(s,\hat{a}_{\pi^K})>\epsilon\}\cup \{\max_{a\in\mathcal{A}}{Q^{\pi^*}(s,a)}-Q^{\pi^*}(s,\hat{a}_{\pi^K})=0\}$. We know $\{\max_{a\in\mathcal{A}}{Q^{\pi^*}(s,a)}-Q^{\pi^*}(s,\hat{a}_{\pi^K})=0\}$ is a trivial case. It is easy to show that \begin{equation*}
    \begin{aligned}
    \int_{s\in\mathcal{S}}{\sum_{a\in\mathcal{A}}{Q^{\pi^*}(s,a)\{\pi^*(a|s)-\pi_K(a|s)\}\mathbbm{1}\{s:\max_{a\in\mathcal{A}}{Q^{\pi^*}(s,a)}-Q^{\pi^*}(s,\hat{a}_{\pi^K})=0\}d\mu_1(s)}}=0,
    \end{aligned}
\end{equation*} as $\pi^*(a|s)=\pi^K(a|s)$ when $\max_{a\in\mathcal{A}}{Q^{\pi^*}(s,a)}-Q^{\pi^*}(s,\hat{a}_{\pi^K})=0$ always holds. 

Therefore, \begin{equation*}
    \begin{aligned}
    \int_{s\in\mathcal{S}}{\sum_{a\in\mathcal{A}}{Q^{\pi^*}(s,a)\{\pi^*(a|s)-\pi_K(a|s)\}\mathbbm{1}(s\in A_*^C)d\mu_1(s)}}=\\
    \int_{s\in\mathcal{S}}{\sum_{a\in\mathcal{A}}{Q^{\pi^*}(s,a)\{\pi^*(a|s)-\pi_K(a|s)\}\mathbbm{1}\{s:\max_{a\in\mathcal{A}}{Q^{\pi^*}(s,a)}-Q^{\pi^*}(s,\hat{a}_{\pi^K})>\epsilon\}d\mu_1(s)}}.
    \end{aligned}
\end{equation*}

We can show that when $\max_{a\in\mathcal{A}}{Q^{\pi^*}(s,a)}-Q^{\pi^*}(s,\hat{a}_{\pi^K})>\epsilon$ holds, \begin{equation*}
    \begin{aligned}
    \max_{a\in\mathcal{A}}{Q^{\pi^*}(s,a)}-Q^{\pi^*}(s,\hat{a}_{\pi^K}) \le 2\max_{a\in\mathcal{A}}{|Q^{\pi^*}(s,a)-\tilde{Q}_K(s,a)|}.
    \end{aligned}
\end{equation*} To show this, note that $\max_{a\in\mathcal{A}}{Q^{\pi^*}(s,a)}-Q^{\pi^*}(s,\hat{a}_{\pi^K}) = \max_{a\in\mathcal{A}}{Q^{\pi^*}(s,a)}-\tilde{Q}_K(s,\hat{a}_{\pi^K})+\tilde{Q}_K(s,\hat{a}_{\pi^K})-Q^{\pi^*}(s,\hat{a}_{\pi^K})$. It is trivial to show \begin{equation*}
    \tilde{Q}_K(s,\hat{a}_{\pi^K})-Q^{\pi^*}(s,\hat{a}_{\pi^K})\le \max_{a\in\mathcal{A}}{|Q^{\pi^*}(s,a)-\tilde{Q}_K(s,a)|}.
\end{equation*} Furthermore, we can get $\max_{a\in\mathcal{A}}{Q^{\pi^*}(s,a)}-\tilde{Q}_K(s,\hat{a}_{\pi^K}) = \max_{a\in\mathcal{A}}{Q^{\pi^*}(s,a)}-\tilde{Q}_K(s,a\in{\argmax_{a\in\mathcal{A}}{{Q^*(s,a)}}})+\tilde{Q}_K(s,a\in{\argmax_{a\in\mathcal{A}}{{Q^*(s,a)}}})-\tilde{Q}_K(s,\hat{a}_{\pi^K})$. Then we know that $\max_{a\in\mathcal{A}}{Q^{\pi^*}(s,a)}-\tilde{Q}_K(s,a\in{\argmax_{a\in\mathcal{A}}{{Q^*(s,a)}}})\le\max_{a\in\mathcal{A}}{|Q^{\pi^*}(s,a)-\tilde{Q}_K(s,a)|}$ and $\tilde{Q}_K(s,a\in{\argmax_{a\in\mathcal{A}}{{Q^*(s,a)}}})-\tilde{Q}_K(s,\hat{a}_{\pi^K})\le0$, which implies that $\max_{a\in\mathcal{A}}{Q^{\pi^*}(s,a)}-\tilde{Q}_K(s,\hat{a}_{\pi^K})\le\max_{a\in\mathcal{A}}{|Q^{\pi^*}(s,a)-\tilde{Q}_K(s,a)|}$.

To bound $\max_{a\in\mathcal{A}}{|Q^{\pi^*}(s,a)-\tilde{Q}_K(s,a)|}$, we know that \begin{equation*}
    \begin{aligned}
    \max_{a\in\mathcal{A}}{{|Q^{\pi^*}(s,a)-\tilde{Q}_K(s,a)|}^2 }\le
    \sum_{a\in\mathcal{A}}{{|Q^{\pi^*}(s,a)-\tilde{Q}_K(s,a)|}^2 }\le\\
    \frac{1}{\min_{a\in\mathcal{A}}{\mu_2(a)}}\sum_{a\in\mathcal{A}}{{|Q^{\pi^*}(s,a)-\tilde{Q}_K(s,a)|}^2 \mu_2(a)}.
    \end{aligned}
\end{equation*} 

Then, it can be obtained that \begin{equation}\label{eqa:37}
    \begin{aligned}
    \int_{s\in\mathcal{S}}{\sum_{a\in\mathcal{A}}{Q^{\pi^*}(s,a)\{\pi^*(a|s)-\pi_K(a|s)\}}\mathbbm{1}(s\in A_*^C)d\mu_1(s)}=\\
    \int_{s\in\mathcal{S}}{(\max_{a\in\mathcal{A}}{Q^{\pi^*}(s,a)}-Q^{\pi^*}(s,\hat{a}_{\pi^K}))\mathbbm{1}(s\in A_*^C)d\mu_1(s)}\le \\
    \int_{s\in\mathcal{S}}{\frac{{(2\max_{a\in\mathcal{A}}{|Q^{\pi^*}(s,a)-\tilde{Q}_K(s,a)|})}^2}{(\max_{a\in\mathcal{A}}{Q^{\pi^*}(s,a)}-Q^{\pi^*}(s,\hat{a}_{\pi^K}))}\mathbbm{1}(s\in A_*^C)d\mu_1(s)}\le\\
    \frac{4}{\epsilon\min_{a\in\mathcal{A}}{\mu_2(a)}}\int_{s\in\mathcal{S}}{\sum_{a\in\mathcal{A}}{{|Q^{\pi^*}(s,a)-\tilde{Q}_K(s,a)|}^2 \mu_2(a)}\mathbbm{1}(s\in A_*^C)d\mu_1(s)}\le\\
    \frac{4}{\epsilon\min_{a\in\mathcal{A}}{\mu_2(a)}}\int_{s\in\mathcal{S}}{\sum_{a\in\mathcal{A}}{{|Q^{\pi^*}(s,a)-\tilde{Q}_K(s,a)|}^2 \mu_2(a)}d\mu_1(s)}=\\
    \frac{4}{\epsilon\min_{a\in\mathcal{A}}{\mu_2(a)}}\left\Vert Q^{\pi^*}(s,a)-\tilde{Q}_K(s,a)\right\Vert_{\mu,2}^2.
    \end{aligned}
\end{equation}

From Theorem \ref{theorem_1}, we already have 
\begin{equation}\label{eqa:38}
    \begin{aligned}
    \left\Vert Q^{\pi^*}-\tilde{Q}_K\right\Vert_{2,\mu}^2=O_p(\gamma^2(1-\gamma)^4\times[|\mathcal{A}|n^{-1}(n^{\alpha^*}{(\log{n})}^{\xi^*}+d_1^*{\kappa}){(\log{n})}^{\xi^*+1}\\+L_{\mathcal{F}_0}{(e^{-\zeta\kappa}-e^{-\zeta m}+n^{-2\Delta})}]+\frac{\gamma^{2K}}{(1-\gamma)^2}R_{max}^2).
    \end{aligned}
\end{equation} 
Then plug in equation \eqref{eqa:38} to equation \eqref{eqa:37} and it leads to \begin{equation}\label{eqa:39}
    \begin{aligned}
    \int_{s\in\mathcal{S}}{\sum_{a\in\mathcal{A}}{Q^{\pi^*}(s,a)\{\pi^*(a|s)-\pi_K(a|s)\}}\mathbbm{1}(s\in A_*^C)d\mu_1(s)}\\=O_p(\frac{1}{\epsilon\min_{a\in\mathcal{A}}{\mu_2(a)}}\gamma^2(1-\gamma)^4\times[|\mathcal{A}|n^{-1}(n^{\alpha^*}{(\log{n})}^{\xi^*}\\+d_1^*{\kappa}){(\log{n})}^{\xi^*+1}+L_{\mathcal{F}_0}{(e^{-\zeta\kappa}-e^{-\zeta m}+n^{-2\Delta})}]+\frac{\gamma^{2K}}{(1-\gamma)^2}R_{max}^2).
    \end{aligned}
\end{equation}

Then from equation \eqref{eqa:29}, equation \eqref{eqa:32} and equation \eqref{eqa:39}, we can get \begin{equation*}
    \begin{aligned}
    \int_{s\in\mathcal{S}}{\sum_{a\in\mathcal{A}}{Q^{\pi^*}(s,a)\{\pi^*(a|s)-\pi_K(a|s)\}d\mu_1(s)}}\\=O_p(\epsilon^{\eta+1}+\frac{1}{\epsilon\min_{a\in\mathcal{A}}{\mu_2(a)}}\gamma^2(1-\gamma)^4\times[|\mathcal{A}|n^{-1}(n^{\alpha^*}{(\log{n})}^{\xi^*}\\+d_1^*{\kappa}){(\log{n})}^{\xi^*+1}+ L_{\mathcal{F}_0}{(e^{-\zeta\kappa}-e^{-\zeta m}+n^{-2\Delta})}]+\frac{\gamma^{2K}}{(1-\gamma)^2}R_{max}^2).
    \end{aligned}
\end{equation*}

If we take \begin{equation*}
    \begin{aligned}
    \epsilon^{\eta+1}=\frac{1}{\epsilon\min_{a\in\mathcal{A}}{\mu_2(a)}}\gamma^2(1-\gamma)^4\times[|\mathcal{A}|n^{-1}(n^{\alpha^*}{(\log{n})}^{\xi^*}+d_1^*{\kappa}){(\log{n})}^{\xi^*+1}+\\L_{\mathcal{F}_0}{(e^{-\zeta\kappa}-e^{-\zeta m}+n^{-2\Delta})}]+\frac{\gamma^{2K}}{(1-\gamma)^2}R_{max}^2,
    \end{aligned}
\end{equation*} then the following result is obtained: 
\begin{equation}\label{eqa:42}
    \begin{aligned}
    \int_{s\in\mathcal{S}}{\sum_{a\in\mathcal{A}}{Q^{\pi^*}(s,a)\{\pi^*(a|s)-\pi_K(a|s)\}d\mu_1(s)}} =O_p(\{\gamma^2(1-\gamma)^4\times[|\mathcal{A}|n^{-1}(n^{\alpha^*}{(\log{n})}^{\xi^*}\\+d_1^*{\kappa}){(\log{n})}^{\xi^*+1}+L_{\mathcal{F}_0}{(e^{-\zeta\kappa}-e^{-\zeta m}+n^{-2\Delta})}]+\frac{\gamma^{2K}}{(1-\gamma)^2}R_{max}^2\}^{\frac{\eta+1}{\eta+2}}).
    \end{aligned}
\end{equation}

Then using similar process, we can show the result on Lebesgue measure, \begin{equation}\label{eqa:43}
    \begin{aligned}
    \int_{s\in\mathcal{S}}{\sum_{a\in\mathcal{A}}{Q^{\pi^*}(s,a)\{\pi^*(a|s)-\pi_K(a|s)\}ds}} =O_p(\{\gamma^2(1-\gamma)^4\times[|\mathcal{A}|n^{-1}(n^{\alpha^*}{(\log{n})}^{\xi^*}\\+d_1^*{\kappa}){(\log{n})}^{\xi^*+1}+L_{\mathcal{F}_0}{(e^{-\zeta\kappa}-e^{-\zeta m}+n^{-2\Delta})}]+\frac{\gamma^{2K}}{(1-\gamma)^2}R_{max}^2\}^{\frac{\eta+1}{\eta+2}}).
    \end{aligned}
\end{equation}

Combine equation \eqref{eqa:save_e10_e11}, equation \eqref{eqa:42} and equation \eqref{eqa:43}, the result of Theorem \ref{theorem_2} can be obtained. 

\end{proof}



\subsection{Proof of Lemma \ref{lemma_6_1}}

The Lemma \ref{lemma_6_1} is:
For the estimated Q-function obtained in iteration $K$, $\tilde{Q}_K$ in Algorithm \ref{alg:algorithm1}, we have 
\begin{equation*}\label{equation_1}
    \begin{aligned}
    \left\Vert Q^*-\tilde{Q}_K\right\Vert_{2,\mu}^2 \le 2 \gamma^2(1-\gamma)^4\epsilon_{max}^2\phi_{\mu,\sigma}^{2}+8\frac{\gamma^{2K}R_{max}^2}{(1-\gamma)^2},
    \end{aligned}
\end{equation*} where $\epsilon_{max}=\max_{k\in\{1,2...,K\}}{\left\Vert T\tilde{Q}_{k-1}-\tilde{Q}_{k}\right\Vert_{\sigma}}$.

\begin{proof}
First, we denote $\rho_k=T\tilde{Q}_{k-1}-\tilde{Q}_k$ and define the operator $P^\pi$ as follow:
\begin{equation}\label{eqa:p_pi_def}
    \begin{aligned}
    (P^\pi Q)(s,a)=E[Q(S',A')|S'~P(.|s,a),A'~\pi(.|S')].
    \end{aligned}
\end{equation}
We can rely on this Lemma from \citet{Fan2020Qlearn}:
\begin{lemma}[Lemma C.2 in \citet{Fan2020Qlearn}]\label{lemma_C_2_Fan}
\begin{equation*}
    \begin{aligned}
    Q^*-\tilde{Q}_l\le \sum_{i=k}{l-1}{\gamma^{l-1-i}(P^{\pi^*})^{l-1-i}\rho_{i+1}+\gamma^{l-k}(P^{\pi^*})^{l-k}(Q^*-\tilde{Q}_k)}
    \end{aligned}
\end{equation*} for $\forall k,l\in\mathbb{Z},0\le k<l\le K-1$.
\end{lemma}
If we plug in $l=K, k=0$, we have \begin{equation*}
    \begin{aligned}
    Q^*-\tilde{Q}_K\le \sum_{i=0}{K-1}{\gamma^{K-1-i}(P^{\pi^*})^{K-1-i}\rho_{i+1}+\gamma^{K}(P^{\pi^*})^{K}(Q^*-\tilde{Q}_0)}
    \end{aligned}
\end{equation*}
We can denote $U_i=(P^{\pi^*})^{K-1-i}$ for $i=0,1,...,K-1$ and $U_K=(P^{\pi^*})^K$.
Then we have \begin{equation*}
    \begin{aligned}
    |Q^*(s,a)-\tilde{Q}_K(s,a)|\le \sum_{i=0}^{K-1}{\gamma^{K-1-i}(U_i|\rho_{i+1}|)(s,a)}+\gamma^K(U_K|Q^*-\tilde{Q}_0|)(s,a).
    \end{aligned}
\end{equation*}
By taking square on both sides, we have \begin{equation*}
    \begin{aligned}
    {|Q^*(s,a)-\tilde{Q}_K(s,a)|}^2 \le [\sum_{i=0}^{K-1}{\gamma^{K-1-i}(U_i|\rho_{i+1}|)(s,a)}]^2+2 \sum_{i=0}^{K-1}{\gamma^{K-1-i}(U_i|\rho_{i+1}|)(s,a)}\\\times \gamma^K(U_K|Q^*-\tilde{Q}_0|)(s,a)+\gamma^{2K}[(U_K|Q^*-\tilde{Q}_0|)(s,a)]^2.
    \end{aligned}
\end{equation*} That is, \begin{equation} \label{eqa:50}
    \begin{aligned}
    E_\mu[{|Q^*(s,a)-\tilde{Q}_K(s,a)|}^2] \le E_\mu\{[\sum_{i=0}^{K-1}{\gamma^{K-1-i}(U_i|\rho_{i+1}|)(s,a)}]^2\} \\+ E_\mu\{2 \sum_{i=0}^{K-1}{\gamma^{K-1-i}(U_i|\rho_{i+1}|)(s,a)}\times \gamma^K(U_K|Q^*-\tilde{Q}_0|)(s,a)\} \\+ E_\mu\{\gamma^{2K}[(U_K|Q^*-\tilde{Q}_0|)(s,a)]^2\}.
    \end{aligned}
\end{equation}

For the third term $E_\mu\{\gamma^{2K}[(U_K|Q^*-\tilde{Q}_0|)(s,a)]^2\}$, we can use the bound \begin{equation}\label{eqa:51}
    \begin{aligned}
    E_\mu\{\gamma^{2K}[(U_K|Q^*-\tilde{Q}_0|)(s,a)]^2\}\le 4\gamma^{2K}V_{max}^2=\frac{4\gamma^{2K}R_{max}^2}{(1-\gamma)^2}.
    \end{aligned}
\end{equation} This equation \eqref{eqa:51} holds because of the following reason: we know $|Q^*(s,a)-\tilde{Q}_0(s,a)|\le2V_{max},\forall (s,a)$ and this implies that $|(U_K|Q^*-\tilde{Q}_0|)(s,a)|=|({(P^{\pi^*})}^K|Q^*-\tilde{Q}_0|)(s,a)|\le 2V_{max},\forall (s,a)$.

For the cross term in \eqref{eqa:50}, we have \begin{equation}
    \begin{aligned}\label{eqa:50_cross_term}
    E_\mu\{2 \sum_{i=0}^{K-1}{\gamma^{K-1-i}(U_i|\rho_{i+1}|)(s,a)}\times \gamma^K(U_K|Q^*-\tilde{Q}_0|)(s,a)\}\le\\
    E_\mu\{[\sum_{i=0}^{K-1}{\gamma^{K-1-i}(U_i|\rho_{i+1}|)(s,a)}]^2\}+
    E_\mu\{\gamma^{2K}[(U_K|Q^*-\tilde{Q}_0|)(s,a)]^2\}.
    \end{aligned}
\end{equation}

Then the main focus should be the bound on $E_\mu\{[\sum_{i=0}^{K-1}{\gamma^{K-1-i}(U_i|\rho_{i+1}|)(s,a)}]^2\}$. We have \begin{equation*}
    \begin{aligned}
    E_\mu\{[\sum_{i=0}^{K-1}{\gamma^{K-1-i}(U_i|\rho_{i+1}|)(s,a)}]^2\}=
    \sum_{0\le i,j\le K-1}{\gamma^{2K-2-i-j}E_{\mu}[(U_i|\rho_{i+1}|)(s,a)(U_j|\rho_{j+1}|)(s,a)]}.
    \end{aligned}
\end{equation*} Then we need to handle $E_{\mu}[(U_i|\rho_{i+1}|)(s,a)(U_j|\rho_{j+1}|)(s,a)]$ for $0\le i,j\le K-1$. 

We have \begin{equation*}
    E_{\mu}[(U_i|\rho_{i+1}|)(s,a)(U_j|\rho_{j+1}|)(s,a)]=E_{\mu}[{(P^{\pi^*})}^{K-1-i}|\rho_{i+1}|(s,a){(P^{\pi^*})}^{K-1-j}|\rho_{j+1}|(s,a)].
\end{equation*}


Recall that $(P^\pi Q)(s,a)=E[Q(S',A')|S'\sim P(.|s,a),A'\sim \pi(.|S')]$.  There is one key observation \begin{equation*}
    \begin{aligned}
    [(P^\pi Q)(s,a)]^2=E[Q(S',A')|S'\sim P(.|s,a),A'\sim \pi(.|S')]\\\times E[Q(S',A')|S'\sim P(.|s,a),A'\sim \pi(.|S')]  \\\le
    E[Q^2(S',A')|S'\sim P(.|s,a),A'\sim \pi(.|S')].
    \end{aligned}
\end{equation*} Here we would abuse the notation $P^{\pi}\mu$ to represent the distribution of next stage $(S',A')$ such that $S'\sim P(.|s,a),A'\sim \pi(.|S')$ (when notation $P^{\pi}$ is followed by a function on $\mathcal{S}\times\mathcal{A}$, it is regarded as the operator defined in equation \eqref{eqa:p_pi_def}); when it is followed by a distribution on $\mathcal{S}\times\mathcal{A}$, it is regarded as a transformation of distribution as is discussed here.   

Then it can be shown that \begin{equation*}
    \begin{aligned}
    E_\mu\{[(P^\pi Q)(s,a)]^2\}\le E_\mu\{E[Q^2(S',A')|S'\sim P(.|s,a),A'\sim \pi(.|S')]\}
    =E_{P^\pi \mu}[Q^2(s,a)].
    \end{aligned}
\end{equation*} We can further get \begin{equation*}
    \begin{aligned}
    E_\mu\{[(P^{\pi_1}P^{\pi_2}...P^{\pi_l} Q)(s,a)]^2\}\le E_{{\pi_1}P^{\pi_2}...P^{\pi_l} \mu}[Q^2(s,a)].
    \end{aligned}
\end{equation*} For simplicity, let's denote ${\pi_1}P^{\pi_2}...P^{\pi_l} \mu$ as $\tilde{\mu}_l$.

For two functions $f_1,f_2$ on $\mathcal{S}\times\mathcal{A}$, we know \begin{equation*}
    \begin{aligned}
    E_\mu[(P^{\pi^*})^{K-1-i}f_1(P^{\pi^*})^{K-1-j}f_2]\le \{E_\mu\{[(P^{\pi^*})^{K-1-i}f_1]^2\}E_\mu\{[(P^{\pi^*})^{K-1-j}f_2]^2\}\}^{\frac{1}{2}}.
    \end{aligned}
\end{equation*}

It can be obtained that \begin{equation*}
    \begin{aligned}
    E_\mu\{[(P^{\pi^*})^{K-1-i}f_1]^2\}\le E_{(P^{\pi^*})^{K-1-i}\mu}[f_1^2]=\int_{\mathcal{S}\times\mathcal{A}}{f_1^2(s,a)d\tilde{\mu}_{K-1-i}(s,a)}
    \\=\int_{\mathcal{S}\times\mathcal{A}}{|f_1^2(s,a)\frac{d\tilde{\mu}_{K-1-i}(s,a)}{d\sigma(s,a)}|d\sigma(s,a)}
    \le\int_{\mathcal{S}\times\mathcal{A}}{|f_1^2(s,a)|d\sigma(s,a)}\times \left\Vert \frac{d\tilde{\mu}_{K-1-i}(s,a)}{d\sigma(s,a)}\right\Vert_{\infty,\sigma}
    \\=\left\Vert f_1\right\Vert_{2,\sigma}^2 \times \omega_\infty(K-1-i;\mu,\sigma). 
    \end{aligned}
\end{equation*} Recall that definition of $\omega_\infty(i;\mu,\sigma)$ in equation \eqref{eqa:def_concentration_coef}. 
Similarly, we can have \begin{equation*}
    \begin{aligned}
    E_\mu\{[(P^{\pi^*})^{K-1-j}f_2]^2\}\le\left\Vert f_2\right\Vert_{2,\sigma}^2 \times \omega_\infty(K-1-j;\mu,\sigma).
    \end{aligned}
\end{equation*}

If we plug in $f_1=|\rho_{i+1}|$ and $f_2=|\rho_{j+1}|$, we can get \begin{equation*}
    \begin{aligned}
    E_{\mu}[{(P^{\pi^*})}^{K-1-i}|\rho_{i+1}|(s,a){(P^{\pi^*})}^{K-1-j}|\rho_{j+1}|(s,a)] \\
    \le \{E_\mu\{[(P^{\pi^*})^{K-1-i}|\rho_{i+1}|]^2\} E_\mu\{[(P^{\pi^*})^{K-1-j}|\rho_{j+1}|]^2\} \}^{\frac{1}{2}}\\
    \le \left\Vert \rho_{i+1}\right\Vert_{2,\sigma} \left\Vert \rho_{j+1}\right\Vert_{2,\sigma} [\omega_{\infty}(K-1-i;\mu,\sigma)\omega_{\infty}(K-1-j;\mu,\sigma)]^{\frac{1}{2}}.
    \end{aligned}
\end{equation*}

Then, \begin{equation*}
    \begin{aligned}
    E_\mu\{[\sum_{i=0}^{K-1}{\gamma^{K-1-i}(U_i|\rho_{i+1}|)(s,a)}]^2\}
    \le \sum_{0\le i,j\le K-1}\gamma^{2K-2-i-j}\left\Vert \rho_{i+1}\right\Vert_{2,\sigma} \left\Vert \rho_{j+1}\right\Vert_{2,\sigma}\\\times [\omega_{\infty}(K-1-i;\mu,\sigma)\omega_{\infty}(K-1-j;\mu,\sigma)]^{\frac{1}{2}}.
    \end{aligned}
\end{equation*}

We can define \begin{equation*}
    \begin{aligned}
    \epsilon_{max}=\max_{i\in\{1,2...,K\}}{\left\Vert \rho_{i}\right\Vert_{2,\sigma}}=\max_{i\in\{1,2...,K\}}{\left\Vert T\tilde{Q}_{i-1}-\tilde{Q}_i\right\Vert_{2,\sigma}},
    \end{aligned}
\end{equation*} so that $\epsilon_{max}^2=max_{i\in\{1,2...,K\}}\left\Vert\rho_{i}^2\right\Vert_{2,\sigma}$.
Thus, it can be shown that \begin{equation*}
    \begin{aligned}
    E_\mu\{[\sum_{i=0}^{K-1}{\gamma^{K-1-i}(U_i|\rho_{i+1}|)(s,a)}]^2\}\le\epsilon_{max}^2\{\sum_{i=0}^{K-1}{\gamma^{i}[\omega_{\infty}(i;\mu,\sigma)]^\frac{1}{2}}\}^2.
    \end{aligned}
\end{equation*}

We've already shown that $\frac{1}{(1-\gamma)^2}\sum_{m=0}^{\infty}{\gamma^{m-1}(m+1)[\omega_\infty(m;\mu,\sigma)]^{\frac{1}{2}}}\le \phi_{\mu,\sigma}$. Then, we know \begin{equation*}
    \begin{aligned}
    \sum_{i=0}^{K-1}{\gamma^{i}[\omega_{\infty}(i;\mu,\sigma)]^\frac{1}{2}}<(1-\gamma)^2\gamma\times\frac{1}{(1-\gamma)^2}\sum_{i=0}^{\infty}{\gamma^{i-1}(i+1)[\omega_{\infty}{i;\mu,\sigma}]^{\frac{1}{2}}}\le (1-\gamma)^2\gamma\phi_{\mu,\sigma}.
    \end{aligned}
\end{equation*}
Hence, \begin{equation}\label{eqa:66}
    \begin{aligned}
    E_\mu\{[\sum_{i=0}^{K-1}{\gamma^{K-1-i}(U_i|\rho_{i+1}|)(s,a)}]^2\}<\epsilon_{max}^2 (1-\gamma)^4\gamma^2\phi_{\mu,\sigma}^2.
    \end{aligned}
\end{equation}

Then combine equation \eqref{eqa:50}, equation \eqref{eqa:51}, equation \eqref{eqa:50_cross_term} and equation \eqref{eqa:66}, we can have the result of Lemma \ref{lemma_6_1}.


\end{proof}

\subsection{Proof of Proposition \ref{statement_1}}
The Proposition \ref{statement_1} is: The function class \begin{equation*}
    \mathcal{F}_1(L,\{\{d_i\}^{L+1}_{i=1},d_0=m+m_0\},s,V_{max},\hat{G},\kappa)
\end{equation*} is a subset of \begin{equation*}
    \mathcal{F}_2(L,\{\{d_i\}^{L+1}_{i=1},d_0=\kappa+m_0\},s+d_1\kappa,V_{max},\hat{G},\kappa).
\end{equation*}
\begin{proof}
Note that the vector $z^*$ recovered from the first $\kappa$ PCA values of $z$ has a relation with the corresponding vector containing the $\kappa$ PCA values: $z^*=\hat{U}{\hat{\Sigma}}^{\frac{1}{2}}\begin{pmatrix}
I_{\kappa} \\ 0_{(m-\kappa)\times\kappa}
\end{pmatrix}\hat{v}$. By Definition \ref{def_F_12} in Section \ref{sec:def_two_func_class}, $\forall f_1\in\mathcal{F}_1(L,\{\{d_i\}^{L+1}_{i=1},d_0=m+m_0\},s,V_{max},\hat{G},\kappa)$, we have $f_1(x,z,a)=f_0^1(x,z^*,a)$ with $f_0^1(.,a)\in\mathcal{F}_{SReLU}(L,\{d_i\}_{i=0}^{L+1},s,V_{max})$. We can denote the first layer in the sparse ReLU network $f_0^1(.,a)$ by $\sigma(W_1^1\begin{pmatrix}x\\z^*\end{pmatrix}+b_1^1)$. We know $W_1^1\begin{pmatrix}x\\z^*\end{pmatrix}=\begin{pmatrix}W_{11}^1 & W_{12}^1\end{pmatrix}\begin{pmatrix}x\\z^*\end{pmatrix}=W_{11}^1 x+W_{12}^1 z^*$, with $W_{12}^1 z^*=W_{12}^1 \hat{U}{\hat{\Sigma}}^{\frac{1}{2}}\begin{pmatrix}
I_{\kappa} \\ 0_{(m-\kappa)\times\kappa}
\end{pmatrix}\hat{v}$.

Then we would like to know $\exists f_2\in\mathcal{F}_2(L,\{\{d_j\}^{L+1}_{j=1},d_0=\kappa+m_0\},s+d_1\kappa,V_{max},\hat{G},\kappa)$, such that $f_2(x,z,a)=f_1(x,z,a)$. We know $f_2(x,z,a)=f_0^2(x,\hat{v},a)$ with \begin{equation*}
    f_0^2(.,a)\in\mathcal{F}_{SReLU}(L,\{d_i\}_{i=0}^{L+1},s+\kappa d_1,V_{max})
\end{equation*} by definition. The first layer of $f_0^2$ can be denoted in a similar way: $\sigma(W_1^2\begin{pmatrix}x\\\hat{v}\end{pmatrix}+b_1^2)$, with $W_1^2\begin{pmatrix}x\\\hat{v}\end{pmatrix}=\begin{pmatrix}W_{11}^2 & W_{12}^2\end{pmatrix}\begin{pmatrix}x\\\hat{v}\end{pmatrix}=W_{11}^2 x+W_{12}^2 \hat{v}$. We can set $W_{11}^2=W_{11}^1,b_1^1=b_1^2$ and $W_{12}^2=W_{12}^1 \hat{U}{\hat{\Sigma}}^{\frac{1}{2}}\begin{pmatrix}
I_{\kappa} \\ 0_{(m-\kappa)\times\kappa}
\end{pmatrix}$, which gives $\sigma(W_1^1\begin{pmatrix}x\\z^*\end{pmatrix}+b_1^1)=\sigma(W_1^2\begin{pmatrix}x\\\hat{v}\end{pmatrix}+b_1^2)$. The parameters in other layers in $f_0^2$ will be set to be the same as $f_0^1$.

Then we know $\forall f_1\in\mathcal{F}_1(L,\{\{d_j\}^{L+1}_{j=1},d_0=m+m_0\},s,V_{max},\hat{G},\kappa)$, $\exists f_2\in\mathcal{F}_2(L,\{\{d_j\}^{L+1}_{j=1},d_0=\kappa+m_0\},s+d_1\kappa,V_{max},\hat{G},\kappa)$ such that \begin{equation*}
    f_1(x,z,a)=f_0^1(x,z^*,a)=f_0^2(x,\hat{v},a)=f_2(x,z,a).
\end{equation*} The reason why the sparsity in $f_0^2$ is increased to $s+\kappa d_1$ is that the weight $W_{12}^2=W_{12}^1 \hat{U}{\hat{\Sigma}}^{\frac{1}{2}}\begin{pmatrix}
I_{\kappa} \\ 0_{(m-\kappa)\times\kappa}
\end{pmatrix}\in\mathbb{R}^{d_1\times\kappa}$ may no longer be sparse, despite that $W_{12}^1$ is sparse. 

\end{proof}

\subsection{Proof of Lemma \ref{lemma_1}}
The Lemma \ref{lemma_1} is: Under the same assumptions of the Theorem \ref{theorem_1},  
\begin{equation*}
\sup_{g\in\mathcal{F}_{2}}{\inf_{f\in\mathcal{F}_{2}}{\left\Vert f-Tg \right\Vert_\sigma^2}}\le 2\sup_{f^{'}\in\mathcal{G}_{0}}{\inf_{f_0\in\mathcal{F}_{0}}{\left\Vert f_0-f^{'} \right\Vert_\sigma^2}}+2L_{\mathcal{F}_0}{\left\Vert Z-Z^{*}\right\Vert_{\sigma}^{2}},
\end{equation*}
 where $L_{\mathcal{F}_0}=\sup_{f\in\mathcal{F}_0}{\sup_{x\ne y}{\frac{{|f(y)-f(x)|}^{2}}{\left\Vert y-x\right\Vert^{2}}}}$. 
$Z$ is the high-frequency vector from the distribution $\sigma$ and $Z^{*}=(\sum_{k=1}^{\kappa_j}{\hat{U}_k{\hat{U}_k}^{T}})Z$ is the vector recovered from $\kappa$ PCA values of $Z$.

\begin{proof}
We have $\sup_{g\in\mathcal{F}_{2}}{\inf_{f\in\mathcal{F}_{2}}{\left\Vert f-Tg \right\Vert_\sigma^2}} \le \sup_{f'\in\mathcal{G}_{0}}{\inf_{f_2\in\mathcal{F}_{2}}{\left\Vert f_2-f' \right\Vert_\sigma^2}}$, which relies on the assumption: $Tg\in\mathcal{G}_0,\forall g\in\mathcal{F}_2$.  From the Proposition \ref{statement_1}, we have $\mathcal{F}_{1}\subset\mathcal{F}_{2}$. Then we have $\sup_{f'\in\mathcal{G}_{0}}{\inf_{f_2\in\mathcal{F}_{2}}{\left\Vert f_2-f' \right\Vert_\sigma^2}} \le \sup_{f'\in\mathcal{G}_{0}}{\inf_{f_1\in\mathcal{F}_{1}}{\left\Vert f_1-f' \right\Vert_\sigma^2}}$.

Then we know  \begin{equation*}
\begin{aligned}
    \sup_{f'\in\mathcal{G}_{0}}{\inf_{f_1\in\mathcal{F}_{1}}{\left\Vert f_1-f' \right\Vert_\sigma^2}}=\sup_{f'\in\mathcal{G}_{0}}{\inf_{f_1\in\mathcal{F}_{1}}{E_\sigma\{{[f_1(x,z,a)-f'(x,z,a)]}^2\}}}\\=\sup_{f'\in\mathcal{G}_{0}}{\inf_{f_0\in\mathcal{F}_{0}}{E_\sigma\{{[f_0(x,z^*,a)-f'(x,z,a)]}^2\}}}
\end{aligned}
\end{equation*} by definition of $\mathcal{F}_1$.

We know \begin{equation*}
    \begin{aligned}
    E_\sigma\{{[f_0(x,z^*,a)-f'(x,z,a)]}^2\}=E_\sigma\{{[f_0(x,z^*,a)-f_0(x,z,a)+f_0(x,z,a)-f'(x,z,a)]}^2\}\\\le 2E_\sigma\{{[f_0(x,z^*,a)-f_0(x,z,a)]}^2\}+2E_\sigma\{{[f_0(x,z,a)-f'(x,z,a)]}^2\}.
    \end{aligned}
\end{equation*} Furthermore, the term $E_\sigma\{{[f_0(x,z^*,a)-f_0(x,z,a)]}^2\}$ satisfies that
\begin{equation*}
     E_\sigma\{{[f_0(x,z^*,a)-f_0(x,z,a)]}^2\} \\ \le L_{\mathcal{F}_0}{\left\Vert Z-Z^{*}\right\Vert_{\sigma}^{2}}.
\end{equation*}

Note that $L_{\mathcal{F}_0}$ is the Lipschitz constant of ReLU networks in the class $\mathcal{F}_0$. Based on chain rule and the fact that $\left\Vert W \right\Vert_2^2 \le m_W  \left\Vert W \right\Vert_{\infty}^2$ for matrix $W$ with $m_W$ columns, we know that $L_{\mathcal{F}_0}$ is always bounded: $L_{\mathcal{F}_0}\le\prod_{l}^{L}{d_{l-1}}$.

So that we know \begin{equation*}
    E_\sigma\{{[f_0(x,z^*,a)-f'(x,z,a)]}^2\}\le 2L_{\mathcal{F}_0}\left\Vert Z-Z^{*}\right\Vert_{\sigma}^{2}+2E_\sigma\{{[f_0(x,z,a)-f'(x,z,a)]}^2\},
\end{equation*} $\forall f_0\in\mathcal{F}_0,f'\in\mathcal{G}_0$. Then we can take $\sup$ and $\inf$ on both sides to have \begin{equation*}
    \begin{aligned}
    \sup_{f'\in\mathcal{G}_{0}}{\inf_{f_0\in\mathcal{F}_{0}}{E_\sigma\{{[f_0(x,z^*,a)-f'(x,z,a)]}^2\}}}\le 2L_{\mathcal{F}_0}{\left\Vert Z-Z^{*}\right\Vert_{\sigma}^{2}} +\\2\sup_{f'\in\mathcal{G}_{0}}{\inf_{f_0\in\mathcal{F}_{0}}{E_\sigma\{{[f_0(x,z,a)-f'(x,z,a)]}^2\}}},
    \end{aligned}
\end{equation*} with $E_\sigma\{{[f_0(x,z,a)-f'(x,z,a)]}^2\}=\left\Vert f_0-f' \right\Vert_\sigma^2$.

So that we eventually get \begin{equation*}
    \sup_{g\in\mathcal{F}_{2}}{\inf_{f\in\mathcal{F}_{2}}{\left\Vert f-Tg \right\Vert_\sigma^2}}\le 2L_{\mathcal{F}_0}{\left\Vert Z-Z^{*}\right\Vert_{\sigma}^{2}} +2\sup_{f'\in\mathcal{G}_{0}}{\inf_{f_0\in\mathcal{F}_{0}}{\left\Vert f_0-f' \right\Vert_\sigma^2}}.
\end{equation*} 

\end{proof}

\subsection{Proof of Lemma \ref{lemma_2}}
The Lemma \ref{lemma_2} is: For the high-frequency vector $Z$, we assume $E(Z)=0$ and the the eigenvalues of $Cov(Z)$ following this exponential decaying trend $\lambda_j=O(e^{-\zeta j})$. We also assume the estimation of the $Cov(Z)$ satisfies that $\left\Vert \hat{U}_j-U_j\right\Vert=O_p(n^{-\Delta})$. Then we have
\begin{equation*}
\left\Vert  Z-Z^{*}\right\Vert_{\sigma}^2=O_p(e^{-\zeta \kappa}-e^{-\zeta m} + n^{-2\Delta} )     
\end{equation*}

\begin{proof}
Note that \begin{equation*}
    \begin{aligned}
    \left\Vert  Z-Z^{*}\right\Vert_{\sigma}^2=E[{(Z-Z^*)}^T(Z-Z^*)]=E[Z^T{(\sum_{j=\kappa+1}^{m}{\hat{U}_j{\hat{U}_j}^T})}^2 Z]=\sum_{j=\kappa+1}^{m}{E[tr(Z^T\hat{U}_j{\hat{U}_j}^T Z)]}\\=\sum_{j=\kappa+1}^{m}{tr[E(Z Z^T)\hat{U}_j{\hat{U}_j}^T]}=tr[E(Z Z^T)\sum_{j=\kappa+1}^{m}{\hat{U}_j{\hat{U}_j}^T}].
    \end{aligned}
\end{equation*}

We know $E(Z Z^T)=Cov(Z)=\sum_{i=1}^{m}{\lambda_i U_i{U_i}^T}$. So that we will have \begin{equation*}
    \begin{aligned}
    \left\Vert  Z-Z^{*}\right\Vert_{\sigma}^2=tr[\sum_{i=1}^{m}{\lambda_i U_i{U_i}^T}\sum_{j=\kappa+1}^{m}{ \hat{U}_j{\hat{U}_j}^T}]=tr[\sum_{i=1}^{m}{\lambda_i U_i{U_i}^T}\sum_{j=\kappa+1}^{m}{ (\hat{U}_j-U_j+U_j){(\hat{U}_j-U_j+U_j})}^T]\\= tr[\sum_{i=1}^{m}{\lambda_i U_i{U_i}^T}\sum_{j=\kappa+1}^{m}{ U_j{U_j}^T}] +  tr[\sum_{i=1}^{m}{\lambda_i U_i{U_i}^T}\sum_{j=\kappa+1}^{m}{ (\hat{U}_j-U_j){U_j}^T}] \\+tr[\sum_{i=1}^{m}{\lambda_i U_i{U_i}^T}\sum_{j=\kappa+1}^{m}{ U_j{(\hat{U}_j-U_j)}^T}] +tr[\sum_{i=1}^{m}{\lambda_i U_i{U_i}^T}\sum_{j=\kappa+1}^{m}{ (\hat{U}_j-U_j){(\hat{U}_j-U_j)}^T}].
    \end{aligned}
\end{equation*} That is, we need to bound four terms here.

The first term \begin{equation*}
    tr[\sum_{i=1}^{m}{\lambda_i U_i{U_i}^T}\sum_{j=\kappa+1}^{m}{ U_j{U_j}^T}]=tr[\sum_{j=\kappa+1}^{m}{ \lambda_j U_j {U_j}^T U_j {U_j}^T}] = \sum_{j=\kappa+1}^{m}{tr(\lambda_j U_j  {U_j}^T)}=\sum_{j=\kappa+1}^{m}{\lambda_j}.
\end{equation*} Since we already know that the eigenvalues $\lambda_j$ satisfies that $\lambda_j=O(e^{-\zeta j})$, so that we have $\sum_{j=\kappa+1}^{m}{\lambda_j}=O(e^{-\zeta (\kappa+1)}\frac{1-e^{-\zeta(m-\kappa)}}{1-e^{-\zeta}})=O(e^{-\zeta \kappa}-e^{-\zeta m})$.  Then we know the first term \begin{equation*}
    tr[\sum_{i=1}^{m}{\lambda_i U_i{U_i}^T}\sum_{j=\kappa+1}^{m}{ U_j{U_j}^T}]=O(e^{-\zeta \kappa}-e^{-\zeta m}).
\end{equation*}

The second term is \begin{equation*}
    \begin{aligned}
    tr[\sum_{i=1}^{m}{\lambda_i U_i{U_i}^T}\sum_{j=\kappa+1}^{m}{ (\hat{U}_j-U_j){U_j}^T}]=\sum_{i=1}^{m}{\sum_{j=\kappa+1}^{m}{tr[\lambda_i U_i {U_i}^T (\hat{U_j}-U_j){U_j}^T]}}\\=\sum_{i=1}^{m}{\sum_{j=\kappa+1}^{m}{tr[\lambda_i {U_j}^T U_i {U_i}^T (\hat{U_j}-U_j)]}}=\sum_{j=\kappa+1}^{m}{\lambda_j {U_j}^T (\hat{U}_j-U_j)}.
    \end{aligned}
\end{equation*}  We know that each term of $U_j$ is $O_p(1)$, so that $|{U_j}^T (\hat{U}_j-U_j)|\le \left\Vert  {U_j}^T \right\Vert\left\Vert  \hat{U}_j-U_j\right\Vert=O_p(n^{-\Delta})$ and $\sum_{j=\kappa+1}^{m}{\lambda_j {U_j}^T(\hat{U}_j-U_j)}=O_p(n^{-\Delta}(e^{-\zeta \kappa}-e^{-\zeta m}))$.

The third term is similar to the second term, we have \begin{equation*}
    \begin{aligned}
    tr[\sum_{i=1}^{m}{\lambda_i U_i{U_i}^T}\sum_{j=\kappa+1}^{m}{ U_j{\hat{U}_j-U_j}^T}]=\sum_{i=1}^{m}{\sum_{j=\kappa+1}^{m}{tr[\lambda_i U_i {U_i}^T U_j{(\hat{U_j}-U_j)}^T]}}\\=\sum_{j=\kappa+1}^{m}{\lambda_j{(\hat{U}_j-U_j)}^T U_j}=O_p(n^{-\Delta}(e^{-\zeta \kappa}-e^{-\zeta m})).
    \end{aligned}
\end{equation*}

Then the fourth term is \begin{equation*}
    \begin{aligned}
    tr[\sum_{i=1}^{m}{\lambda_i U_i{U_i}^T}\sum_{j=\kappa+1}^{m}{ (\hat{U}_j-U_j){(\hat{U}_j-U_j)}^T}]=\sum_{i=1}^{m}{\sum_{j=\kappa+1}^{m}{tr[\lambda_i U_i {U_i}^T (\hat{U_j}-U_j){(\hat{U_j}-U_j)}^T]}}\\=\sum_{i=1}^{m}{\sum_{j=\kappa+1}^{m}{tr[\lambda_i  {U_i}^T (\hat{U_j}-U_j){(\hat{U_j}-U_j)}^T U_i]}}=\sum_{i=1}^{m}{\sum_{j=\kappa+1}^{m}{\lambda_i  [{U_i}^T (\hat{U_j}-U_j)]^2}}\\=O_p(n^{-2\Delta}e^{-\zeta})=O_p(n^{-2\Delta}).
    \end{aligned}
\end{equation*}

Thus, it can be shown that \begin{equation*}
    \left\Vert  Z-Z^{*}\right\Vert_{\sigma}^2=O_p((e^{-\zeta \kappa}-e^{-\zeta m}) + n^{-\Delta}(e^{-\zeta \kappa}-e^{-\zeta m})+ n^{-2\Delta} )=O_p(e^{-\zeta \kappa}-e^{-\zeta m} + n^{-2\Delta} ).
\end{equation*}

\end{proof}

\subsection{Proof of Proposition \ref{statement_2}}
The Proposition \ref{statement_2} is: When $\delta=\frac{1}{n}$, $\log{N_{\delta,2}}\le C_1|\mathcal{A}|(n^{\alpha^*}(\log{n})^{\xi^{*}}+d_1^*\kappa)(\log{n})^{1+\xi^{*}}$ for some constant $C_1$.

\begin{proof}
To prove this result, we try to show for the function class $\mathcal{F}_3=\{f_3:S\rightarrow\mathbb{R}: f_3(x,z)=f(x,\begin{pmatrix}I_\kappa & 0_{\kappa\times(m-\kappa)}\end{pmatrix}{\hat{\Sigma}}^{-\frac{1}{2}} {\hat{U}}^T z)=f(x,\hat{v}),f\in\mathcal{F}_{SReLU}(L^*,{\{d_j^*\}}_{j=0}^{L+1},s^*,V_{max}),   d_0=\kappa+m_0\}$. Recall the definition of $\mathcal{F}_2$ at Section \ref{sec:def_two_func_class}, we know that $\forall f\in\mathcal{F}_2$, $f(.,a)\in\mathcal{F}_3$. That is, $\mathcal{F}_3$ is the function class for modeling action component $a$ of $Q(s,a)$. There is a relation between cardinality of the minimal $\sigma$-covering set of $\mathcal{F}_2$, $N_{\frac{1}{n},2}$ and cardinality of the minimal $\sigma$-covering set of $\mathcal{F}_3$, $N_{\frac{1}{n},3}$. With similar arguments to the step $|\mathcal{N}(\delta,\mathcal{F}_0,\left\Vert.\right\Vert_\infty)|\le{|\mathcal{N}_\delta|}^{|\mathcal{A}|}$ in equation (6.26) of \citet{Fan2020Qlearn}, we can show that \begin{equation}\label{eqa:cardinality_relation}
    \log{N_{\frac{1}{n},2}}\le |\mathcal{A}|\log{N_{\frac{1}{n},3}}.
\end{equation}

Here we can argue that $\mathcal{F}_3$ is also a type of sparse ReLU network and directly use Lemma 6.4 in \citet{Fan2020Qlearn}. We can show that $\mathcal{F}_3\subset\mathcal{F}_{SReLU,3}$, where $\mathcal{F}_{SReLU,3}$ is a simplified notation for $\mathcal{F}_{SReLU}(L^*+1, \{{\{d_i\}}_{i=0}^{L+2}:d_0=m, d_{i+1}=d_i^*,i\in\{0,1,2,...,L+1\},s^*+m\kappa,V_{max})$. That is, the one extra layer of $\mathcal{F}_3$ is actually the linear transition $\hat{v}=\begin{pmatrix}I_\kappa & 0_{\kappa\times(m-\kappa)}\end{pmatrix}{\hat{\Sigma}}^{-\frac{1}{2}} {\hat{U}}^T z$. We don't know whether the weight matrix $\begin{pmatrix}I_\kappa & 0_{\kappa\times(m-\kappa)}\end{pmatrix}{\hat{\Sigma}}^{-\frac{1}{2}} {\hat{U}}^T $ is sparse, so that we need to increase the original sparsity to $s^*+m\kappa$.

The cardinality of $\delta$-covering set of sparse ReLU network is already bounded in Lemma 6.4 of \citet{Fan2020Qlearn}, and we can apply it here. We know the cardinality of $\delta$-covering set of $\mathcal{F}_3$,$N_{\frac{1}{n},3}$ satisfies that \begin{equation*}
    \begin{aligned}
    \log{N_{\frac{1}{n},3}}\le \log{|\mathcal{N}(\delta=\frac{1}{n},\mathcal{F}_{SReLU,3},\left\Vert .\right\Vert_\infty)|}\le(s^*+m\kappa+1)\log{[2n(L^*+2)\prod_{l=0}^{L+1}{d_l^*+1}]}
    \end{aligned}
\end{equation*}.

Here $L^*=O({(\log{n})}^{\xi^*})$,$m+m_0\le min_{j\in\{1,2,...,L\}}{d_j^*}\le max_{j\in\{1,2,...,L\}}{d_j^*}=O(n^{\xi^*})$ and $s^*\asymp n^{\alpha^*}{(\log{n})}^{\xi^*}+d_1^*\kappa$ for some constant $\xi^*>1+2\xi$. So that \begin{equation*}
    \begin{aligned}
    \log{N_{\frac{1}{n},3}} \le C (n^{\alpha^*}{(\log{n})}^{\xi^*}+d_1^*\kappa+m\kappa)\log{[2n(L^*+2)\prod_{l=0}^{L+1}{d_l^*+1}]} \\ \le C'(n^{\alpha^*}{(\log{n})}^{\xi^*}+d_1^*\kappa)\log{[2n(L^*+2)\prod_{l=0}^{L+1}{d_l^*+1}]}\\ \le C''(n^{\alpha^*}{(\log{n})}^{\xi^*}+d_1^*\kappa)[\log{n}+\log{L^*}+L^* \log{(n^{\xi^*})}]\\ \le C_1 (n^{\alpha^*}{(\log{n})}^{\xi^*}+d_1^*\kappa) {(\log{n})}^{\xi^*+1},
    \end{aligned}
\end{equation*} where $C,C',C'',C_1$ are all constants.

Then from equation \eqref{eqa:cardinality_relation}, we know $\log{N_{\frac{1}{n},2}}\le C_1|\mathcal{A}| (n^{\alpha^*}{(\log{n})}^{\xi^*}+d_1^*\kappa) {(\log{n})}^{\xi^*+1}$.
\end{proof}

\section{Details of Selecting Number of Principal Components}\label{s:appendix_kappa_select}
An optimal number of principal components, $\kappa^*$ that balances the bias and variance follows the order of $\kappa^*\asymp \log(n)$.
\begin{proof}
We know the terms involving $\kappa$ in Theorem \ref{theorem_2} is $\kappa|\mathcal{A}|d_1^* n^{-1}\log^{\xi^*+1}{n}+ C L_{\mathcal{F}_0}e^{-\zeta\kappa}$ for some constant $C>0$. We can take the derivative of this term with respect to $\kappa$. Then we know when $\kappa^*=\frac{1}{\zeta}[\log(C\zeta L_{\mathcal{F}_0}{|\mathcal{A}|}^{-1} {d_1^*}^{-1}) + \log{n}-(\xi^*+1)\log(\log{n})]$, this term will be minimized. So that we know the optimal $\kappa^*$ that balances the bias and variance must satisfy $\kappa^*\asymp \log(n)$.

This $\log{n}$ guideline applies when $n$ is large, as it is derived based on the theorems of asymptotic trend. On the other hand, $\kappa\le m$ needs to hold, as number of principal components cannot exceed the original dimension. Thus, when $n$ increases to infinity, eventually we will take $\kappa^* = m$. That is, when training sample size is extremely large, we no longer need to reduce the dimensions of high frequency part and our Algorithm \ref{alg:algorithm1} will be equivalent to the original neural fitted Q-iteration.

\end{proof}




\begin{table}[h!]
\caption{Estimate of value difference of four policies.}
\label{t:table1}
\vskip 0.15in
\begin{center}

\begin{sc}
\begin{tabular}{lcccr}
\toprule
Difference & $\pi_k^{PCA}-\pi_k^{ALL}$ & $\pi_k^{PCA}-\pi_k^{AVE}$ & $\pi_k^{PCA}-\pi_k^{BOTTLE}$\\
\midrule
Mean    & 1.163 & 1.399 & 1.153  \\
Margin of Error  & 0.181 & 0.224 & 0.145 \\
\bottomrule
\end{tabular}
\end{sc}
\end{center}
\vskip -0.1in
\end{table}

\begin{table}[h!]
\caption{Value estimate of the four policies.}
\label{t:supplementary_table}
\vskip 0.15in
\begin{center}

\begin{sc}
\begin{tabular}{lcccr}
\toprule
Value Estimate & $\pi_k^{PCA}$ & $\pi_k^{ALL}$ & $\pi_k^{AVE}$ & $\pi_k^{BOTTLE}$\\
\midrule
Mean & -8.762 & -9.926 & -10.162 & -9.916 \\
Standard Error & 0.055 & 0.096 & 0.121 & 0.086 \\
\bottomrule
\end{tabular}
\end{sc}
\end{center}
\vskip -0.1in
\end{table}

\end{document}